%% file: new_main.tex
\documentclass{article}
\usepackage[letterpaper]{geometry}  % nice margins

\usepackage{siunitx}
\usepackage{amsmath,amssymb}
\usepackage[square,numbers,compress]{natbib}
\usepackage{bbm}
\usepackage{dsfont}
\usepackage{xcolor}
\usepackage{lscape}
\usepackage{comment}
\usepackage{float}
\usepackage[demo]{graphicx}
\usepackage[ruled,vlined,linesnumbered]{algorithm2e}
\usepackage[font=small,labelfont=bf]{caption}
\usepackage{enumitem}
\usepackage{wrapfig}
\usepackage{hyperref}
\hypersetup{colorlinks=true,urlcolor=blue,citecolor=brown}  % Avoid clash on Arxiv
\usepackage{subcaption}
\usepackage{booktabs}
\usepackage[utf8]{inputenc}
\usepackage{authblk}
\usepackage[capitalize]{cleveref}
\usepackage{changepage}
\usepackage{multirow}
\usepackage{longtable}
\usepackage{tabularx}
\usepackage{algorithmic}

\DeclareUnicodeCharacter{2212}{-}
\newcommand{\bigCI}{\mathrel{\text{\scalebox{1.07}{$\perp\mkern-10mu\perp$}}}}
\definecolor{pipe_green}{rgb}{0.44, 0.68, 0.28}
\definecolor{pipe_blue}{rgb}{0.27, 0.45, 0.85}
\definecolor{pipe_orange}{rgb}{0.93, 0.49, 0.19}

\newtheorem{assumption}{Assumption}
\Crefname{table}{Table}{Tables}

\title{From Observational Data to Clinical Recommendations: A Causal Framework for Estimating Patient-level Treatment Effects and Learning Policies}

\author[1]{Rom Gutman}
\author[1]{Shimon Sheiba}
\author[2]{Omer Noy Klein}
\author[1]{Naama Dekel Bird}
\author[3,4]{Amit Gruber}
\author[3,4]{Doron Aronson}
\author[3,4]{Oren Caspi}
\author[1,5]{Uri Shalit}
\affil[1]{Faculty of Data and Decision Sciences, Technion, Israel}
\affil[2]{Blavatnik School of Computer Science, Tel-Aviv University, Israel}
\affil[3]{Department of Cardiology, Rambam Health Care Campus, Israel}
\affil[4]{The Bruce Rappaport Faculty of Medicine, Technion, Haifa, Israel}
\affil[5]{Department of Statistics and Operations Research, School of Mathematical Sciences, Tel Aviv University, Israel}
\date{}
\begin{document}

\maketitle

\begin{abstract}
We propose a framework for building patient-specific treatment recommendation models, building on the large recent literature on learning patient-level causal models and inspired by the target trial paradigm of \citet{Hernan2016UsingAvailable}. We focus on safety and validity, including the crucial issue of causal identification when using observational data. We do not provide a specific model, but rather a way to integrate existing methods and know-how into a practical pipeline. We further provide a real world use-case of treatment optimization for patients with heart failure who develop acute kidney injury during hospitalization. The results suggest our pipeline can improve patient outcomes over the current treatment regime.

\end{abstract}

\section{Introduction}
\input{01_intoduction.tex}

\section{Preliminaries}\label{sec:background}
\input{02_cate.tex}

\section{Modeling Framework}\label{pipeline}
\input{03_cate_pipeline_steps}

\section{Case study: acute care}\label{acute}
\input{04_acute_results}

\section{Related work} 
\input{06_related_work}

\section{Discussion}\label{sec:discuss}
\input{07_Discussion.tex}

\newpage

\bibliographystyle{unsrtnat}
\bibliography{combined_refs.bib} 

\bigskip
\bigskip
\bigskip
\bigskip
\bigskip
\bigskip
\section*{Appendix}
\input{09_appendix}

\end{document}

%% file: 01_intoduction.tex
The rapid accumulation of patient health data is driving an increased interest in using the data to learn models which recommend patient-specific treatments \citep{bica2021fromchallenges, Powell2021TenStudy, meid2020usingdata, kent2018personalizedeffects, eini2022tell, Prosperi2020CausalHealthcare}. The goal is to use patient health data, such as electronic medical records (EMRs), to learn individualized treatment rules to improve patient care compared to existing treatment policies. Ideally, the learned individualized treatment rules will help physicians make better decisions about the best treatment for each patient. This idea holds a particular interest in scenarios where clinicians experience high uncertainty in treatment decision-making. For example, consider the case of acute heart failure (AHF) patients who have developed acute kidney injury (AKI). Caregivers face a challenging trade-off when deciding the optimal level of diuretics to administer, having to decide whether to target the congestion problem or the organ perfusion problem (intra-vascular volume) \citep{Boulos2019TreatmentInjury,damman2014renalmeta-analysis,tang2010cardiorenal}. 
The estimation of optimal patient-level treatments is a causal problem  \cite{Rubin1974EstimatingStudies, holland1987causalstudies, pearl2009causality}, as it requires evaluating and comparing the causal effect of each treatment under consideration. Thus, learning treatment recommendation models from patient health data requires extra care, both due to the biases inherent in causal data analysis \citep{hernan2011withepidemiology, shalit2019candata} and the high stakes of the task.  %\rom{maybe push later} 

However, despite their wide use, there are real concerns about the possible biases of observational data, chief among them the possibility of unobserved confounding between treatment and outcome leading to bias in the effect estimates. Furthermore, while observational health data is widely used to estimate average treatment effects (ATEs), much less work has been done on estimating patient-level effects using such data \cite{shi2022learning}. 

Indeed, estimating patient-level treatment effects is a daunting task -- we cannot hope to correctly estimate the true effect for each and every patient. However, we claim that in many cases estimating the accurate effect for every patient is not necessary for having a positive impact. 
Rather, in cases where there are no clear treatment guidelines, a reasonable goal for a treatment recommendation system would be to make recommendations that, if followed, would improve the overall average patient outcomes compared to the current practice \citep{shalit2019candata, fernandezloria2021causalmatters}. This idea is formalized in the notion of \emph{policy value} (see \cref{eval}).
Importantly, recommendations need not be given for all patients -- in order to ensure safety, recommendations may be deferred for patients for whom the recommendation model shows high uncertainty regarding which treatment is best. 

We propose the Target Recommendation System, a framework for safely learning and rigorously evaluating patient-level treatment recommendation models from patient health data.

Our goal is to enable researchers to responsibly develop treatment recommendation models that will lead to overall better patient outcomes. The framework includes suggested guidelines for assessing whether learning such models is feasible for a given clinical task and dataset, and best practices for applying and validating the many patient-level causal estimation methods currently available. 

Given a clinical question of interest (``How should we manage diuretics for heart failure patients with kidney injury?''), the goal of the proposed framework is to help practitioners address two questions: (1) Is the clinical question answerable given the available data? And, if positive, (2) How to safely estimate a treatment recommendation model from data, and how to evaluate its potential value?

Toward these goals, we first give a set of sufficient conditions regarding the clinical task and the available data. We propose guidelines for addressing the questions of hidden confounding, and more generally addressing the problem of \emph{causal identification} \citep{hernan2020causal}, i.e. the conditions under which using the available data for estimating the desired causal effects is possible in the first place. 
We then propose a workflow for learning and validating a treatment recommendation model, integrating a large selection of recent work in statistical machine learning and causal inference. Our focus here is not on any specific algorithm, but rather on how to best make use of the rapidly growing literature in the field. 
% We will supply the code base that supports our framework upon publication.
% \footnote{code available at: \url{https://github.com/RomGutman/robust_cate_policy}

We apply our framework to a real-world case study involving patients with a severe and difficult-to-treat condition: patients hospitalized with AHF, who developed AKI during their hospital stay. We estimate and validate treatment recommendation policies, showing that they could improve patient welfare compared to current practice.

%% file: 02_cate.tex
In this section, we give the basic definitions needed for developing the framework. A full formal introduction to causal inference is beyond the scope of this paper, and we refer the readers to \citep{hernan2020causal, pearl2009causality, chernozhukov2024applied, wager2024causal}
References on more specific concepts are given throughout the section.

\subsection{The potential outcomes framework and the conditional average treatment effect}\label{ci}
Our goal is to estimate the effect of a treatment, or intervention, $T$ on an outcome $Y$ given covariates $X$. We consider a binary treatment, although our framework can be generalized to multiple treatments.
We assume for each unit (i.e., patient) there exist two \emph{potential outcomes} \citep{Rubin1974EstimatingStudies}  $Y^1, Y^0$, where $Y^t$ is the outcome that would have occurred had the patient received treatment $t$.

Under the potential outcomes framework, the Average Treatment Effect (ATE) is defined as $
\mathbb{E}[Y^1 - Y^0]$. The ATE can be interpreted as the difference between the outcome of treating all the population with $T=1$ versus treating them all with $T=0$.
The Conditional Average Treatment Effect (CATE) \citep{Imbens2004NonparametricReview}
is defined as the treatment effect conditioned on $X=x$: 
\begin{equation}
    \label{eq:CATE_def}
    \tau(x) := \mathbb{E}[Y^1 - Y^0\mid X =x].
\end{equation}
The CATE represents the average gain or loss from changing the treatment for the sub-population with covariates $X=x$. In this paper, $X$ is assumed to be a relatively high-dimensional vector, and in practice is usually unique for each patient.

We further assume we observe a sample of $n$ patients $\{(x_i, t_i, y_i)\}_{i=1}^n$, where $x_i \in \mathcal{X} \subset \mathbb{R}^d$ represents the baseline (pre-treatment) covariates of the $i$-th patient, $t_i \in \{0,1\}$ is a treatment given in a single point in time, and $y_i \in \mathcal{Y}$ corresponds to a continuous or binary outcome of interest.

Only one of the potential outcomes $Y^t$ can ever be observed for each patient \cite{Rubin1974EstimatingStudies}, a major challenge known as ``The Fundamental Problem of Causal Inference'' \cite{holland1987causalstudies}. Thus, estimation of either of the above causal quantities using the observed outcomes $y_i$ requires that some set of causal identification conditions be satisfied, conditions which we outline in \cref{subsq:causal_conditions} below.
If these conditions are indeed satisfied, unbiased estimates of the ATE and CATE can be obtained from the patient data using estimation methods, as described in \cref{causal_estimate_methods}.

\subsection{Conditions for causal effect identifiability}\label{subsq:causal_conditions}
Treatment effects can be estimated from observational data under a set of four jointly sufficient conditions \citep{ROSENBAUM1983TheEffects, Rubin1980RandomizationComment}, commonly used in causal inference: \textbf{stable unit treatment value assumption (SUTVA)}~\citep{Rubin1980RandomizationComment}, \textbf{consistency} ~\citep{Rubin1980RandomizationComment}, \textbf{common support (overlap, positivity)} \citep{ROSENBAUM1983TheEffects} and \textbf{ignorability (conditional independence)}~ \citep{ROSENBAUM1983TheEffects}.
\begin{assumption} [SUTVA] \label{as:SUTVA}
    The potential outcome of any unit will not be affected by the treatment assignment of another unit, and, there are no different forms or versions of each treatment level, which lead to different potential outcomes.
\end{assumption}
\begin{assumption} [Consistency] \label{as:consis}
    $Y = Y^1 \cdot T + Y^{0}\cdot(1-T)$ \\ 
    The observed outcome $Y=y_i$ for each patient is in fact the potential outcome of the patient under the treatment $T=t_i$ that they had received.
\end{assumption}

\begin{assumption} [Common Support] \label{as:overlap}
    $p(T=t\mid X=x)>0 \quad \forall x \in \mathcal{X},t \in \{0,1\}$ \\ 
    All the patients have a non-zero probability of receiving each possible treatment $T$.
\end{assumption}
\begin{assumption} [Ignorability] \label{as:ignore} $\{Y^1,Y^0\}\bigCI T\mid X$ \\
This conditional independence statement implies that there are no unmeasured variables that affect both the treatment assignment and the outcome (a.k.a. \emph{confounders}).
\end{assumption}

While all causal effect estimation methods must rely on identification conditions, some of these conditions are provably untestable from data. Prominently, it is well-known that there is no data-dependent test for the validity of ignorability  
(\cref{as:ignore}) \citep{Cole2008ConstructingModels, pearl2009causality, hernan2020causal}. This condition is essential because hidden confounding can lead to non-vanishing bias in causal estimates, even with an unbounded sample size~\citep{pearl2009causality}. As a mitigation strategy, methods of \emph{sensitivity analysis} have been developed to test the stability of the treatment recommendations under varying levels of hidden confounding, including for CATE \citep{tan2006,yadlowsky2022bounds,kallus2019interval,jesson2021quantifying,jin2021sensitivity,yin2022conformal,oprescu2023b}. See \cref{subsec:uncert} for more details. 

The quantity in \cref{as:overlap} is commonly denoted by $e(x):=~p(T=t\mid X=x)$ and is known as the \emph{propensity score}. This is the treatment assignment probability for a patient with observed covariates $x$, as reflected in the observed data, i.e. it encodes the actual clinical practice. The propensity score is often estimated from data and used for causal identification and estimation \citep{austin2011anstudies}, and we discuss it in detail in \cref{propensity explained}.

\subsection{Methods for CATE estimation}\label{causal_estimate_methods}

Under the identification assumptions specified in \cref{subsq:causal_conditions}, the CATE function $\tau\left(x\right)$ can be estimated from  observed data as: $$\tau(x)=\mathbb{E}[Y\mid{T=1, x}] - \mathbb{E}[Y\mid{T=0, x}].$$
This is also known in the literature as heterogeneous treatment effect (HTE) (e.g., \citep{kent2018personalizedeffects}).
In recent years, many methods have been proposed to generate estimates $\widehat{\tau}$ of $\tau$. 

A commonly used set of approaches for estimating the CATE function is known as meta-learners \citep{Kunzel2019MetalearnersLearning}, where the CATE estimation is decomposed into several supervised learning subproblems. The simplest approach, named \emph{S-Learner} \cite{Kunzel2019MetalearnersLearning}, fits a single model $f(x,t)$ using the entire sample, with $T$ acting as a feature: ${f(x,t)}\approx \mathbb{E}[Y\mid{T=t, x}]$. CATE is then estimated as: $\widehat{\tau}(x_i) = f(x_i,1) - f(x_i,0).$ The so-called \emph{T-learner} \cite{Kunzel2019MetalearnersLearning} method fits two separate models $f_t(x)\approx \mathbb{E}[Y\mid{T=t, x}]$, each to the population receiving the corresponding treatment $t$, and estimate the CATE as $\widehat{\tau}(x_i) = f_1(x_i) - f_0(x_i).$
In both cases, the model(s) $f$ can be learned using any supervised learning model. As the S-Learner fits a single model for both treatment arms, the predicted outcomes of both arms share the same noise, thereby potentially reducing variance. However, the treatment covariate might get washed out and result in a zero-biased estimation of CATE. The T-learner approach offers more flexibility by employing separate models for each treatment arm, but it does not leverage the common components of the response, which can introduce both bias and unneeded variance, particularly when the sizes of the treatment groups differ.

Several works have expanded the meta-learners approach with the goal of overcoming some of the shortcomings of both the S- and T-Learners \citep{Kunzel2019MetalearnersLearning, nie2021quasi, kennedy2020towards}. These methods utilize the learner's estimation with additional knowledge, such as propensity scores, to establish what are known as pseudo-outcomes. For instance, the \emph{X-learner} \citep{Kunzel2019MetalearnersLearning} overcomes imbalanced treatment arm population sizes by establishing pseudo-outcomes using both arms of the T-learner with a weighting function. 
The \emph{R-learner} \citep{nie2021quasi} and \emph{DR-learner} \citep{kennedy2020towards} combine the propensity score with the outcome predictions to establish pseudo-labels which are in turn fitted using yet another model. See \citep{okasa2022meta} for a detailed overview of these methods.

Other approaches aim to estimate the CATE function directly. A well-known approach is the Causal Forest \citep{Wager2018EstimationForests}, a tree-based non-parametric method that directly models the treatment effect similar to the well-known random forest \citep{Breiman2001RandomForests}. Additional examples are causal boosting \citep{powers2018somedimensions}, the doubly-robust targeted maximum likelihood estimation (TMLE) \citep{Gruber2010AOutcome.,vanderLaan2015TargetedRule}, Gaussian process-based methods \citep{Cheng2019Patient-SpecificProcesses}, and deep learning-based approaches \citep{Shalit2017EstimatingAlgorithms,Johansson2018LearningDesigns, Louizos2017CausalModels, shi2019adaptingeffects}.
A comprehensive description of existing methods is provided by \citet{bica2021fromchallenges}.

\subsection{Causal and statistical uncertainty}\label{subsec:uncert}

In \cref{causal_estimate_methods} we described methods for estimating treatment effects. Building on these, incorporating uncertainty estimation is important for enhancing the reliability of these models as they work in tandem with clinicians. 

There are two primary sources of uncertainty in CATE estimates: statistical and causal. The statistical uncertainty, common in supervised learning, includes finite sample variance, inherent noise, and model misspecification, among others. This uncertainty has been extensively studied with both Bayesian methods \citep{Chipman2010BART:Trees, Hill2011BayesianInference, rasmussen2003gaussian} and frequentist approaches \citep{Wager2018EstimationForests, shafer2008tutorial, lei2021conformal}. The causal uncertainty stems from the reliance of CATE estimators on unverifiable assumptions, as mentioned above, notably the assumption of the absence of hidden confounders. This uncertainty is typically addressed by methods of sensitivity analysis which assume some bounded level of hidden confounding. Examples of such approaches tailored for CATE estimation can be found in \citep{tan2006, Yadlowsky2018BoundsFactors, kallus2019interval, jesson2021quantifying, jin2021sensitivity, yin2022conformal, oprescu2023b}. Some of these models account for both sources of uncertainty together \citep{jesson2020identifying, jesson2021quantifying, oprescu2023b}. Notably, causal uncertainty does not diminish with sample size, but might diminish by measuring previously unmeasured confounders.

Methods for estimating both types of uncertainty typically have predefined parameters that represent the allowed uncertainty levels. In this work we will denote these as \(\alpha_{\text{stat}}\) and \(\alpha_{\text{causal}}\), respectively. For example, the parameters can respectively represent the expected confidence level and assumed level of unobserved confounding.

These uncertainty parameters in turn yield a range of possible estimates for the CATE, establishing bounds for the possible values of $\tau(x)$. We denote these bounds as $\left[~\underline{\hat{\tau_{\theta}}}\left(x\right), \overline{\hat{\tau_{\theta}}}\left(x\right)~\right]$, where $\theta$ represents the set of used uncertainty parameters, for example $\theta =\left(\alpha_{\text{stat}}, \alpha_{\text{causal}}\right)$.

Importantly, when the range $\left[~\underline{\hat{\tau_{\theta}}}\left(x\right), \overline{\hat{\tau_{\theta}}}\left(x\right)~\right]$ includes both positive and negative values, it implies that it is uncertain whether the treatment is beneficial or harmful for patients with covariates $x$ for the defined uncertainty parameters $\theta$. As we will see below, in such cases we might want to defer making a treatment recommendation.

\subsection{Learning and evaluating treatment policies}\label{eval}
In this work, the main motivation for estimating the CATE function $\hat{\tau}$ is to inform a treatment policy for achieving optimal outcomes. Formally, let $\pi: X \rightarrow \mathcal{T}$ be a treatment policy that maps a patient's feature vector $x \in \mathcal{X}$ to a treatment assignment $t\in \mathcal{T}$. This is also known as an individual treatment rule (ITR). %The most straightforward way to obtain a policy from a CATE estimate is by choosing the treatment according to the sign of the CATE estimate; if higher outcomes are desirable, a reasonable policy would thus be $\pi(x)=\mathrm{I}_{\widehat{CATE}(x)>0}$.
Here we will consider the more general problem of learning a policy that either makes a treatment recommendation or alternately \emph{defers} the recommendation due to insufficient certainty \citep{jesson2020identifying}.

In this paper, we use CATE estimates to construct a policy. Given a CATE estimate $\hat{\tau}$, we apply a decision rule $\psi$ producing a policy $\pi$ such that $$\pi(x_i) = \psi\left(\tau\left(x_i\right)\right).$$ 
A standard approach for binary treatments is to assign treatment according to the sign of a CATE model $\hat{\tau}$, such that $\psi_{sign}(x_i) = \mathbbm{1}_{\hat{\tau}(x_i) \geq 0}$.

Given a policy $\pi$, we are interested in evaluating the expected patient outcomes if treatments were assigned according to $\pi$, and comparing these expected outcomes with outcomes under other treatment policies, e.g., the actual treatment assignment observed in the data. Generally, a good policy would be one that is better than current practice. The expected outcome under a certain policy is known as the \textit{policy value}, defined as: $$V\left(\pi\right):= \mathbb{E}\left[Y^{\pi(X)}\right].$$
Similarly to ATE and CATE, this estimand also suffers from the fundamental problem of causal inference, as for any given $x_i$, the quantity $Y^{\pi(X)}$ is a counterfactual whenever $\pi\left(x_i\right) \ne t_i$. Thus, the estimation of $V\left(\pi\right)$ requires causal identification conditions, and the same ones given in \cref{subsq:causal_conditions} are commonly used. More generally the problem of estimating $V\left(\pi\right)$ resembles that of estimating ATE, and similar approaches are often employed in this context. Most commonly, these methods rely either on outcome modeling (``plug-in'', ``direct methods''), weighted methods (inverse propensity weighting/scoring -- IPW)\citep{Qian2011PerformanceRules}, or a combination of both in a doubly-robust manner \citep{dudik2014doubly, montoya2022estimators}.

While below we mostly focus on learning a treatment policy via the CATE function, there are many approaches for optimizing $\pi(x)$ directly from data, with the aim of learning a policy attaining high values of $V(\pi)$ \citep{Qian2011PerformanceRules, bertsimas2019personalizedapproach, dudik2014doubly, Athey2017EfficientLearning, athey2021policy, Kallus2018BalancedLearning, montoya2022optimal}. Some approaches focus on learning a more complex decision rule $\psi$ \citep{athey2021policy}, combining propensity information with outcome models in a doubly-robust manner \citep{athey2021policy, Kallus2018BalancedLearning} or directly selecting the policy based on maximizing the potential outcome \citep{bertsimas2019personalizedapproach}. 
Several works have expanded this direction using semi-parametric methods \citep{montoya2022optimal,athey2021policy} or using other optimization methods \citep{Kallus2018BalancedLearning}.

%% file: 03_cate_pipeline_steps.tex
\begin{figure}[ht]
    \centering
    \includegraphics[width=.5\linewidth]{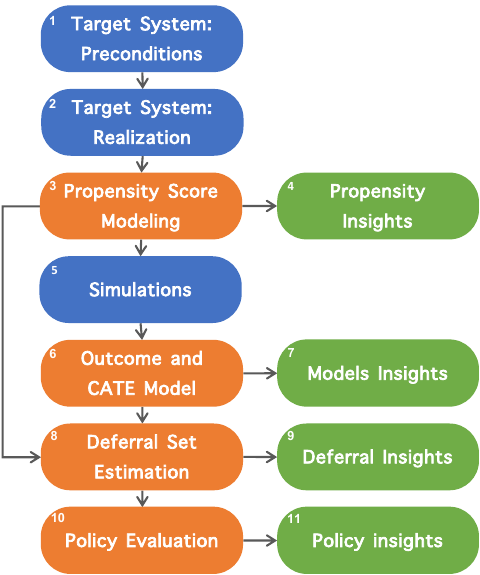}
    \caption{The outline of the target system, including \textcolor{pipe_blue}{\textbf{identification}},  \textcolor{pipe_orange}{\textbf{estimation}} and \textcolor{pipe_green}{\textbf{validation}} steps. Steps \textcolor{pipe_blue}{\textbf{1-2}} refer to defining the clinical and causal question of interests and step \textcolor{pipe_blue}{\textbf{5}} is designed to validate its feasibility. Steps \textcolor{pipe_orange}{\textbf{3,6,8,10}} are for estimating the causal quantities of interest such as the CATE, and establishing a recommendation policy (step \textcolor{pipe_orange}{\textbf{10}}). Steps \textcolor{pipe_green}{\textbf{4,7,9,11}} are for validating and evaluating the corresponding estimation steps,  where we recommend best practices to ensure each step's validity.} 
    \label{fig:pipeline}
\end{figure}
\label{target_trial_system}

We introduce a framework for developing a treatment recommendation system based on patient health data. Given patient covariates $x$, such a system either outputs a recommended treatment $t \in \mathcal{T}$, or a ``defer''  option ($\perp$), meaning that it has no recommendation for the specific case.
\cref{fig:pipeline} provides a high-level map of the framework's steps, where we distinguish between three types of steps:

\begin{itemize}
    \item \textcolor{pipe_blue}{\textbf{Identification:}} The identification steps are designed to help practitioners specify the desired treatment recommendation system and assess the feasibility of constructing it with the available data. This entails a close discussion with the clinical partners to define the problem setting, followed by a preliminary examination of the framework using a semi-synthetic simulation. 
    \item \textcolor{pipe_orange}{\textbf{Estimation:}} The estimating are meant to guide practitioners on how to build the target system models, including methods for estimating propensity scores, outcome and CATE models, and estimating the policy value.
    \item \textcolor{pipe_green}{\textbf{Validation:}} The validation steps are aimed at building confidence in the system’s output and gaining clinical insights. This includes proposed evaluation practices to assess the outcomes of the estimation steps. 
\end{itemize}
The following subsections provide a detailed description of the framework.

\subsection{\textcolor{pipe_blue}{Target Recommendation System: Preconditions}}\label{subsec:preconditions}
The ``Target Trial'' is a framework for estimating average treatment effects from observational data, which has had many successes in recent years \citep{rubin2004teaching, hernan2011withepidemiology,Hernan2016UsingAvailable, dickerman2019avoidable, Powell2021TenStudy}. The framework postulates that one should formulate an ideal clinical trial which would be used to estimate the average effect of the intervention in question and that the analysis of the observational data should emulate this ideal trial. Here, we propose a similar idea for the clinical decision support, framed as the ``Target Recommendation System'' (TRS). TRS helps define the estimands, when will the system be called upon, with what input, and how will it be integrated within the clinical workflow. 

Therefore, we start by formulating the basic questions needed to define the system. These should be discussed between the modeling and clinical experts on the team, including a close understanding of the clinical staff's routines, to gain firsthand insights into their decision-making processes when faced with treatment choices. Ideally, the discussions should take place before one embarks on collecting and analyzing patient data \cite{sendak2020adelivery}, as they bear on the causal assumptions needed for obtaining valid estimates from observational data (\cref{subsq:causal_conditions}), as well as the general applicability of the proposed system. The basic points to address are as follows:

\textbf{Single treatment decision at a well-defined time-point.} The TRS should focus on a single clinical decision which occurs at a well-defined time point in the clinical workflow. Examples include selecting among a set of possible medications, determining whether to embark on a pre-specified procedure (e.g., a surgery), whether to take a diagnostic action (e.g., a specific type of biopsy). This also includes preliminary inclusion and exclusion criteria, giving a preliminary definition of the target population for which the decision and the system are relevant.

The importance of a well-specified decision time-point (called time-zero by \citet{Hernan2016UsingAvailable}) is double. First, this will likely be the point in the clinical workflow where the system will be called upon to provide a recommendation. Thus, any input to the TRS must be based on data that is available before the decision time-point. 
Additionally, pre-specifying the decision time-point helps us grapple with questions of confounding: since confounders are factors that affect the clinical decision (and the outcome), understanding what information was available to the decision maker at decision time-point allows us to narrow down the set of potential confounders, addressing the ignorability assumption (\cref{subsq:causal_conditions}).

\textbf{Small action space.} We focus on treatment decisions with a relatively small set of treatment options (actions). Furthermore, these actions should in principle be applicable to most patients within the target population. Large action spaces usually imply a small sample size for at least some of the actions, and are much more likely to lead to violations of the common support assumption (\cref{subsq:causal_conditions}).

\textbf{Ambiguity or lack of clinical guidelines.} If the decision about the assignment of the treatment in question follows clear clinical guidelines, then it is very hard to generate evidence for the treatment recommendation system, as the overlap assumption (\cref{subsq:causal_conditions}) is likely violated: Patients of certain characteristics would (nearly) always receive the treatment specified by the guidelines, and there will be no data about the response of these patients to the treatment which goes against the guidelines. 
On the other hand, the absence of clear clinical guidelines grants clinicians flexibility in selecting a treatment, potentially leading to variability between clinicians, which in turn could lead to similar patients receiving different treatments.

Moreover, cases where there are no clinical guidelines are often the cases where clinicians are open to, and in need of, assistance in determining the best treatment for a given patient.

\textbf{Well-defined and meaningful outcomes.} The treatment decision should have a clinical outcome that is well-defined and acknowledged by the domain experts as clinically meaningful when deciding on treatment. This is the outcome which the TRS will aim to optimize.\\

Assuming the above preconditions have been discussed and affirmatively addressed, the output of this stage should be the following: (1) \textbf{who are the clinical practitioners} expected to use the system; (2) \textbf{who are the patients} for which the recommendation is needed (3) the \textbf{time point} within the clinical course when a recommendation will either be solicited by the clinician or offered by the system; (4) what information do we expect to be available as \textbf{input} to the system at this time point; (5) the exact types of recommendations the system is expected to present to the clinical staff, i.e. the system's \textbf{outputs}. 

In cases where the above conditions are not met, one can try and redefine the treatment recommendation system. This includes, but is not limited to, redefining the treatment decision, restricting the recommendation to a smaller sets of treatments, or redefining the target population. One might also conclude that the task is unfeasible, which we refer to as an \emph{exit point}. 

In \cref{sec:clinical-questions} we present suggested questions that can aid in defining exactly the setting and context of the treatment recommendation system.

\begin{figure}[ht]
    \centering
    \includegraphics[width=.5\linewidth]{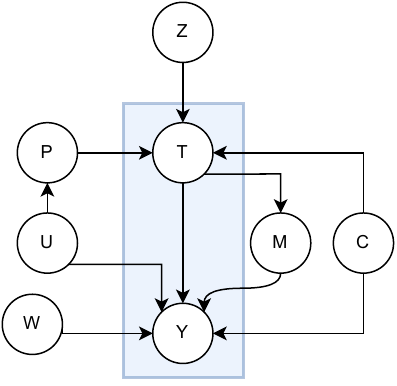}
    \caption{A schematic illustration of the different types of causal variables for the question of the causal effect of the treatment $T$ on the outcome $Y$; this is a modified version of Figure 1 from \citet{tennant2021use}. $Z$ is an instrumental variable, $M$ is a mediating variable, $W$ is an effect modifier, $C$ is a \emph{measured} confounder, $U$ is an \emph{unmeasured} confounder, and $P$ is a proxy observation of some part of $U$ (e.g., a blood test might be a proxy for the state of a physiological system). }
    \label{fig:causal_variables}
\end{figure}

\subsection{\textcolor{pipe_blue}{Target Recommendation System: Realization}}\label{subsec:ident-data}

This stage aims to assess whether the assumptions for causal effect identifiability discussed in \cref{subsq:causal_conditions} hold in the available data.

We begin by evaluating the SUTVA condition (\cref{as:SUTVA}). This involves confirming that treatments are abundantly available, preventing a scenario where one patient's treatment assignment compromises another patient's access to treatment. In addition, treatments for contagious 
diseases might also violate SUTVA, as treating one patient might cause or prevent other patients from falling ill.
Also, treatments that are nominally considered the same should in fact agree across different settings in either their mechanism of effect or their medical protocols; for example, a procedure must not be done in substantially different ways across two hospital units participating in the same study. 

For the consistency assumption (\cref{as:consis}) to hold, we must ensure that treatment allocations are accurately recorded in the dataset.

We discuss the common support condition (\cref{as:overlap}) in detail in \cref{propensity explained}. 

In the rest of this subsection, we focus on the ignorability assumption (\cref{as:ignore}). Notably, we are dealing here with potential confounders for actions taken by trained human decision-makers. Since humans typically take into account a limited number of factors in their decision-making process \citep{simon1955behavioral, kushniruk2001analysis, patel2002emerging}, this reduces the possible scope of hidden confounders, although non-conscious factors should be taken into account as well.

We propose two alternative approaches to evaluating whether the ignorability assumption holds. The first is to build a causal graph \citep{pearl2009causality} of all the covariates in the data and in addition any other factors that might influence the decision making process of the clinicians. This should be done in close consultation with the clinical experts. Notably, the causal graph could include variables that are not represented as-is in the dataset.
The causal graph allows researchers to 
map different variables and their relationships, and potentially use the \textit{backdoor criterion} \citep{pearl1995causal}, a widely used condition for causal identifiability. \citet{tennant2021use} offer a practical guideline on how to include and report such graphs.

Following \citet{tennant2021use}, in \cref{fig:causal_variables} we provide a schematic illustration of the different types of variables and how they affect each other. A simplified approach to the analysis is to make sure that all confounders ($C$) are represented in the data, or, when that is impossible, that relevant proxies ($P$) exist for confounders that are not in the data. For example, one can use hematocrit and hemoglobin biomarkers as proxies for congestion in AHF patients \citep{duarte2015prognostic}.
In addition, including effect modifiers ($W$) is beneficial in terms of reducing variance for outcome modeling. On the other hand, including instrumental variables ($Z$) in the analysis could increase both bias and variances and should be avoided \citep{ding2017instrumental, sauer2013covariate, huenermund2022choice}. Mediating variables ($M$) occur after the treatment and thus should also be discarded.

However, we note that constructing a causal graph that comprehensively captures the multitude of potential variables involved in a clinical problem is often challenging and time-consuming. Thus, in \cref{sec:variable_selection} we provide an alternative to the causal graph approach: a step-by-step protocol addressing the question of confounding and which covariates should be included and excluded from the analysis.

As in the last step (\cref{subsec:preconditions}), in cases where the data ultimately does not meet the causal identification requirements, the researcher should re-consider the settings, data acquisition setup, and the research question itself.

\subsection{\textcolor{pipe_orange}{Propensity Score Modeling}}\label{propensity explained}
In this step, we focus on the estimation of propensity score $e(x)$, as defined in \cref{subsq:causal_conditions}.
The purpose of estimating it in our framework is twofold. First, to evaluate the population for which the common support assumption holds (\cref{subsq:causal_conditions}), and for whom we can plausibly estimate the CATE function. Second, propensity scores are used both in some CATE estimators (e.g. X-learner \citep{Kunzel2019MetalearnersLearning} and R-learner \citep{nie2021quasi}) and in policy value estimators and baselines, as we will see in \cref{subsec:polval}. We elaborate on the use of propensity score as an overlap evaluator in the validation step presented in \cref{propensity_insights}.

From a modeling perspective, estimating the propensity score is a relatively straightforward task of estimating the conditional probability of the binary variable $T$ conditioned on a covariate vector $X$. We recommend using and comparing several models for modeling the propensity score, such as (regularized) logistic regression, XGBoost \citep{Chen2016XGBoost}, and outcome adaptive Lasso \citep{Shortreed2017}. 

As the propensity score should ideally represent the true probability of treatment conditioned on the confounders, in-sample and out-of-sample calibration is an important metric for propensity model evaluation \cite{gutman2022propensity}. Calibration can be improved for any given model by using post-processing methods \citep{platt1999probabilistic, zadrozny2001obtaining}. 

We note that standard discrimination metrics for classifiers such as accuracy or the area under the receiver-operator characteristics curve (AUROC) should be treated here more as descriptive measures of the propensity score. For instance, when a model achieves an AUROC of $1$, it signifies that the model assigns a higher score ($\hat{p}(T=1 \mid X=x)$) to each unit $x$ with treatment $T=1$ compared to units with treatment $T=0$. Although this behavior can be desirable in standard classification problems and would indicate a strong predictor, in the context of propensity estimation such an AUROC score suggests a possible lack of overlap.

\begin{figure}[ht]
    \centering
    \includegraphics[width=.8\linewidth]{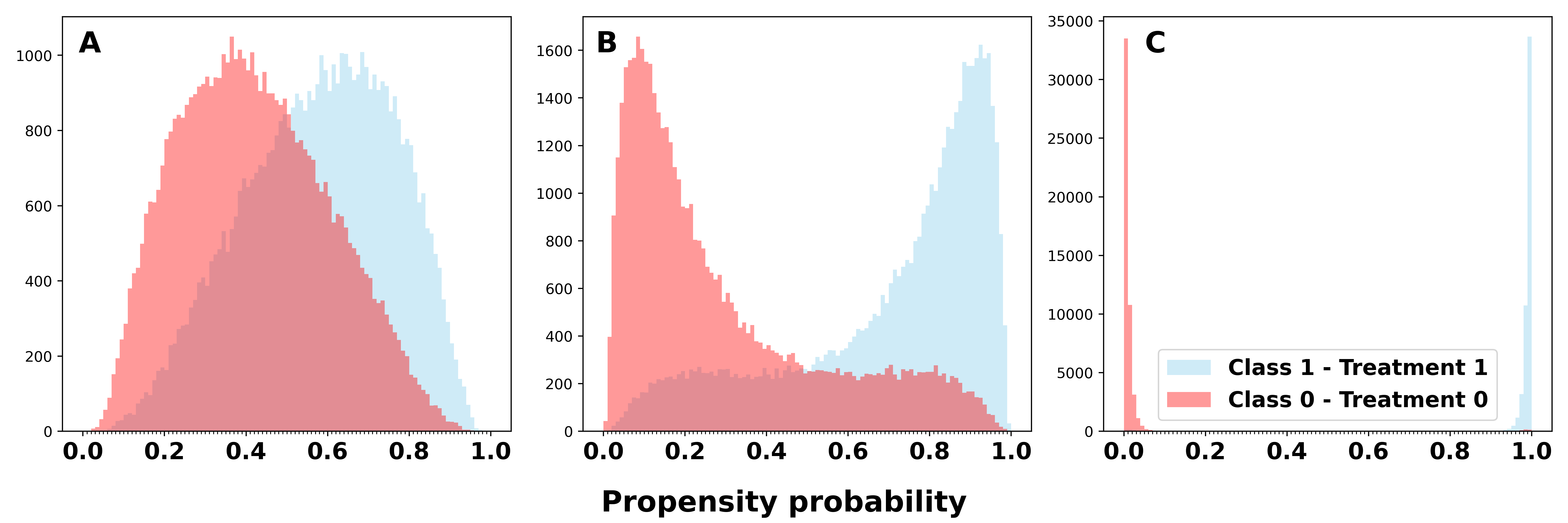}
    \caption{Schematic illustration of propensity overlap, presenting the distribution of propensity scores (x-axis) of patients received treatment $T=1$ (blue) or treatment $T=0$ (red). 
    \textbf{A} Strong overlap, where minor trimming could be considered.
    \textbf{B} Mild overlap, where trimming for extreme values is advised.\\
    \textbf{C} Non-overlap case, where the researchers should re-consider the research question (i.e. using this data as-is is not advised).}
    \label{fig:prop_demo}
\end{figure}
\subsection{\textcolor{pipe_green}{Propensity score model insights}}\label{propensity_insights}

After fitting and evaluating the propensity score models, the researcher should validate their clinical face validity and use them to evaluate whether common support (\cref{as:overlap}) holds for a substantial sub-population.

As a clinical face validity check, we recommend evaluating the models using standard interpretability tools such as examining the weights of linear models, or using SHAP values \citep{lundberg2017unified}. The results can then be evaluated with the clinical team. The idea is to ensure that the important factors for clinical decision-making are reflected in the propensity score model. If concerns are raised (e.g. the model indicates some variable is highly predictive of treatment, but the clinicians find this unreasonable based on their experience), further analysis of the models and data is required. 

If the models are deemed clinically reasonable, our focus shifts to a further objective of the propensity score model, namely assessing the extent of common support: to what degree could each unit plausibly receive each of the treatments. Towards that end, we recommend visualizing the distribution of propensity scores across the treatment arms. \cref{fig:prop_demo} presents a stylized demonstration of different levels of overlap: \cref{fig:prop_demo}A exhibits near-perfect common support, \cref{fig:prop_demo}B shows good common support, while \cref{fig:prop_demo}C demonstrates no common support. The first two cases illustrate adequate representation for each treatment arm, although trimming might be necessary, as we discuss in the next paragraph. However, the last case (\cref{fig:prop_demo}C) raises a warning sign, indicating a complete lack of common support. In such instances, the research objectives should be reevaluated, as there is likely no reliable statistical method for estimating meaningful treatment recommendations.

Even if common support is satisfied for most patients (as in \cref{fig:prop_demo}A and \cref{fig:prop_demo}B), some patients might still be outside the common support region. As part of our framework, we propose that recommendations for patients such as these, i.e. those with extreme propensity scores, be \emph{deferred}. The reasons are twofold: First, patients with extreme propensity scores are those who typically receive only one type of treatment, and therefore we cannot reliably estimate their counterfactual outcomes \citep{dehejia1999causal,fogarty2016discrete, hernan2020causal, shi2019adaptingeffects}. Furthermore, patients with extreme propensity scores are typically those for whom clinicians do not require recommendations from a data-driven model, as they are consistently treated with a single treatment.

Deferral by propensity score is performed by settings lower and upper bounds (denoted $\eta_l$ and $\eta_h$, respectively), such that the overlap set is defined as:
\begin{align}\label{eq:Soverlap}
    S_{\text{overlap}} := \left\{x \in X \mid \eta_l \leq \hat{e}\left(x\right) \leq \eta_h \right\}.
\end{align}

When determining the thresholds of the overlap region ($\eta_l$ and $\eta_h$), factors such as the feature space dimensionality and the number of patients under each treatment arm in the low overlap regions should be taken into consideration \citep{crump2009dealing, d2021overlap,fogarty2016discrete}. 
The goal is for the downstream CATE models to have sufficient support for estimating counterfactuals.

An additional validity check is to characterize the above populations for which common support does not hold, i.e. patients who according to the propensity model are treated nearly deterministically with a single treatment option. We propose consulting with the clinical partners whether in their experience treatment for this population is indeed chosen unambiguously. The goal is to further test whether the model makes clinical sense and to highlight possible omissions in the data or model.

We further propose testing whether vulnerable populations tend to have more extreme, or less extreme, propensity score values compared to the rest of the population, indicating possible systemic biases in the clinical population. Such biases should be highlighted and examined down the pipeline, as a possible target for improved recommendations.

Finally, we note that the notion of deferral is related to, but distinct from, what is commonly known as \emph{trimming} \citep{crump2009dealing, d2021overlap} in the observational study literature: the practice of removing from the study units with extreme propensity scores. When estimating an ATE, trimming raises issues regarding the target population \citep{crump2009dealing, d2021overlap}, as the ATE is only estimated for the non-trimmed subpopulation. However, in our case, we do not suffer from this problem, as we are aiming for individual recommendations rather than ATE estimates.

\subsubsection{Sub-population characterization}\label{subsec:subpop_char}

Throughout this framework, we come up with a need to characterize in an interpretable way certain patient sub-populations. For example, here we wish to characterize the patients outside of the overlap (common support) zone, i.e. those with extreme propensity scores. We thus give some general recommendations for the task of characterizing a sub-population. 

We first recommend generating some descriptive statistics of the patients included and excluded from the sub-population, and presenting them in simple tables or histograms. 

We then recommend fitting a linear model or a single decision-tree $f: X \rightarrow Z$, where $Z$ is the sub-population label (e.g. ``deferred'' and ``not deferred''), and visualizing the top coefficients that contribute the most to the model's classification. For a linear model, this can be measured by absolute coefficient value. One can further use SHAP values visualized over linear or non-linear models \citep{lundberg2017unified}.

\subsection{\textcolor{pipe_blue}{Semi-synthetic outcome simulation}}\label{simulation_step}
    
The previous steps are meant to establish whether the data is suitable for the task, mainly from a causal perspective. In this step, the goal is to gain a better understanding of whether the data suffices from the statistical perspective, i.e. in terms of sample size and the ability to fit machine learning models which are adequate to the task.

Toward that end, we propose using a semi-synthetic outcome simulation. The simulation serves two main purposes: (1) Debugging and testing whether, under relatively favorable conditions, the proposed framework can use data to uncover a treatment policy that is reasonably close to the optimal one. (2) Estimating whether the sample size is sufficient to discover treatment effects and plausibly estimate a policy that is superior to current practice, if such a policy exists.

We propose creating a simulation based on the existing covariates $X$ and observed treatment assignments $T$, where the only simulated component are the potential outcomes for each unit, $(Y^1_i,Y^0_i)$. Thus, one can obtain the ``true'' treatment effect (within the context of the simulation) for each unit by computing $Y^1_i-Y^0_i$. Once the potential outcomes are simulated, one can estimate the optimal policy $\pi^*$ and its optimal policy value $V(\pi^*)$ under the simulation, as well as the policy value of the current policy, $\pi_{\text{current}}(x_i) = t_i$, and the policy value of any other policy of interest.

We propose using several different functional and parametric forms for the simulated outcomes, e.g. linear, tree-based etc. The simulated potential outcomes should meet the following criteria: (1) they should have the same scale as the true outcomes, (2) exhibit similar variance as the true outcomes, (3) possess a clinically reasonable distribution of CATE values, and (4) allow for explicitly controlling varying levels of divergence between current practice and optimal practice.

Next we follow the \textcolor{pipe_orange}{estimation} phase steps of the framework using the simulated data:
The first of these steps is estimating CATE models (denoted $\mathcal{A}$) based on the simulated data (see \cref{subseq:Fit multiple CATE}), derive a corresponding policy $\pi_\mathcal{A}$ (\cref{eval}), and estimate its policy value $\hat{V}(\pi_\mathcal{A})$ (see \cref{subsec:polval}).

We recommend running the simulation multiple times with varying settings and random seeds. The analysis should then focus on the degree to which policy value estimators (\cref{subsec:polval}) agree with the ground-truth policy values, using measures such as Pearson correlation, scatter plots etc., and the degree to which CATE models managed to lead to a beneficial policy, by comparing their policy values to policy values of baseline policies (\cref{policy_insights}) on the one hand and the optimal policy on the other hand. Most importantly, we recommend checking (1) whether the estimated policy value for the policy $\pi_\mathcal{A}$ is close to its true policy value under the simulation, i.e. is $\hat{V}(\pi_\mathcal{A}) \approx V(\pi_\mathcal{A})$; (2) whether the learned policy $\pi_\mathcal{A}$ has a better policy value than the current treatment policy, i.e. is $V(\pi_\mathcal{A}) > V(\pi_{current})$; and (3) whether the policy value is reasonably close to the optimal policy, $\hat{V}(\pi_\mathcal{A}) \approx V(\pi^{*})$.

In \cref{appendix_simulation} we provide the simulation we created for our case study with detailed explanations of how we aimed to achieve the simulation goals.

\subsection{\textcolor{pipe_orange}{Outcome and CATE modeling}} \label{subseq:Fit multiple CATE}

Having established an overlap set (\cref{eq:Soverlap}), the next step is estimating the CATE function for members of the set $S_{\text{overlap}}$. The field of CATE estimation has seen significant growth in recent years, with numerous models and methods being developed and actively researched, as discussed in \cref{causal_estimate_methods}. In this step, our focus is on addressing the major difficulties of CATE estimation, rather than advocating for any specific method.

A major challenge in CATE estimation is the absence of a held-out set, and more generally the difficulty of evaluating and comparing the accuracy of CATE estimates. This is unlike regression or classification models, where held-out error can be used for evaluation and model selection. Thus, practitioners might be at loss with regard to which of the many possible options for CATE modelling should they use \citep{curth2023magic}.

To address this challenge, we propose a two-pronged approach that can be applied to any of the aforementioned methods.
First, considering that most CATE models involve regression or classification models as components (e.g. all the meta-learner approaches), we recommend evaluating these ``component models'' in a manner consistent with the evaluation of traditional models. Although such evaluations are not sufficient for selecting between different families of CATE meta-learners (e.g., T-learner vs. X-learner), they can still be utilized to compare and reject poor-performing models within a given family. For instance, if we intend to use a random forest regression model as a component in a meta-learner, and we find that its held-out mean squared error is exceedingly high (for example it is approximately equal to the overall variance of the regression target), it is advisable to exclude this model from the CATE estimation process. Having said this, it is important to note that a CATE model may use an underlying regression or classification model that demonstrates superior performance in terms of held-out accuracy, yet still leads to inferior accuracy in CATE estimation and the subsequent derivation of a treatment policy \citep{curth2021really}. Thus, the goal of this first step is merely to reject the worst-performing models, not to choose the best one.

Second, once distinctly under-performing models have been weeded out, we suggest using \emph{held-out policy value} as the primary metric for model evaluation and selection. While not a perfect metric, it serves as the closest proxy to the ultimate goal of the treatment recommendation model, which is to improve patient welfare. We discuss the policy value in detail in \cref{policy_insights} below.

\subsection{\textcolor{pipe_green}{Outcome and CATE model insights}}\label{subsq:cate_insights}

For CATE models that have reasonably performing components and competitive policy value performance, we propose a series of validity tests.

First, we recommend conducting an error analysis of the component regression or classification models.
This analysis aims to identify the strengths and weaknesses of the component models, detect sub-populations where the models might perform poorly, and highlight significant discrepancies among them. These insights, to some extent, can also feed back into the previous step and guide the selection of component models.

We further advise employing standard interpretability methods for the component models and involving clinical experts in examining the findings. If the models heavily rely on covariates that appear to be incongruous, it is important to investigate this phenomenon as it could indicate fragility in the models. However, it is worth noting that highly predictive models may rely on administrative data, such as the timing of tests, and these should not be dismissed outright \citep{agniel2018biases, Gutman2022WhatOutcome}.

We then suggest estimating standard correlation metrics, such as Pearson and Spearman correlation, between the estimates of the selected CATE models. Strong agreement among models with different parametric assumptions provides more confidence in their results. On the other hand, if the correlations between models are low it may indicate the presence of unobserved confounding or suggest that each model explains different aspects of the treatment effect variance, with no clear determination of which is accurate. Therefore, a low correlation between the outputs of different CATE models serves as a warning sign regarding the validity of the estimates.

Next, we propose visualizing the CATE calibration graph as outlined in \citep{athey2019estimating} to illustrate the data heterogeneity.
This method divides the dataset into segments based on quantiles of the estimated CATE values. In each segment, we calculate the ATE using a method such as Augmented Inverse Probability Weighting (AIPW) \citep{aipw} or some other ATE estimation technique. Then, we plot the estimated ATE within each segment versus the average predicted CATE within this segment. This may indicate how well the predictor identified homogeneous groups in the heterogeneous population, by evaluating how aligned those two estimates are.

Finally, the CATE estimates should be used to infer the ATE and compare it to any known effects from the existing literature, if such exist. A wide divergence from any existing ATE estimates is not necessarily indicative of a mistake, but should encourage a deeper dive into potentially substantial differences between the setting and population used in the study and the settings and populations used in previous studies. 

\subsection{\textcolor{pipe_orange}{Deferral set estimation}}\label{subsec:deferral}

As we focus on high-stakes clinical recommendations, we wish to create models that refrain from unfounded recommendations. 
This requires assessing the uncertainty of CATE estimates and deferring decisions where this uncertainty is high.

As discussed in \cref{subsec:uncert}, we recall that for CATE models there exist both statistical and causal uncertainty. We propose estimating both sources of uncertainty jointly (e.g. using the method proposed by \citet{jesson2021quantifying, yadlowsky2022bounds,oprescu2023b}), and use the estimated uncertainties to establish a deferral rule $\text{Rej}$ parameterized by uncertainty parameters $\theta$:
\begin{align}
    \text{Rej}_{\theta}: \mathcal{X} \rightarrow \left\{0, 1 \right\},
\end{align} 
where $\text{Rej}_{\theta}(x)=1$ implies that, given uncertainty parameters $\theta$, for a patient with features $X=x$ we defer making a treatment recommendation. A common rule, as detailed in \cref{subsec:uncert}, is to defer the decision for a sample 
if its joint uncertainty interval includes zero, i.e.$ 0 \in \left[~\underline{\hat{\tau}_\theta}\left(x\right), \overline{\hat{\tau}_\theta}\left(x\right)~\right]$. 

For each of the methods jointly modeling statistical and causal uncertainty the user must specify the uncertainty parameters; the statistical parameter is typically a prediction interval or a Bayesian credible interval, while the causal parameter is often a bound on the level of divergence between the true and observed propensity scores \citep{tan2006}.
Importantly, we note that we are not restricted to standard prediction interval levels such as setting $\alpha_{\text{stat}} = 0.95$. We might be willing to tolerate more uncertainty than that in order to defer fewer recommendations. We encourage varying the degrees of accepted statistical and causal uncertainties ($\alpha_{\text{stat}}$, $\alpha_{\text{causal}}$) to understand their impact on the proportion of deferrals across the population, and on the held-out policy value of the non-deferred patients.

Importantly, the final deferral rule $\text{Rej'}_{\theta}$ should also include the patients who were ``trimmed'' during the propensity evaluation (i.e., $x_i \notin S_{\text{overlap}}$ (\cref{propensity_insights})). Therefore, we recommend a combined deferral rule. For example, given a propensity estimator $\hat{e}\left(x\right)$ and using uncertainty intervals with uncertainty parameter $\theta$, we obtain a deferral rule:
\begin{align}
    \text{Rej'}_{\theta}\left(x\right)= 
\begin{cases}
    1, & \text{if } \eta_l > \hat{e}\left(x\right) \lor \hat{e}\left(x\right) > \eta_h \\
    1 & \text{if } 0 \in \left[~\underline{\hat{\tau}_\theta}\left(x\right), \overline{\hat{\tau}_\theta}\left(x\right)~\right]\\
    0 & \text{otherwise}
\end{cases}.
\end{align}

\subsection{\textcolor{pipe_green}{Deferral set insights}}\label{subsec:defer_insight}

Once a deferral rule $\text{Rej'}_{\theta}$ has been determined, we propose examining and characterizing the deferred patient population, understanding which populations tend to be deferred more than others. A special emphasis should be put on the role of vulnerable populations who might end up being under-served or over-served by the recommendation system. 

To this end, we recommend using similar methods as described in \cref{subsec:subpop_char} and consulting with the domain experts whether the characterization of the deferred sub-population makes clinical sense.

\subsection{\textcolor{pipe_orange}{Policy}}\label{subsec:polval}

The main purpose of our proposed framework is to generate a \emph{policy}: a recommendation for the treatment of eligible patients. These recommendations are generated based on a model's CATE estimate, coupled with a decision rule converting the CATE estimate to a recommendation.
Formally, given a CATE estimation method $\mathcal{A}$ yielding CATE estimates $\widehat{\tau}_\mathcal{A}$ (\cref{subseq:Fit multiple CATE}), a deferral rule $\text{Rej'}_{\theta}$ (\cref{subsec:deferral}), and a decision rule $\psi : \mathbb{R} \rightarrow \mathcal{T}$, the policy for a patient with features $x$ is defined as:
\[
    \pi_\mathcal{A}\left(x\right)= 
\begin{cases}
    \perp & \text{if } \text{Rej'}_{\theta}(x) = 1 \\
    \psi\left(\widehat{\tau}_\mathcal{A}(x)\right) & \text{otherwise.}
\end{cases}
\]

\noindent As mentioned in \cref{eval}, there are several possible choices for the decision rule $\psi$, where the most commonly used is simply the sign of $\hat{\tau}$: $\psi(\hat{\tau}_\mathcal{A}\left(x\right) ) = \mathbbm{I}_{\left\{\hat{\tau}_\mathcal{A}\left(x\right) \geq 0\right\}}$ (or some threshold other than 0, depending on the context). 

The quality of a policy is assessed via the policy value $V\left(\pi\right)$ (\cref{eval}), evaluated on the held-out data. For policies with a defer option such as the ones we propose, the policy value estimate has two components. When the policy defers, we assume the action and its outcome would match the policy observed historically in the data (i.e. current assigned treatment); thus, for a unit $x_i$ with observed treatment $t_i$ and outcomes $y_i$ we assume the outcome under ``defer'' would simply be $y_i$, i.e. we assume the clinicians would treat this patient as they always do. For emphasis we denote this outcome $Y_{\text{factual}}$. When the policy does not defer and gives a treatment recommendation, we estimate the policy value using standard methods. Formally, given a policy function $\pi: \mathbb{X} \rightarrow \left\{0,1,\perp\right\}$, where $\perp$ is deferral, the policy value of $\pi$ is given by: 
\begin{equation}\label{eq:policy_val}
    V\left(\pi\right) = \mathbb{E}\left[Y\left(\pi\left(x\right)\right) \mid \pi(x) \ne \perp \right]p\left(\pi \left(x\right) \ne \perp \right) + \mathbb{E}\left[Y_{\text{factual}} \mid \pi(x) = \perp \right]p\left(\pi \left(x\right) = \perp \right),
\end{equation}
where $\mathbb{E}\left[Y\left(\pi\left(x\right)\right) \mid \pi(x) \ne \perp \right]$ describe what will be the average outcome $Y$ using policy $\pi$.
As this quantity is a causal quantity which requires estimating unobserved potential outcomes, causal identifying assumptions are required to hold for this quantity to be estimated. In \cref{eval} we discuss some of the methods that have been suggested to estimate policy value \citep{dudik2014doubly, Athey2017EfficientLearning, athey2021policy, Kallus2018BalancedLearning, leete2019balanceddata, imai2023experimental}, where common approaches include \emph{Inverse Propensity Weighting} (IPW) as well as \emph{Doubly-Robust} (DR) approaches which combine the propensity and outcome models \citep{dudik2014doubly}. We denote them as $V^{IPW}$ and $V^{DR}$, respectively. All of these should be used on held-out data, since the policy is learned and thus naive estimates might suffer from overfitting. 
In this work we suggest using the DR methods as given by \citep{dudik2014doubly,lunceford2004stratificationstudy}, with bootstrap sampling. See \cref{policy_val_eq_appendix} for details.

\subsection{\textcolor{pipe_green}{Provide insights on chosen policy}}\label{policy_insights}

Once several policies $\pi$ (derived from different CATE estimators, rejection rules, etc.) and their corresponding estimated policy values $\hat{V}\left(\pi\right)$ are obtained, we suggest a series of validation steps to aid in policy selection. These steps focus on two main aspects: (1) Understanding the value of the policies and (2) Assessing the clinical sensibility of the policies.

For the first aim, we wish to ensure the learned policies outperform several baseline approaches: ``Doctors'', which is the treatment as it is assigned in the historical data (i.e., the correct practice), ``Random'', which randomly assigns treatment in the same proportion as the historical policy, ``Propensity'', which assigns treatment based on the estimated propensity scores (e.g. it assign $T=1$ if $\hat{e}(x)>0.5$), and ``Treat-all-with-t'', which assigns all patients to treatment $T=t$; i.e. there is one such policy for each treatment arm.

We propose the following approaches for comparing policies: (1) Direct and visual comparison, and (2) rank graph comparison. 
\begin{enumerate}[label=\arabic*.]
    \item \textbf{Direct and visual comparison:} First, we suggest comparing the policies' performance in terms of held-out policy value within each bootstrap round, counting overall ``wins''. See \cref{tab:acute_boot_compare} for an example. 
    
    Additionally, it is useful to visually examine the overall performance of various policies graphically, e.g. using box plots. This comparison may provide insight into the distribution of policy values; see \cref{fig:acute_policy_boxplot} for an illustration.

    \item \textbf{Rank graph comparison}: 
    We further propose measuring the policy values as function of treatment proportion \citep{Shalit2017EstimatingAlgorithms, imai2023experimental}. This analysis assumes each policy can provide a ranking of patients by treatment benefit. The graph then presents the policy's value under a varying threshold $q \in [0 \, , 1]$, which represents assigning $T=1$ only to a proportion $q$ of all the patients. Thus, the extreme thresholds ($q=0, q=1$) for each policy have the same policy values as  ``Treat-all-with-$t$`` ($t=0, t=1$) policies. See \cref{fig:acute_policy_val_th}   for an example of such a graph, and \cref{sec:rank_graph} for technical details.
    
\end{enumerate}

Following the above analysis, some conclusions might be made: If ``Treat-all-with-t'' policies are the best policies, it might suggest a fixed treatment decision is better than using personalized policy. If the policy ``Doctors'' has the highest value, it suggests current practice would not be improved by the model. Finally, if the ``propensity'' baseline is the best performing, it suggests that a sort of ``clinical wisdom of the crowds'' phenomenon might be the best option, as the propensity score estimator aggregates the clinicians' decision making process. Unlike the previous two cases, here one might move on with the propensity policy as a the basis for the treatment recommendation model going forward. 

Following the selection of policy model, we recommend characterizing the different sub-populations affected by the policy recommendations. Specifically, we suggest examining: (1) patients in each treatment recommendation arm, and (2) patients whose recommendations are in agreement or disagreement with the actual care as reflected in the data.
The characterization of treatment assignment policies is also known as policy summarization, which is a challenging task that is still an open area of research \citep{lage2019toward, matsson2022case}. We recommend characterizing the above-mentioned sets with the approaches described in \cref{subsec:subpop_char}.

Finally, we suggest plotting the \emph{Average Potential Outcome} ($\mathbb{E}\left[Y \mid i \in S_{sub} \right]$) of the above sub-populations $S_{sub}$ and present it in a tree-shape known as \emph{Outcome Tree}. See \cref{fig:acute_outcome_tree} for an example.

%% file: 04_acute_results.tex
In this case study, we explore a clinical dilemma arising when treating hospitalized patients suffering from acute kidney injury (AKI) as a consequence of acute decompensated heart failure (ADHF), also known as type 1 cardiorenal syndrome (CRS-1). When treating patients suffering from CRS-1, clinicians grapple with a difficult choice: whether to prioritize decongesion or intravascular volume and end-organ perfusion (very simplistically: whether to prioritize treating the heart condition or the kidney failure) \citep{Boulos2019TreatmentInjury, damman2014renalmeta-analysis, tang2010cardiorenal}.  This conundrum manifests itself in decisions around volume optimization therapies, such as adjusting the dosage of loop diuretics for patients. As of now, clear guidelines for managing this situation remain elusive \citep{Boulos2019TreatmentInjury,mcdonagh2021}. For additional information on ADHF and AKI, please see \cref{AKI and ADHF}.
 In the subsequent sections, we trace the steps of our proposed framework (\cref{pipeline}) as applied to this clinical challenge.

\subsection{Target Recommendation System: Preconditions}\label{acute_target_trail}

The target recommendation system was formulated over several months of in-depth discussions with the clinical team, including on-site visits to observe patient treatment in practice. Consequently, in alignment  with the preconditions specified in \cref{subsec:preconditions}, we have framed the target system as follows:
Our objective is to aid in determining the optimal diuretics dosage for patients hospitalized with ADHF who develop AKI during their hospital stay. We identify the decision point as the first time a laboratory result indicating a rise of $>0.3\frac{mg}{dl}$ in serum creatinine from the baseline (admission) value \citep{Boulos2019TreatmentInjury, damman2014renalmeta-analysis, tang2010cardiorenal}, providing a 48-hour window for the treatment decision.

Two options for diuretic dosage are considered: (1)\textbf{``Decrease''}: reduction ($T=0$), or (2) \textbf{``Increase''}: maintenance or increase ($T=1$). The outcome we evaluated relates to renal function \citep{mcdonagh2021}: Specifically, renal function is evaluated by the percent return to levels of creatinine at baseline (\emph{RTB}), where 100\% denotes complete return to baseline and 0\% indicates no change \citep{Boulos2019TreatmentInjury}. In general higher RTB values are desirable, though other clinical factors could be taken into consideration. The RTB measure is calculated as follows:
\begin{align}
\label{eq:RTB}
  RTB := \frac{crea_{d}-crea_{o}}{crea_d-crea_b},
\end{align}
\noindent where $crea_d$ is the lab result at the treatment decision point, $crea_o$ is the lab result at the outcome point (the last measurement within a week from the decision point), and $crea_b$ is the patient's baseline creatinine value taken at time of hospital admission.

We note that as currently defined, our formulation violates SUTVA as there are two versions of treatment for $T=1$: increasing or maintaining the dosage are not the same thing. Nonetheless, the two versions represent similar clinical intent of focusing on prioritizing decongestion by either maintaining or increasing the diuretic dosage. Thus, we can think of this clinical decision to focus on cardiac function as representing a closely aligned, if not the identical, treatment arm. In upcoming work, we look into a three-treatment-arm analysis of this decision.

\subsection{Target Recommendation System: Realization}
During our on-site visits and discussions with clinicians, we have made efforts to identify any possible factors that might affect treatment decisions and outcomes. We believe that following this process we have identified a comprehensive set of confounders for the model. All factors mentioned by our clinical collaborators as potentially impacting their treatment decisions were documented in the hospital electronic health record system\footnote{The system is in-house and maintained by the hospital, and includes forms created in the past by the clinical team in order to collect data about ADHF and AKI patients}. 

Additionally, we identified covariates affecting the outcome for improved prediction accuracy, based on insights from a study on a related patient cohort \cite{Gutman2022WhatOutcome} and based on domain expertise from our clinical partners (\cref{subsec:ident-data}). 

Consequently, we included approximately 200 covariates, covering all lab test results and timings, patient demographics, prior hospitalizations, other medications at admission and before the treatment decision, and the time elapsed from admission to the treatment decision. Missing data was imputed using the median of the training set, along with a missingness indicator. Time series data such as lab tests were used with summary statistics: mean, median, 10th and 90th percentiles, standard deviation, and first and last measurement for each signal from the admission time until the decision point. See a full list in \cref{tab:acute_table_one} in the appendix.

\subsection{Data}\label{subseq:acute_data}
We used data from a cohort of 6,940 patients, encompassing 12,027 patient visits, gathered between 2004 and 2015 at a large university medical center \citep{Gutman2022WhatOutcome}. The eligibility criteria were patients admitted primarily due to heart failure, as per the European Society of Cardiology criteria \citep{Mutlak2018TricuspidRisk, Aronson2013PulmonaryFailure, Ponikowski20162016Failure},
who during hospitalization experienced AKI. As stated above AKI was defined as any instance of serum creatinine increase exceeding $>0.3\frac{mg}{dl}$ from the value at admission. Applying this criterion yielded a study cohort of 2,157 stays.
This study follows the Declaration of Helsinki and has been approved by the institutional review committee on human research. See detailed description of the cohort in \cref{tab:acute_table_one} in the appendix. 

We note that while we made the utmost effort to include all possible confounders, we believe that going into a prospective trial we will be able to further improve the quality of our dataset, especially by applying definitions of treatments which more precisely align with current care. We will explore these refinements in future work. 

\subsection{Estimation}
In the following subsections, we describe the results following the outline in \cref{pipeline}. Some further results are detailed in \cref{apx_sec:acute_policy_val}.
The data were split into train, validation, and test sets, containing 1305, 322, and 530 stays, respectively.

\subsubsection{Propensity model}\label{acute_prop}

\begin{figure}[!htb]
    \centering
    \includegraphics[width=.7\textwidth]{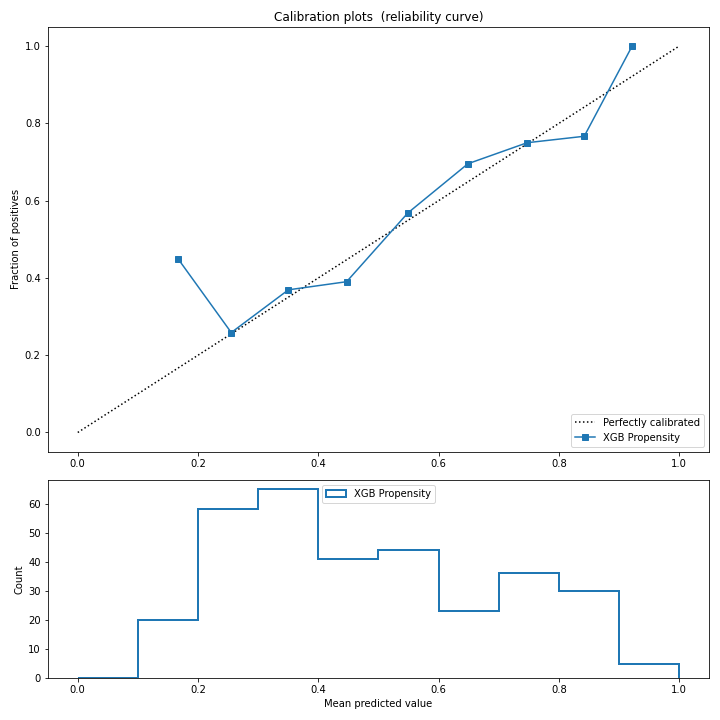}
    \caption{Calibration curve of propensity estimator, using XGBoost, on the validation set. }
    \label{fig:prop_calibration_cruve}
\end{figure}

For propensity score and overlap estimation (\cref{propensity explained}), we fit a model using XGBoost \citep{Chen2016XGBoost}. \cref{tab:acute_prop_metrics} includes all the relevant metrics for the propensity model; in particular, we found the model to be well-calibrated, see \cref{fig:prop_calibration_cruve}. 

\subsubsection{Propensity model insights}\label{acute_prop_insights}
To gain a better understanding of the propensity model (following \cref{propensity_insights}), we used SHAP \citep{Lundberg2017APredictions} to extract the features most correlated with the predictions, see \cref{fig:acute_prop_features}. We then consulted with domain experts who checked the validity of our findings,  approving that the top features are indeed likely to correlate with the clinical decision. 
Common support was selected based on the training data, and it contains subjects with propensity scores ranging from 0.21 to 0.9, see \cref{fig:acute_prop_triming}.

Out of the 530 patients in the test set, 461 were in the region of common support.

\subsubsection{Semi-synthetic outcome simulation}\label{acute_simulation}
As discussed in \cref{simulation_step}, we performed a simulation study whose goals are to ascertain whether causal effects can be uncovered from the data under reasonably favorable assumptions. Given the data of patients in the common support region (\cref{acute_prop}), we simulated potential outcomes for both treatment arms. The generated outcomes were chosen such that the CATE is a linear function of the covariates; see \cref{appendix_simulation}. 
Specifically, the linear CATE was designed to be a weighted average of the clinical decision maker's implied assessment of the best treatment as given by a log-linear propensity score estimator, and an arbitrary vector.
We run 6 simulation rounds with different seeds.

For the analysis, we used XGBoost, Ridge \& Lasso Regression, and BART \citep{Chipman2010BART:Trees, Hill2011BayesianInference}, and fitted models separately on both treatment arms, a practice known as ``T-Learner'' (see \cref{causal_estimate_methods}). 
We also used causal forest \citep{Wager2018EstimationForests} as a direct estimator of the CATE function.
Moreover, we used three ensemble-based policies: 
``Average'', which takes the average outcome prediction of all ML models as the estimated potential outcome and determines the treatment based on it; ``Majority'', which casts a vote where each ML model decides the treatment recommendation, and the majority vote determines the recommendation; and ``Consensus'', which provides a recommendation only for patients on which all ML models agree regarding the treatment, deferring all other recommendations. 

These policies were compared against the baselines ``Doctors'' 
, ``Random'', ``Propensity'', and``Treat-all-with-t'' (\cref{policy_insights}), as well as an ``Optimal'' policy (which we can derive since this is a simulation), as per \cref{simulation_step}.

\begin{table}[!htbp]
\centering
\begin{tabular}{l|cc|c}
Policy & $\hat{V}^{\text{DR}}\left(\pi\right)$ &  $\hat{V}^{\text{IPW}}\left(\pi\right)$ & $V\left(\pi\right)$ \\ \hline
XGBoost \citep{Chen2016XGBoost} & -0.378 (0.082) & -0.417 (0.069) & -0.392 (0.082) \\ 
Causal Forest \citep{Wager2018EstimationForests} & -0.325 (0.098) & -0.365 (0.081) & -0.334 (0.096) \\ 
Ridge & -0.442 (0.083) & -0.472 (0.080) & -0.432 (0.074)\\ 
Lasso & -0.447 (0.098) & -0.488 (0.093) & -0.450 (0.095) \\ 
BART \citep{Chipman2010BART:Trees} & -0.405 (0.088) & -0.430 (0.082) & -0.397 (0.073) \\ 
\hline
Average & -0.446 (0.094) & -0.469 (0.088) & -0.451 (0.087) \\ 
Majority & -0.372 (0.107) & -0.795 (0.049) & -0.339 (0.087) \\ 
Consensus & -0.444 (0.091) & -0.478 (0.085) & -0.442 (0.084) \\
\hline
Propensity & 0.079 (0.104)  & 0.063 (0.130) & 0.067 (0.101)\\ 
Increase all & 0.1 (0.115) & 0.096 (0.105) & 0.116 (0.116) \\ 
Decrease all & -0.16 (0.074) & -0.192 (0.072) & -0.136 (0.063) \\
Doctors & -0.081 (0.077) & -0.081 (0.077) & -0.081 (0.077) \\ 
\hline
\textbf{Optimal} & - & - & \textbf{-0.524 (0.074)} \\
\end{tabular}
\caption{Mean and SEM of simulation results over 6 randomization runs. For each policy we provided the Doubly Robust (DR) $\hat{V}^{\text{DR}}\left(\pi\right)$ and Inverse Propensity Weighting (IPW) $\hat{V}^{\text{IPW}}\left(\pi\right)$ estimated policy value, along with the real policy values $V\left(\pi\right)$. We also provide the optimal policy value, based on the simulated potential outcomes values. Lower value is better.}
\label{tab:acute_simulation_results}
\end{table}

\begin{figure}[!htbp]
    \centering
    \begin{subfigure}{0.45\textwidth}
        \includegraphics[width=1.\textwidth]{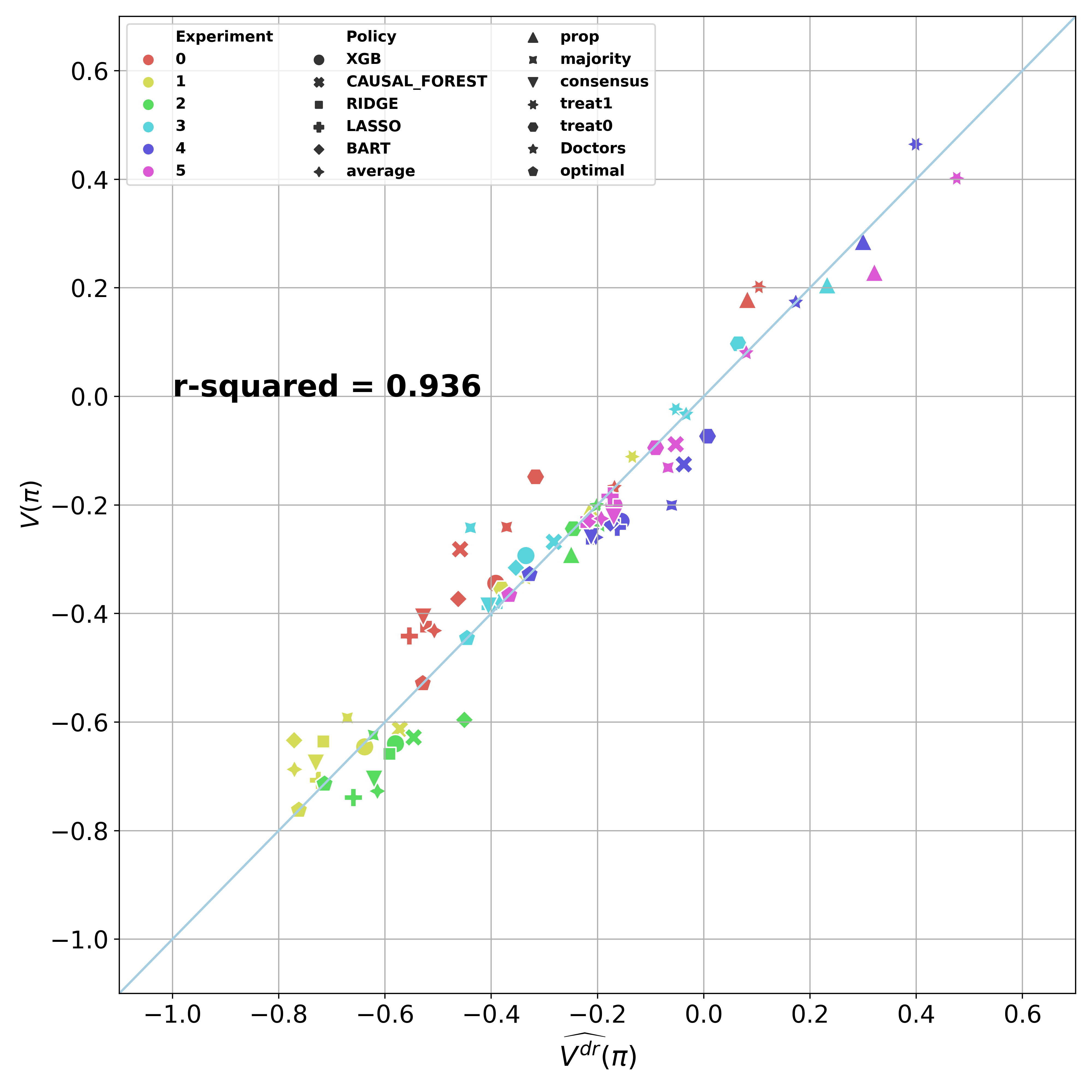}
        \caption{Simulation: Estimated policy values using doubly robust (DR) estimation vs. the true policy values.}
        \label{fig:sim_policy_val_dr_scatter}
    \end{subfigure}
    \hfill
    \begin{subfigure}{0.45\textwidth}
        \includegraphics[width=1.\textwidth]{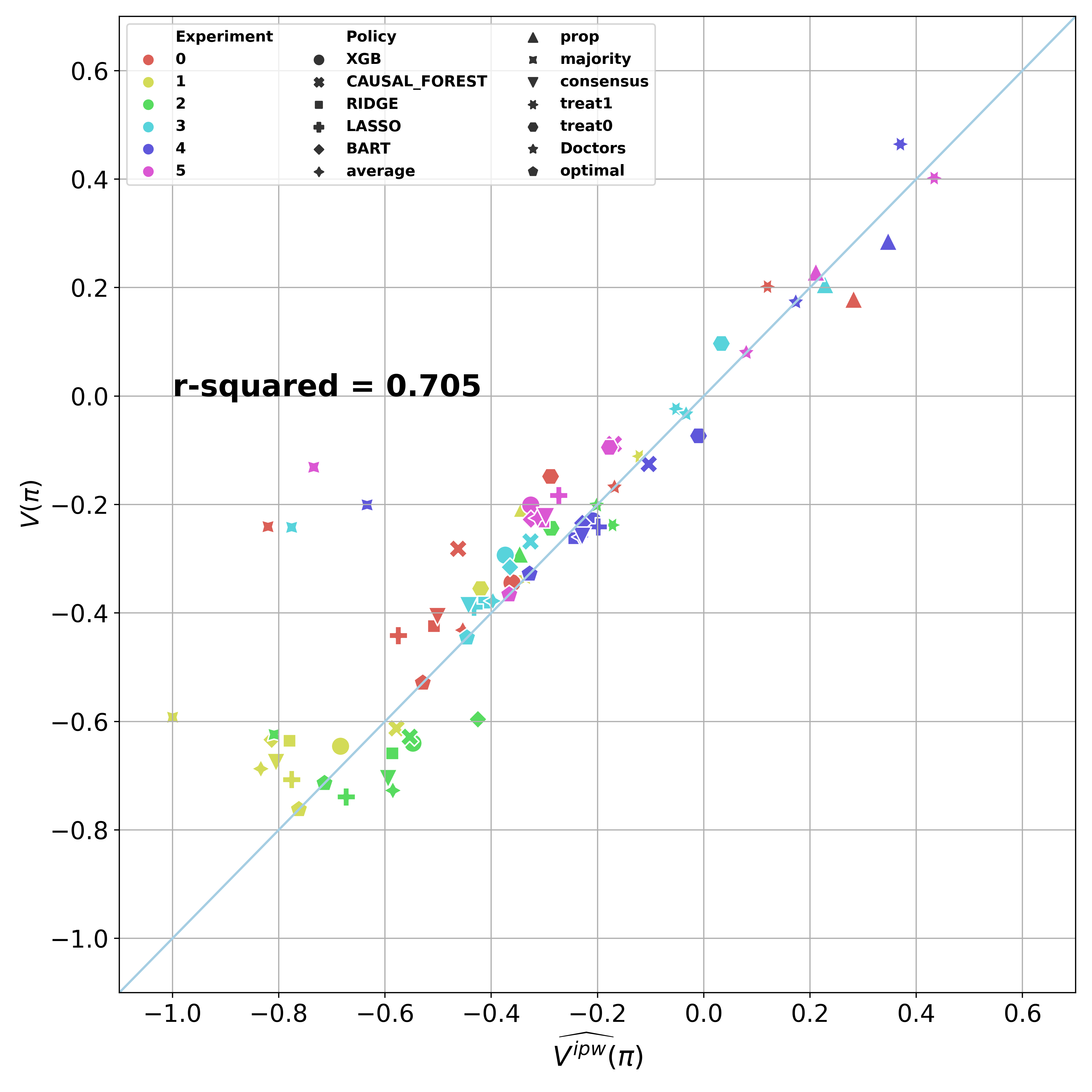}
        \caption{Simulation: Estimated policy values using inverse propensity weighting (IPW) estimation vs. the true policy values.}
        \label{fig:sim_policy_val_IPW_scatter}
    \end{subfigure}
    \caption[simScatter]{Scatter plot of the true policy value versus the estimated policy values, using DR (\cref{fig:sim_policy_val_dr_scatter}) and IPW (\cref{fig:sim_policy_val_IPW_scatter}) methods. The graph represents 6 simulation runs, using multiple policies: T-learner of XGboost (``XGB''), Ridge (``RIDGE''), Lasso (``LASSO''), and BART (``BART''), Causal Forest as a direct estimator (``CAUSAL\_FOREST''), average (``average''), majority (``majority'') and consensus (``consensus'') as Ensemble-based policies, and the baseline policies -- propensity based policy (``prop''), ``Increase'' to all (``treat1''), ``Decrease'' to all (``treat0''), Current Policy (``Doctors'') and the optimal policy.}
    \label{fig:acute_sim_policy_scatter}
\end{figure}

In \cref{simulation_step}, we suggested 3 indications that are required from the simulation to test whether the data is suitable for the task at hand.
In the rest of the sub-section, we analyze the results, presented in \cref{tab:acute_simulation_results} and \cref{fig:acute_sim_policy_scatter}, in light of those goals.

First, the results indicate that the estimated policy value is close to its real policy value.
This suggests that, under our current framework and evaluation method, the estimated policy value is similar to the real policy value. In \cref{fig:acute_sim_policy_scatter} we can further see that DR policy estimation is better aligned with the actual policy. 

We note that using the same CATE estimator as a plug-in estimator for policy value estimation can cause a bias in estimation, known as \emph{congeniality bias} \citep{curth2023magic}. The results in \cref{fig:acute_sim_policy_scatter} further indicate that the DR estimator has limited bias and gives a good estimation of the policy value.

We further observe that most policies outperform the current practice (``Doctors''), with the exceptions of causal forest, propensity, and majority vote. Unsurprisingly, policies that have linear underlying estimators (Ridge, Lasso) perform well, in tune with the linear nature of the outcome generation process. Along with XGBoost and BART, these policies yielded results that approximate the optimal policy. 

Based on these results, we conclude that our data is sufficiently robust and our modeling approach is appropriate for the task. Further details can be found in \cref{sp:acute_policy_detalied}.

\subsubsection{Outcome and CATE model}\label{acute_outcome}

For the CATE estimator models (\cref{subseq:Fit multiple CATE}), we used the same methods as in \cref{acute_simulation}, and added DragonNet \citep{shi2019adaptingeffects} as a direct estimator of the CATE function.
Although our outcome of interest is \emph{RTB} of creatinine, we note it is a noisy estimand. Looking into the definition of RTB (\cref{eq:RTB}), we see that ``outcome creatinine'' ($crea_o$) is the only unknown measurement at decision time. Therefore, we set $crea_o$ to be the outcome of interest for models at this stage, and transform it using \cref{eq:RTB} for downstream policy value estimation (\cref{acute_policy}). 
In \cref{tab:acute_outcome}, we present the prediction models' performance metrics. The results suggest satisfactory performance, with slight over-fitting for the BART model. 
 It is notable from the results that ``Increase'' prediction task was more difficult. This finding may stem from the diverse nature of the ``Increase'' arm, which includes both increase and stay with the same dosage. In future work, we intend to model them as 2 separate arms.

\begin{table}[!htbp]
\centering
\begin{tabular}{l|l|l|l|l|l|l}
\multicolumn{3}{c|}{Models} & RMSE & MAE & $R^2$ & std \\ \hline
\multirow{8}{*}{$T=0$:\textbf{ Decrease}} & \multirow{4}{*}{Train} & BART & 0.533 & 0.346 & 0.824 & \multirow{4}{*}{1.272} \\
& & XGB & 0.593 & 0.348 & 0.782 & \\
& & RIDGE & 0.744 & 0.455 & 0.657 & \\
& & LASSO & 0.754 & 0.433 & 0.648 & \\ \cline{2-7}
& \multirow{4}{*}{Validation} & BART & 0.808 & 0.538 & 0.58 &  \multirow{4}{*}{1.251}\\
& & XGB & 0.771 & 0.504 & 0.618 & \\
& & RIDGE & 0.772 & 0.491 & 0.617 & \\
& & LASSO & 0.773 & 0.48 & 0.615 & \\ \cline{1-7}
\multirow{8}{*}{$T=1$:\textbf{ Increase}} & \multirow{4}{*}{Train} & BART & 0.6 & 0.411 & 0.781 & \multirow{4}{*}{1.282} \\
& & XGB & 0.659 & 0.44 & 0.735 & \\
& & RIDGE & 0.688 & 0.477 & 0.712 & \\
& & LASSO & 0.808 & 0.533 & 0.602 & \\ \cline{2-7}
& \multirow{4}{*}{Validation} & BART & 0.988 & 0.573 & 0.588 & \multirow{4}{*}{1.545 } \\
& & XGB & 0.99 & 0.556 & 0.587 & \\
& & RIDGE & 1.059 & 0.646 & 0.527 & \\
& & LASSO & 1.021 & 0.581 & 0.56 & \\ \cline{1-7}
\end{tabular}
\caption{Performance metrics for outcome prediction models, in terms of Root Mean Squared Error (RMSE), Mean Absulte Error (MAE) and the correlation coefficient ($R^2$). We present the results for the T-learner models: BART, XGBoost (``XGB''), Ridge, and Lasso.}
\label{tab:acute_outcome}
\end{table}

\subsubsection{Outcome and CATE model insights}\label{sc:acute_outcome_insights}
In \cref{apx_sec:acute_outcome_models} we present several analysis results (Following \cref{subsq:cate_insights}). 
First, we show the model's error distribution along with an analysis of the important features of each model. We find these closely align with current medical knowledge. 
Furthermore, we outline the correlations between the CATE models, which indicates we have a positive, while small, correlation between the policies. See \cref{tab:pearson_outcome_models,tab:kendall_outcome_models} in the appendix. 
Additionally, we show the CATE calibration graph in \cref{fig:acute_xgb_cate_clib} in the appendix, which indicates our CATE estimates are calibrated in the sense presented in \citet{athey2019estimating}.

\subsubsection{Deferral set}\label{acute_deferal}
As explained in \cref{subsec:deferral}, our framework suggests only providing recommendations for patients for whom we have higher certainty that the recommended treatment will be beneficial. For patients for whom there is high uncertainty, we defer making a recommendation. The initial set of deferrals includes patients that were not part of the propensity-overlap region (see \cref{acute_prop}). Another set of patients to be deferred are patients that have high -- statistical or causal -- uncertainty for their recommendation. We obtain this using Quince \citep{jesson2021quantifying}, with Dragonnet \citep{shi2019adaptingeffects} as an underlying CATE estimator. 
Using a causal sensitivity parameter of $\exp(0.1)$ and a statistical point estimate we end up with an additional 139 (out of 481) deferred patients.

Therefore, we have two sets of deferrals: \emph{``Inclusive''}, which is based only on the common support set, and \emph{``Conservative''}, which includes both common support and high uncertainty deferrals. 
In \cref{apx_sec:acute_deferral_inisghts} we present insights on the trimmed patients, following \cref{subsec:defer_insight}.

\subsubsection{Policy} \label{acute_policy}
Following the CATE model results we established treatment policies following \cref{subsec:polval} using multiple different CATE estimators. For policy value estimation, we used XGBoost (XGB) as our plug-in estimator (see \cref{policy_val_eq_appendix}), and the outcome is given in RTB terms, as described in \cref{acute_target_trail}. 

In the next section, we present the results based on the ``Inclusive'' set. In \cref{fig:ipw_dr_policy_values_defer} and \cref{fig:acute_outcome_tree_conservative} we further present results for the ``Conservative'' set.

\subsubsection{Policy Insights}\label{acute_policy_insight}
Having established our candidate policies, we evaluate them on the actual data and compare them with current policy as performed by the clinicians at the hospital. The results given in this section are on held-out data, taken over 10K bootstrap rounds.

As a first evaluation phase, we investigated the breakdown of our policy recommendation versus the current treatment. Each of the treatment arms was divided into three bins -- whether our policy agrees with the observed treatment, disagrees, or was deferred. 

\begin{figure}[htb]
\centering
\includegraphics[width=0.75\textwidth]{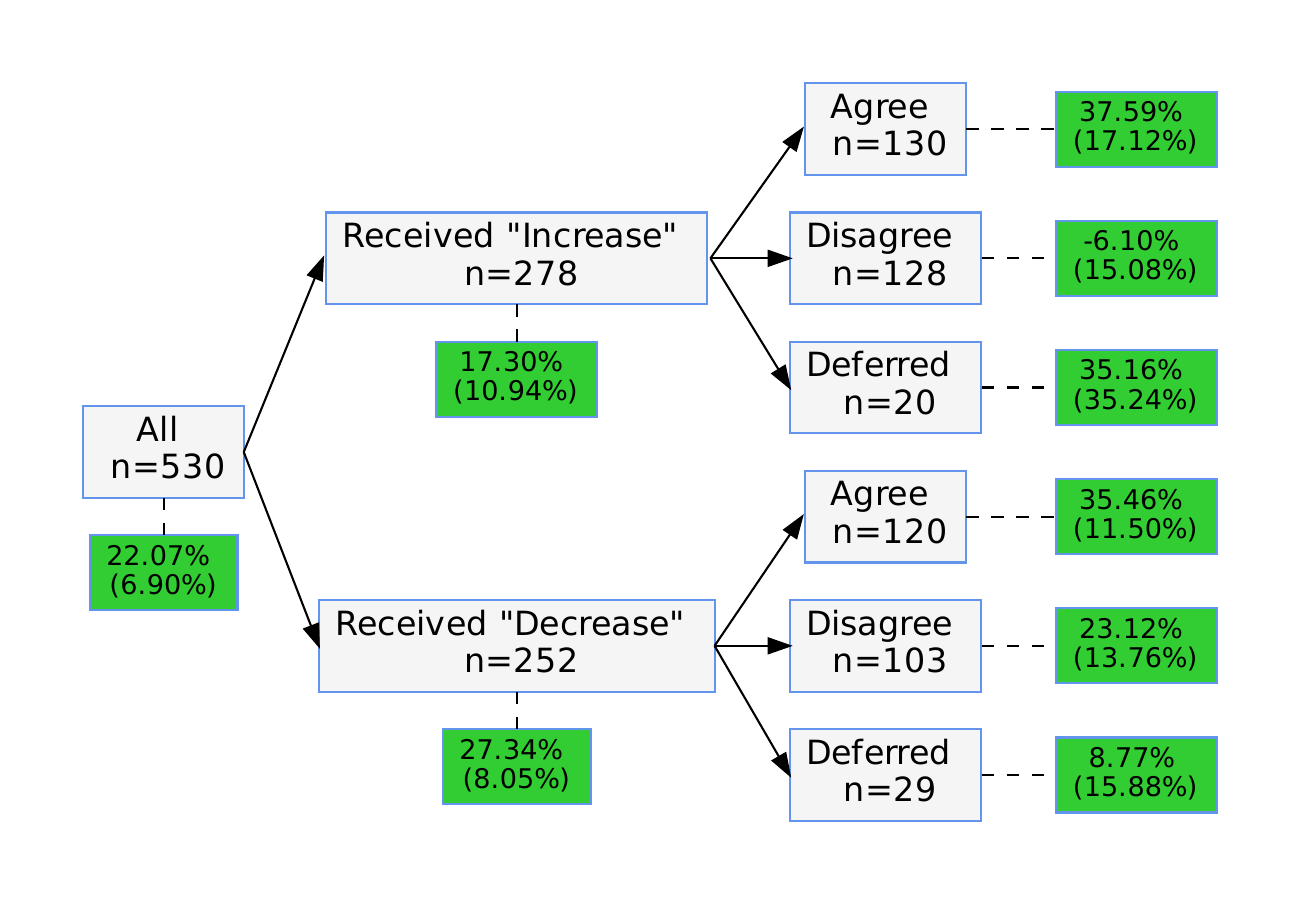}
\caption{Outcome Tree: A graph representing the mean RTB value for each policy group, using XGBoost T-learner as the basis for the policy. The first split (left) is by actual treatment received. The second split (right) is by whether the policy agreed or disagreed with the actual treatment, or whether it deferred. 
The gray boxes represent the number of patients in each subgroup, and the green boxes represent their mean (sem) RTB value.}
\label{fig:acute_outcome_tree}
\end{figure}

\cref{fig:acute_outcome_tree} shows the results in the form of an \emph{Outcome Tree}, as suggested in \cref{policy_insights}, with an XGBoost T-learner as the basis for the policy. The analysis suggests that in cases where the algorithm agrees with the current treatment, the patients have a mean RTB of $37\%$ in ``Increase' and $35.46\%$ ``Decrease'' arms. Those results are better than the cases where the policy disagrees, and better than the overall results in their respective arms.

\begin{table}[h]
\centering
\begin{tabular}{lcc}
Model & $\hat{V}^{\text{DR}}\left(\pi\right)$ &  $\hat{V}^{\text{IPW}}\left(\pi\right)$ \\ \hline
Ridge & 45.8\% & 43.8\% \\
XGBoost & 40.9\% & 40.9\% \\
DragonNet & 37.0\% & 38.3\% \\
Lasso & 31.2\% & 31.5\% \\
Causal Forest & 28.8\% & 28.4\% \\
BART & 20.2\% & 21.5\% \\
\hline
Average & 35.2\% & 35.4\% \\
Majority & 33.4\% & 31.4\% \\
Consensus & 33.9\% & 33.5\% \\
\hline
Propensity & 26.4\% & 32.1\% \\
Random & 22.8\% & 26.5\% \\
Increase all & 13.4\% & 13.3\% \\
Decrease all & 37.6\% & 41.6\% \\
Doctors & 22.1\% & 22.1\%
\end{tabular}
\caption{The policy values of using different policies, using both Doubly Robust (DR) and Inversed Propensity Weighting (IPW) estimators. The results are separated by their type -- first, the estimated policies, second the ensemble policies, followed by the simple baselines.}
\label{tab:acute_policy_value}
\end{table}
Next, we evaluate the policy value.  The results are presented in \cref{tab:acute_policy_value}. They indicate that the policy value of most models is better than the current treatment, as seen when using both DR and IPW policy value estimators. For example, it seems from the evaluation that using either Ridge ($45.8\%$) or XGB ($40.9\%$) as outcome models would almost double the RTB ratio compared to the current treatment policy value ($22.1\%$).

\begin{figure}[h]
\centering
    \includegraphics[width=.65\textwidth]{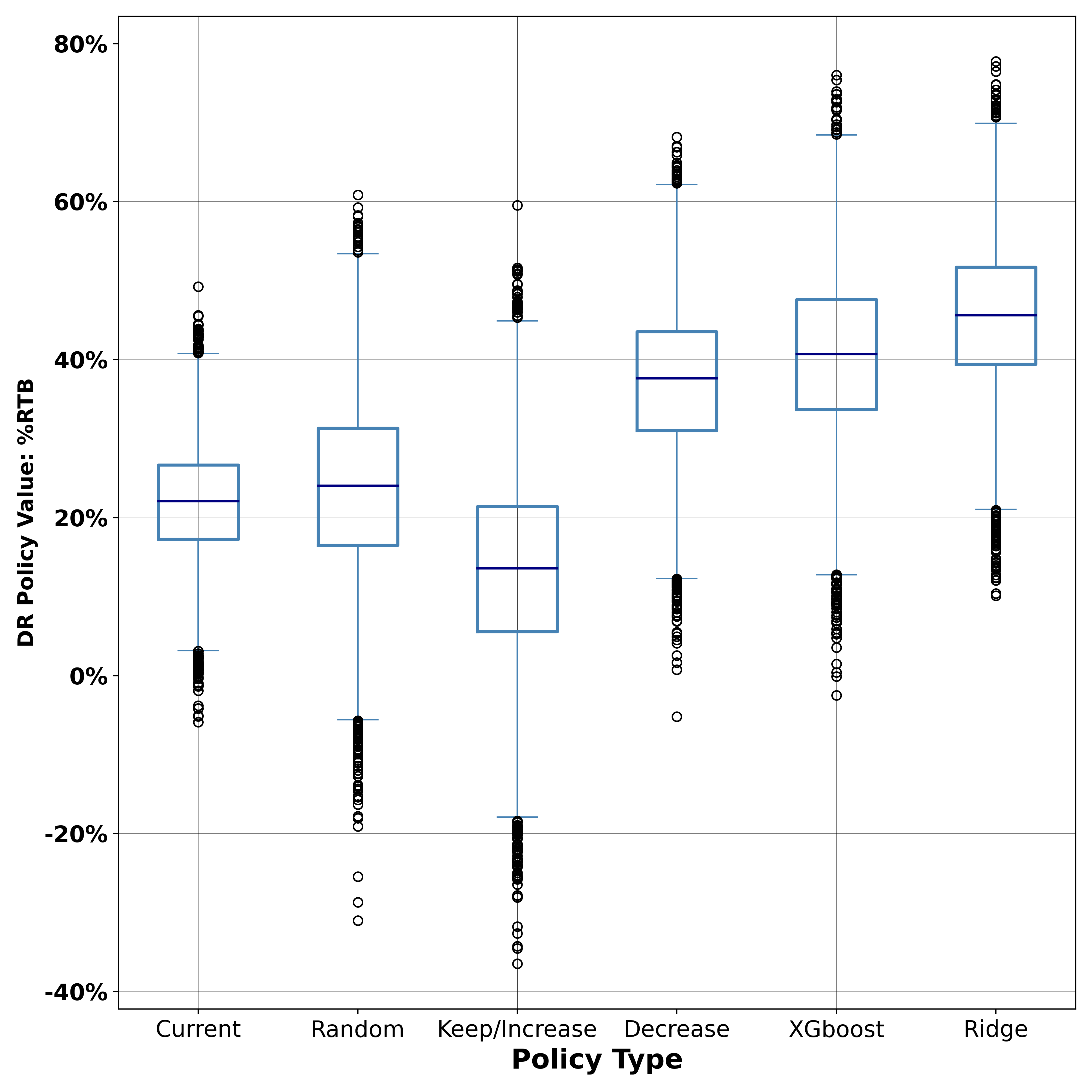}
    \caption{Policy value box-plot, the result of running 10K bootstraps evaluation on held-out data, showing 6 policies: \emph{Current}: current treatment, \emph{Random}: randomly assigning treatment at the same propotion as current treatment, \emph{Keep/Increase}: all patients given ``increase'', \emph{Decrease}:  all patient given ``decrease'', \emph{XGBoost}: T-learner of XGBoost models, \emph{Ridge}: T-learner of Ridge models.}
    \label{fig:acute_policy_boxplot}
\end{figure}

We further present the policy values in box-plots in \cref{fig:acute_policy_boxplot}. For clarity, we show only two best-performing policies (Ridge and XGBoost), the full results are in \cref{apx_sec:acute_policy_val} (\cref{fig:ipw_dr_policy_values} and \cref{fig:ipw_dr_policy_values_defer}).
The results indicate that our recommended policies yield better value to the patients compared to current care, where there is no intersection between their respective inter-quartile ranges. Interestingly, the results suggest that treating all patients with \emph{decrease} will result in better outcomes that the current treatment policy, yielding a value similar to both ridge and XGB. We discuss this finding in the end of the section.

\begin{figure}[h]
\centering
    \includegraphics[width=.65\textwidth]{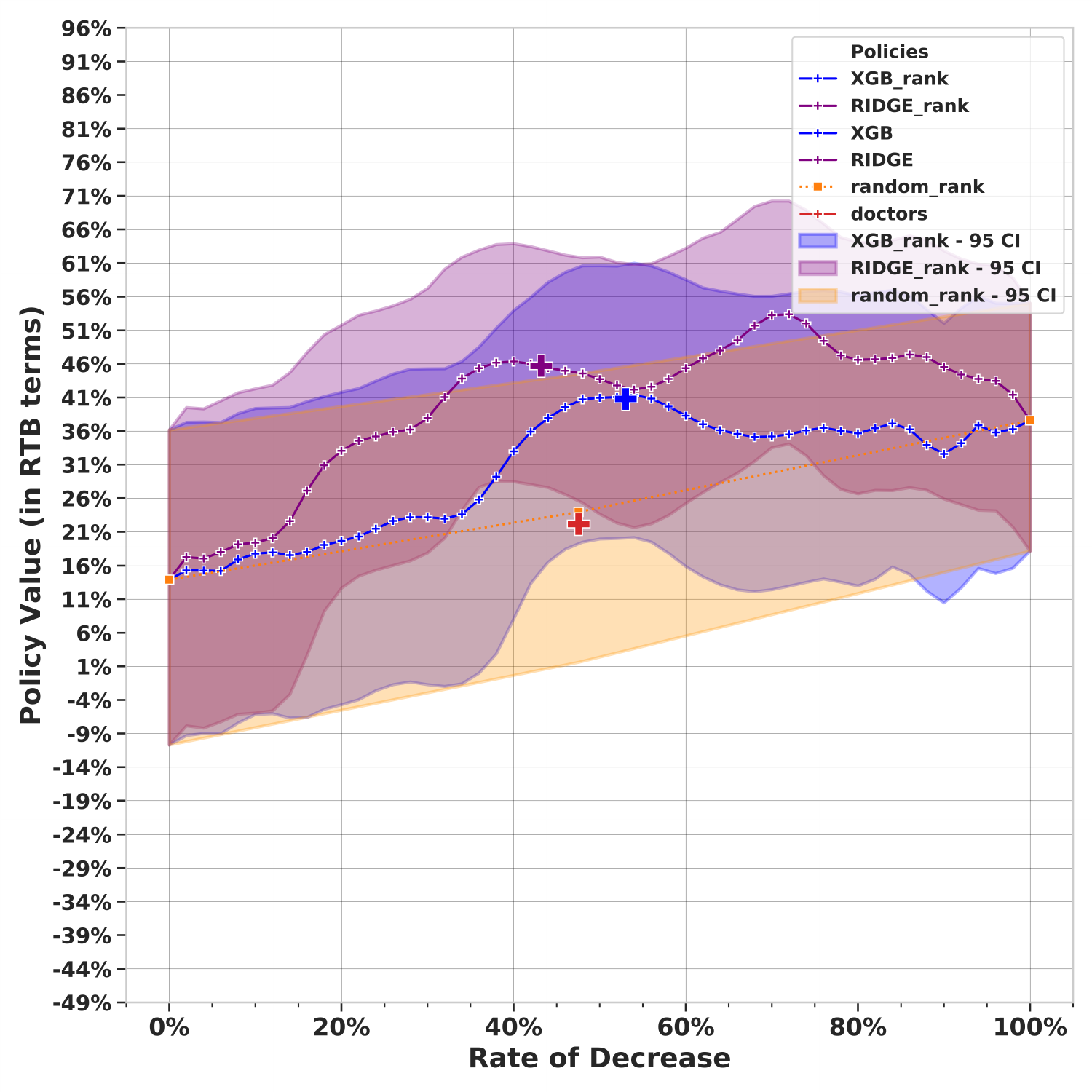}
    \caption{Policy value rank graph, result of running 10K bootstraps evaluation on held-out data, showing two T-learner models: XGBoost (``XGB'') and Ridge (``RIDGE''), compared with ``Doctors'' treatment policy (``doctors''). Each model ``\_rank'' markers (e.g., ``RIDGE Rank'') represent the policy value of providing ``decrease'' recommendation to exactly N\% of the population , based on a ranking of the estimated CATE values given by the model. We used a step-size of 2\%. 
    We further compared it to randomly ranking patients policy (``random\_rank''). The left and right endpoints represent the policy value of ``Treat-all-with-t'' policies, respectively. See explanation in \cref{policy_insights}.}
    \label{fig:acute_policy_val_th}
\end{figure}

In \cref{fig:acute_policy_val_th} we show the rank graph; for clarity we again show only the two best performing policies (Ridge and XGBoost). We can see that for most given thresholds, applying a ridge- or XGBoost-based policy will yield better performance than the current treatment policy across most quantiles, where ridge-based is statistically better with 95\% confident interval. The analysis suggests, similar to \cref{fig:acute_policy_boxplot}, that we cannot reject the possibility that using a model based policy has the same value as simply decreasing dosages for all patients -- we discuss this point below.

In \cref{tab:acute_boot_compare}, we present the result of a direct comparison of policy performances across bootstrap rounds. 
This analysis reveals that the Ridge and XGBoost policies outperform the existing treatment in 9972 and 9850 out of 10000 rounds, respectively. Additionally, these policies surpass most others in performance. Notably, however, we see here too that these policies advantage over the ``Decrease'' policy are not statistically significant.

\begin{table}[htb]
\begin{adjustwidth}{-1cm}{-1cm}
\scriptsize
\begin{tabular}{l|cccccccccccccc}
 & \textbf{BART} & \textbf{CF} & \textbf{Doc} & \textbf{LASSO} & \textbf{Prop} & \textbf{RIDGE} & \textbf{XGB} & \textbf{Avg} & \textbf{Con} & \textbf{Rand} & \textbf{Maj} & \textbf{DN} & \textbf{Dec} & \textbf{Inc} \\ \hline
\textbf{BART}          &     - &          2803 &    2434 &   332 &       1752 &    21 &    21 &      38 &        14 &  2124 &      538 &    138 &    343 &   5820 \\
\textbf{CF} &  7197 &             - &    4956 &  2408 &       3448 &   273 &   538 &    1454 &      1408 &  4342 &     2232 &   1349 &    479 &   7775 \\
\textbf{Doc}       &  7566 &          5044 &       - &  1356 &       2213 &    28 &   150 &     701 &       559 &  4132 &     1064 &    383 &    365 &   8513 \\
\textbf{LASSO}         &  9668 &          7592 &    8644 &     - &       7054 &   456 &   772 &    2025 &      2414 &  7559 &     6818 &   2310 &   2798 &   9679 \\
\textbf{Prop}    &  8248 &          6552 &    7787 &  2946 &          - &   141 &   567 &    1679 &      1692 &  5855 &     4271 &   1259 &   1040 &   8780 \\
\textbf{RIDGE}         &  9979 &          9727 &    9972 &  9544 &       9859 &     - &  7036 &    9007 &      9418 &  9835 &     9905 &   8839 &   8058 &   9988 \\
\textbf{XGB}           &  9979 &          9462 &    9850 &  9228 &       9433 &  2964 &     - &    8261 &      8816 &  9528 &     9735 &   7239 &   6339 &   9962 \\
\textbf{Avg}       &  9962 &          8546 &    9299 &  7975 &       8321 &   993 &  1739 &       - &      6170 &  8702 &     8525 &   4447 &   4325 &   9896 \\
\textbf{Con}     &  9986 &          8592 &    9441 &  7586 &       8308 &   582 &  1184 &    3830 &         - &  8678 &     8668 &   3654 &   3846 &   9918 \\
\textbf{Rand}         &  7876 &          5658 &    5868 &  2441 &       4145 &   165 &   472 &    1298 &      1322 &     - &     3186 &   1041 &   1113 &   8554 \\
\textbf{Maj}      &  9462 &          7768 &    8936 &  3182 &       5729 &    95 &   265 &    1475 &      1332 &  6814 &        - &   1177 &    945 &   9691 \\
\textbf{DN}        &  9862 &          8651 &    9617 &  7690 &       8741 &  1161 &  2761 &    5553 &      6346 &  8959 &     8823 &      - &   4692 &   9920 \\
\textbf{Dec}        &  9657 &          9521 &    9635 &  7202 &       8960 &  1942 &  3661 &    5675 &      6154 &  8887 &     9055 &   5308 &      - &   9487 \\
\textbf{Inc}        &  4180 &          2225 &    1487 &   321 &       1220 &    12 &    38 &     104 &        82 &  1446 &      309 &     80 &    513 &      - \\
\end{tabular}
\end{adjustwidth}
\caption{This table represents the direct comparison of policies, where columns and rows represent the different policies, and each cell represents how many bootstrap rounds out of 10,000 was the policy value of the ``row'' policy better than the ``column'' policy. For example, the T-learner policy with XGB (``XGB'') had a higher policy value 9,850 times (out of 10k rounds) than the current policy (``Doc''), which is $98.5\%$ of the rounds.
The models are: T-learners with backbone ML models -- BART, Lasso, Ridge, and XGBoost (``XGB''), Direct methods -- Causal Forest (``CF''), DragonNet (``DN''), 
baseline methods -- average (``Avg''), consensus (``Con''), majority (``Maj''), Random (``Rand''), and policies representing providing ``Increase'' and ``Decrease'' to all (``Inc'' and ``Dec'' respectively), alongside model based on the propensity score (``Propensity'') and current treatment (``Doc'').}
\label{tab:acute_boot_compare}
\end{table}

Overall, our analysis suggests that a policy that recommends decreasing the dosage for all patients would yield a similar performance to our model-based policy, in terms of patient RTB. While the performance is similar, we note that our recommendations decrease the dosage only for about 43\% - 55\% of the patients which allows more flexibility with respect to other outcomes, which we present next.

\paragraph{Patient Re-Hospitalization} 
As outlined in the opening of \cref{acute}, the clinician's treatment decision is based on trying to maintain a delicate balance between kidney and cardiac function. Thus, we further evaluate our policies on cardiac function-related outcomes.
The cardiac function is estimated by checking if the patient was re-hospitalized due to AHF within 30 days of the treatment decision, which would indicate a failure in managing the patient's heart condition.

We evaluated the policy values using the same approaches described above, where for the DR policy value estimation we used an $L_2$ regularized Logistic-Regression trained to predict 30-day re-hospitalization on the train split, as a plug-in estimator.

\begin{figure}[!htb]
\centering
    \includegraphics[width=.65\textwidth]{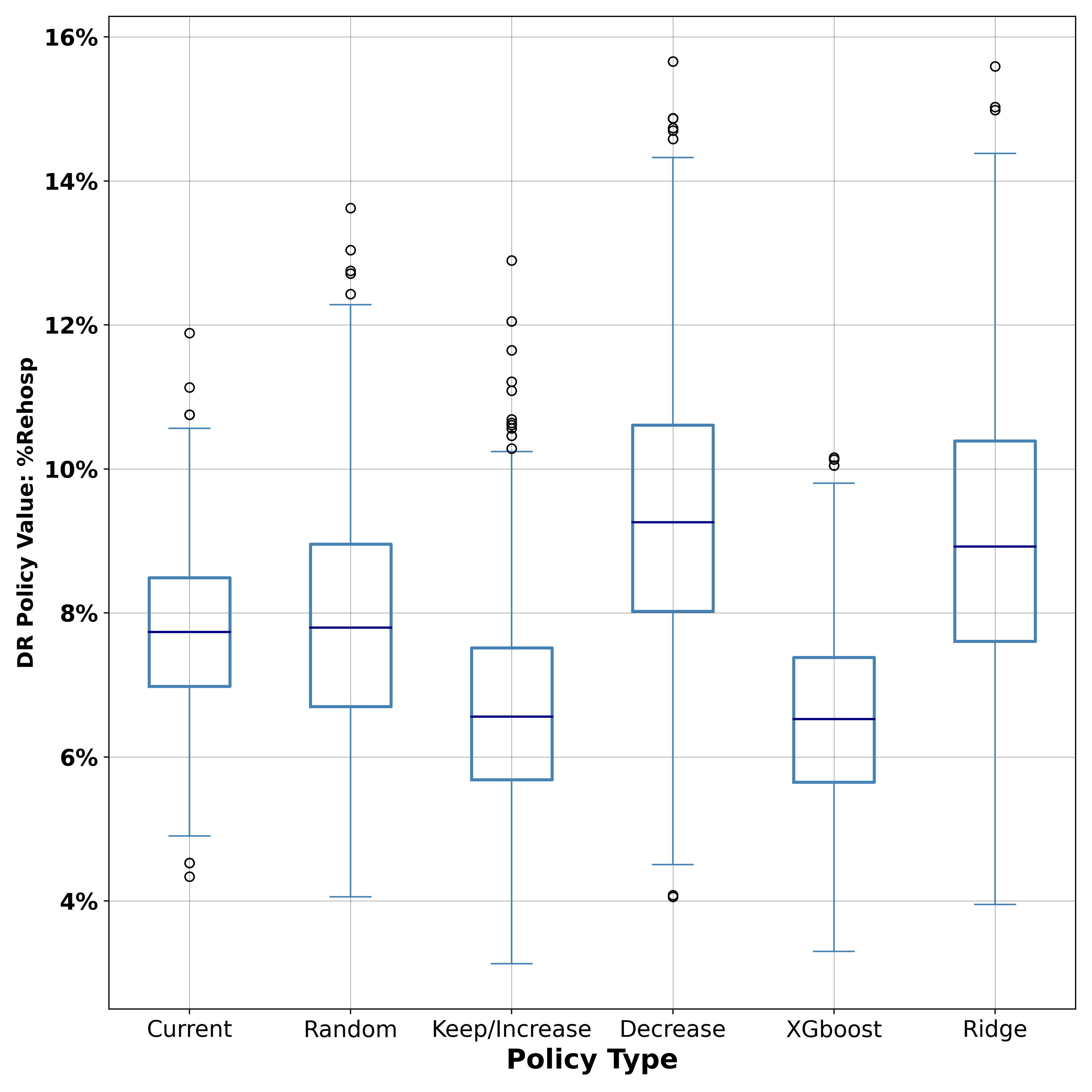}
    \caption{Policy value box-plot, a result of running 10K bootstraps evaluation on held-out data, in terms of re-hospitalization 30 days from decision-point. The DR policy value where estimated using $L_2$-regularized logistic regression estimator.
    The policies are: \emph{Current}: current treatment, \emph{Random}: randomly assigning treatment at the same proportion as current treatment, \emph{Keep/Increase}: all patients given ``increase'', \emph{Decrease}:  all patient given ``decrease'', \emph{XGBoost}: T-learner of XGBoost models, \emph{Ridge}: T-learner of Ridge models.}
    \label{fig:acute_policy_rehosp_boxplot}
\end{figure}

We present the re-hospitalization policy-value box-plot in \cref{fig:acute_policy_rehosp_boxplot}. The results indicate that the XGBoost-based policy we obtained earlier achieves a lower re-hospitalization rate than current care, and similar to the ``Increase'' policy. Furthermore, unlike the analysis of the main outcome, here the ``Decrease'' policy performs worse than current care and is similar to the Ridge-based policy.  
This, combined with the results on the main outcome (\cref{fig:acute_policy_boxplot}), suggests that XGBoost-based policy might balance the two outcomes better than current care and other baselines, improving both kidney and cardiac function.

%% file: 06_related_work.tex
In recent years, several authors sought to give guidelines on how to learn treatment effect and produce a treatment policy from data. In this section, we will provide a short overview of those works and our added contribution.

\textbf{Observational studies} 
\citet{Powell2021TenStudy} and \citet{dang2023causal} suggested guidelines for using observational data for estimation of average treatment effects.
Those works share a similar perspective to ours regarding causal identification and the importance of cross-discipline collaboration. However, our focuses on establishing a policy for treatment recommendation on the individual level, rather than estimating a single overall average effect. 

As mentioned in \cref{causal_estimate_methods}, estimation of \emph{Heterogeneous Treatment Effect} (HTE) from both observational and experimental data is an active area of research \citep{kent2018personalizedeffects, kravitz2004evidence, hayward2005reporting, Dahabreh2016UsingEvidence, Chen2017AScoring, anoke2019approachesconfounding, Luedtke2017EvaluatingSubgroup, ling2023emulate, kent2020predictive}. It mainly focuses on identifying subgroups with differing causal effects.
As in the ATE case, our work is using tools from the HTE literature, but our focus on is on estimating a useful individualized policy rather than estimating effects for sub-groups. 

\citet{sharma2020dowhy} developed a Python package and a framework to estimate causal effects, including in observational cases, in 4 stages closely resembling our suggested framework. While this framework suggests many evaluation methods, they are mostly related to ATE and do not take into consideration the deferral option in policy, and lack guidelines for how to conduct the identification phase. Their framework is not focused on using healthcare data and recommending actions to clinicians. 

\textbf{Individual Treatment Rules (ITR)}
As explained in \cref{eval}, ITR is a common name for a learned function $\pi\left(X\right)$ that assigns a treatment to each individual based on their characteristics. 

\citet{boominathan2020treatment} have suggested a method for forming a policy from data in cases where we know all the outcomes (whether based on biological knowledge, advanced lab tests, etc.), but they are not available to the decision maker during treatment assignment. This setting does not apply to our case. Moreover, the authors do not tackle the identification and evaluation steps we suggest in this work.  
\citet{meid2020usingdata} suggested a general framework to learn ITR in cases where HTE has be found for a given treatment, based on effect modifiers. While their work has similar settings to ours, we go further from the simulation study, and provided the additional required steps of identification and validation for this task. Moreover, we provided steps in cases of high uncertainty and deferral.

\citet{meid2022can} presented a real-word case study, using CATE to recommend personalized treatment. Following similar ideas to our suggested framework, their work explored the estimation and some of the evaluation steps. Our work expands on this by introducing analysis and validation phases for each stage, and our focus on identification. 
Moreover, our framework includes a deferral option for high uncertainty or weak overlap cases. 

\citet{petersen2014causal} suggested a theoretical framework (``Causal Roadmap'') on how to think about estimating causal parameters, as an alternative to the ``Target Trial'' \citep{rubin2004teaching, hernan2011withepidemiology,Hernan2016UsingAvailable, dickerman2019avoidable} discussed in \cref{subsec:preconditions}. Their framework focuses on selecting the most suitable model in reference to the presumed data-generating process and causal assumptions.

In companion papers, \citet{montoya2022optimal} and \citet{montoya2022estimators} demonstrate an application of this theoretical framework to establish ITR, in RCT settings. In \citet{montoya2022optimal} they consider different treatment estimation models and policy learning, again under RCT settings. Following that, \citet{montoya2022estimators} explore policy value estimation.
We differ from this work as we suggest a framework for learning a policy in the observational settings with a focus on \emph{causal identification}, and in addition our discussion of validation steps and the role of the deferral option.

%% file: 07_Discussion.tex
In this work we propose a framework for generating patient-level treatment recommendation models based on patient health data, including observational data.
We provide detailed steps and propose how to validate them, with the goal of developing a reliable and robust treatment recommendation policy.

Rather than prescribing a specific causal effect estimation method, we give a general guideline for developing treatment policies. We recognize that various estimation methods may be more suitable depending on the target system's characteristics and complexity.
We applied our framework to a real-world treatment recommendation challenge in acute healthcare settings. As outlined in \cref{acute_policy}, our framework yields promising results, suggesting that the learned treatment policy may outperform current care and lead both to better renal functions and lower re-hospitalization rates.

A notable challenge in healthcare today is the gap between the rapid development of models and their actual implementation in bedside settings \citep{eini2022tell,Bates2020ReportingIntelligence, rajpurkar2022ai, harris2022clinical, zhang2022shifting, heaven2021hundreds, smit2023causal,balagopalan2024machine}.
Although our work does not directly tackle this gap, it is motivated by it.
Our framework aims to help taking a step towards closing the gap in two key ways. First, we explicitly focus on actionable recommendations as part of the treatment process. Second, given the high stakes of deploying systems in healthcare settings, our framework emphasizes ways of mitigating risks by prioritizing responsible, expert-in-the-loop model development and testing.

While those steps are not exhaustive, we believe they can contribute to addressing the challenges inherent in this gap.
We recognize that our rigorous approach may not be necessary in every scenario. For example, in lower-stakes environments such as e-commerce recommendation systems where speed is crucial, some steps of our framework could be omitted.

While the framework provides promising results, it is essential to check its usefulness ``in the wild'' \cite{McCradden2020ClinicalIntelligence}, where conditions may differ. Testing such systems in real-world settings remains an active area of research \citep{wiesenfeld2022ai,joshi2025ai}. The ultimate test for the type of systems we discuss here would be a clinical trial in the field, comparing clinical outcomes when recommendations are available vs. when they are not. This requires real-time integration with the hospital EHR, as the system will need access to the full set of covariates in order to make real-time recommendations. Moreover, questions of generalization among different hospitals and health settings will also come into play here \citep{behar2023generalization}.

Looking forward, we believe that incorporating more advanced estimation methods should improve the accuracy of our policy. Additionally, improvements such as leveraging multi-modal data, better utilization of time-series information, accounting for survival bias, and leveraging mechanistic knowledge whenever possible can contribute to refining the model's performance.

For instance, we note that the analysis of the patients' re-hospitalization might suffer from competing risks or survival bias: some patients might have died during the 30 days from the decision point, and thus their potential outcome of re-hospitalization is invalid \citep{rubin2006causal1}. Moreover, when accounting for secondary outcomes, such as re-hospitalization, the confounders might differ from those pertaining to the main outcome.

More broadly, there are some crucial areas which our framework does not directly address. First, multiple or continuous treatments, while sharing many similarities with binary treatments, still require special care especially when it comes to propensity score modeling and the accompanying deferral rules. A more difficult challenge is balancing multiple outcomes; for example, in our cardio-renal case, clinicians wish to optimize both renal function and cardiac functions, goals which might be at odds, and a balance which is difficult to quantify. Notions such as the win-ratio \citep{pocock2012win,even2025rethinking} might be relevant in such cases 

Another extension to the framework would be developing dynamic treatment recommendation policies \citep{murphy2003optimal, smit2023causal}. These would support physicians beyond the initial decision-making stage. In cases such as the cardio-renal scenario we discuss, physicians could benefit from guidance on managing the patient's entire treatment plan, especially after the onset of Acute Kidney Injury (AKI). This is a challenging task, as each decision increases the complexity of possible treatments and reduces the overlap between the patients' journeys \citep{gottesman2019guidelines}.

A further promising research effort is modeling directly the deferral decision, as opposed to our current uncertainty based approach, which might be sub-optimal in terms of maximizing the combined expert-model decision making \citep{madras2018predict,mozannar2020consistent,ghoummaidact,gao2023confounding}.
Downstream from our recommendation system is the question of human-computer interaction, and how to present recommendations to clinicians effectively \citep{bansal2021most,jacobs2021machine,meyer2022impact}. 

Finally, beyond the distribution shift due to the difference between retrospective and prospective environments and across hospitals mentioned above, we must also consider shifts caused by the presence of the treatment recommendation system itself \citep{finlayson2021clinician,perdomo2020performative}. This might change the entire treatment regime, beyond the decision point, and thus may impact the estimation accuracy and validity of policy estimation.

%% file: 09_appendix.tex
\appendix

\renewcommand{\thefigure}{S.\arabic{figure}}
\setcounter{figure}{0}
\renewcommand{\thetable}{S.\arabic{table}}
\setcounter{table}{0}
\renewcommand{\thealgocf}{SA\arabic{algocf}}
\setcounter{algocf}{0}

\section{Questions for clinical-algorithmic team collaboration}\label{sec:clinical-questions}
We propose that the system be defined by discussion of the following points with the clinical team.
\begin{enumerate}
    \item \emph{What is the treatment decision for which the clinical team would like assistance from an algorithmic model?}

    \item \begin{enumerate}
        \item \emph{At what point in the clinical workflow would the clinical team want the recommendation?}

        \item \emph{Does the clinical recommendation time point correspond to the time where the decision is currently made?}
   
    \end{enumerate}
\end{enumerate}

\section{Which covariates should be used}\label{sec:variable_selection}
Creating and validating a causal graph that captures all of the hundreds of potential variables involved in a clinical problem is often challenging and time-consuming. We thus propose an alternative that could be useful for many cases of interest.  

To simplify the task, we start by attempting to identify the potential confounders. Towards this goal, we ask experts to answer the following questions: 
\begin{enumerate}[label=\alph*.]
    \item Which factors plausibly affect the treatment decisions as they occurred in the historical data?
    \item Of the above factors, which also plausibly affect the outcome?
\end{enumerate} 

The answer to the second question can also come from the scientific literature at large. Factors that plausibly affect both the outcome and the treatment decisions as they occurred in the cases present in our data are potential confounders. Next, we need to make sure these potential confounders are indeed represented in the data, or, if not available directly, at least have a good proxy. E.g., one can use hematocrit and hemoglobin biomarkers as proxies for congestion in AHF patients \citep{duarte2015prognostic}. If we find an important confounder which is \emph{not} recorded in the data and has no good proxy, then we must either find a way to obtain the required confounder, or pursue a completely different analysis. 
    
Assuming we are satisfied that the bulk of confounding factors is represented in the data, we should consider which covariates we should \emph{discard} due to the possibility of inducing bias. We recommend the following procedure for each covariate in the dataset: 
    \begin{enumerate}[label=\alph*.]
        \item Is it affected by the treatment decision, or does it occur after the treatment decision? If it is, it should be discarded from future analysis. \\
        \\If not:
        \item Does it plausibly affect the treatment decision as reflected in the available historical data (or is a proxy for such a variable)?
        \item Does it plausibly affect the outcome (or is a proxy for such a variable)?
    \end{enumerate}
    
    Having discarded the covariates for which the answer to \textbf{a} is ``yes'', we are left with the following: Covariates for which the answer to \textbf{b} AND \textbf{c} is ``yes''; these are potential confounders and must be included in the analysis. Covariates for which the answer to \textbf{b} is ``no'' and \textbf{c} is ``yes''; these are (potentially) what are known as effect modifiers, and should also be included in the analysis as their inclusion can reduce variance \citep{huenermund2022choice}.  Covariates for which the answer to \textbf{b} is ``yes'' and to \textbf{c} is ``no''; these are potential instruments, and should be discarded from analysis as their inclusion might \emph{increase} variances \citep{ding2017instrumental,huenermund2022choice}. Finally, covariates which do not plausibly affect outcome or treatment, and are not proxies for such variables, should usually be discarded from the analysis as they are unrelated to the problem at hand.

    While the above is a simplification of the more rigorous process of identifying a backdoor blocking set and other causal identification schemes \citep{pearl2009causality}, we find that in many cases this is a more realistic endeavor than constructing a large, accurate, causal graph. 

    We also note that in principle, proxy variables should be used differently from variables which are direct causes of treatment or outcome. However, in practice, we believe this distinction is often far from clear, and there are not many methods for rigorously dealing with proxy variables when the proxy mechanism and graph are unknown \citep{tchetgen2020introduction,shi2020selective,sverdrup2023proximal}. This is an area where we believe further research is called for.

\section{Simulation details}\label{appendix_simulation}
Given real patient data $\{(x_i,t_i)\}_{i=1}^n$ of $n$ patients with $d$ covariates, we wish to simulate two potential outcomes $Y^0, Y^1$ for each patient; we outline the process in \cref{alg:cap}. Our goal is to create a learnable yet somewhat realistic simulated CATE function. Towards that end we simulate the two expected potential outcomes $\mathbb{E}\left[Y^1 \mid x\right]$ and $\mathbb{E}\left[Y^0 \mid x\right]$ as linear functions of the covariates: $w_1^\top x$, and $w_0^\top x$, respectively. This leads to a linear CATE function of the form $\tau(x) = (w_1-w_0)^\top x = \Delta^\top x$ where $\Delta = w_1-w_0$. We generate the vectors $w_0, w_1$ so that their difference $\Delta$ will have a meaningful connection to the underlying data, as we now explain. 

We start by fitting a logistic propensity model $\hat{e}(x) = p(T=1 \mid x)$ using the real treatment assignments,  obtaining a coefficient vector which we denote $\beta_{prop}$. We consider this vector as an embedding of the clinical knowledge driving treatment assignment. We then further generate a random vector with the same norm as $\beta_{prop}$, and combine the two using a weighting parameter $\lambda \in [0, 1]$ to generate $\Delta$. The CATE function is thus a linear function which is a weighted combination of the propensity score coefficient vector and a random vector. The weighting factor $\lambda$ represents the degree to which the clinicians' understanding of the important factors regarding the outcome is correct: For $\lambda=1$, the simulated $Y^1$ for patients who receive treatment $T=1$ will tend to have a higher potential outcome, meaning clinicians made the correct treatment decision (assuming here that higher outcomes are better). 

We draw the potential outcome values themselves from a Gaussian distribution with means $w_1^\top x$, $w_0^\top x$ and variances $\varepsilon_1$, $\varepsilon_0$.
We set the variances to be 1.2 SD times the variance of $ w_t^\top x$ across patients for each outcome arm $t$. Finally, we re-scale the resulting average CATE to match a pre-defined value that represents a clinically reasonable value.

The full simulation procedure and settings are presented in \cref{alg:cap}.

\begin{algorithm}
\caption{Simulation Parameters}
\label{alg:cap}
\begin{algorithmic}[1]
\REQUIRE $\lambda \in [0,1]$, desired effect $C \in \mathbb{R}$
\REQUIRE $\{(x_i,t_i)\}_{i=1}^n$ (baseline data)
\STATE $\beta_{\text{prop}} \gets$ coefficients from logistic regression for $p(t_i=1|x_i)$
\STATE $\beta_{\text{rand}} \sim \mathcal{N}\Bigl(0,\Bigl(\frac{1}{\sqrt{d}}\Bigr)^2\Bigr)$
\STATE Normalize:
\STATE \quad $\beta_{\text{prop}} \gets \frac{\beta_{\text{prop}}}{\|\beta_{\text{prop}}\|_2}$
\STATE \quad $\beta_{\text{rand}} \gets \frac{\beta_{\text{rand}}}{\|\beta_{\text{rand}}\|_2}$
\STATE $\Delta \gets \sqrt{\lambda}\,\beta_{\text{prop}} + \sqrt{1-\lambda}\,\beta_{\text{rand}}$
\STATE $a \gets \frac{1}{n}\sum_{i=1}^n |x_i \cdot \Delta |$
\STATE $\Delta \gets \Delta \cdot \frac{C}{a}$
\STATE $w_0 \sim \mathcal{N}\Bigl(0,\Bigl(\frac{1}{\sqrt{d}}\Bigr)^2\Bigr)$
\STATE $\sigma_0 \gets \mathrm{SD}(w_0^\top X)$
\STATE $w_1 \gets \Delta + w_0$
\STATE $\sigma_1 \gets \mathrm{SD}(w_1^\top X)$
\STATE Initialize empty vectors $Y_0$ and $Y_1$
\FOR{$i = 1$ to $n$}
  \STATE $\varepsilon_{0,i} \sim \mathcal{N}\Bigl(0,\bigl(1.2\,\sigma_0\bigr)^2\Bigr)$
  \STATE $Y_{0,i} \gets w_0^\top X_i + \varepsilon_{0,i}$
  \STATE $\varepsilon_{1,i} \sim \mathcal{N}\Bigl(0,\bigl(1.2\,\sigma_1\bigr)^2\Bigr)$
  \STATE $Y_{1,i} \gets w_1^\top X_i + \varepsilon_{1,i}$
\ENDFOR
\RETURN Policy $\pi^*(x)$ defined as
\[
\pi^*(x)=
\begin{cases}
1, & \text{if } \Delta^\top x < 0,\\[1mm]
0, & \text{otherwise.}
\end{cases}
\]
\RETURN Vectors $Y_0$ and $Y_1$
\end{algorithmic}
\end{algorithm}

\section{Policy value Formulation}\label{policy_val_eq_appendix}
As detailed in \cref{eval} and \cref{policy_insights}, the policy value $V\left(\pi\right)$ is defined in \cref{eq:policy_val}.
In this work we suggest using a flavor of methods suggested by \citep{dudik2014doubly,lunceford2004stratificationstudy}, where each sample $i$ contributes weight $w_i$: $w_i = \frac{\mathbb{I}_{t_i=\pi(x_i)}}{p^{*}(t_i=\pi(x_i)|x_i)}$, and we provide the DR and IPW estimands:

\begin{align*}
    &\widetilde{V}_{\text{DR}}(\pi) =  \frac{1}{\sum_{i=1}^n w_i} \sum_{i=1}^n w_i \cdot (y-\hat{y}^{\pi(x_i)})  + \frac{1}{n} \sum_{i=1}^n\hat{y}^{\pi(x_i)}, \\%\label{eq:dr_policy}
    &\widetilde{V}_{\text{IPW}}(\pi) =  \frac{1}{\sum_{i=1}^n w_i} \sum_{i=1}^n w_i \cdot y,%\label{eq:ipw_policy}
\end{align*}
where $p^{*}(t_i=\pi(x_i)|x_i)$ is the propensity score model trained on all the data, and $\hat{y}^{\pi(x_i)}$ is the estimated potential outcome under the policy $\pi(x_i)$, given by plug-in estimator. The plug-in estimator can be any outcome model trained in the outcome estimation stage (\cref{subseq:Fit multiple CATE}). 

\section{Rank graph Details}\label{sec:rank_graph}
As described in \cref{policy_insights}, we suggest using a rank graph comparison. To perform such analysis, we suggest the following steps. 
Given policies, the threshold for treatment assignment for a given policy is set to be the $q^{th}$ percentile of the CATE value, and the policy value is calculated based on this percentile. I.e., instead of policy $\pi_\mathcal{A}(x_i) = sign(\widehat{\tau_\mathcal{A}}(x_i))$, we define a set of new policies, where each policy is defined according to the $q^{th}$ percentile of the CATE value of policy $\mathcal{A}$: $\pi_\mathcal{A}^{q-rank}(x_i) = \mathbb{I}\left(\widehat{\tau_A}(x_i) > Q(\widehat{\tau_\mathcal{A}},q) \right)$, where $\pi_\mathcal{A}^{q-rank}$ is defined for each $q$, we call this the $q-rank$ policy of $\mathcal{A}$. 
Policy values for each rank-$q$ policy are calculated for $q$ values ranging from $0$ to $1$ (with step size $\delta$), and these values are compared to the random policy and dichotomies policy. The x-axis of the graph is set to be the proportion of patients assigned $T=1$ for each policy and quantile value.

\section{AKI and ADHF}\label{AKI and ADHF}
\textbf{Acute decompensated heart failure (ADHF).} We focus on ADHF as an important use case for learning causal decision support models from healthcare data. ADHF is the most common cause of hospital admission in persons aged $\>$65 years, accounting for over 1,000,000 hospitalizations each year in the U.S. alone\citep{Gheorghiade2009AcuteSyndromes}. The prognosis of patients admitted with ADHF is dismal \citep{Gheorghiade2009AcuteSyndromes}. Seemingly a single homogeneous clinical phenotype presenting with shortness of breath and fluid overload, these patients are a very heterogeneous group in terms of the underlying pathophysiology. This leads to a major challenge in tailoring treatment, as a single uniform approach may prove ineffective or harmful in subgroups of these patients \citep{Laffey2018NegativeWrong.}. Indeed, that may explain the inconclusiveness of virtually all recent randomized controlled trials (RCTs) in ADHF patients and the lack of evidence-based guidelines for their management \citep{McMurray2014AngiotensinNeprilysinFailure, Massie2010RolofyllineFailure, OConnor2011EffectFailure}. \\
\textbf{Acute Kidney Injury (AKI)} is a common and potentially dangerous complication for patients suffering from ADHF. AKI is characterized by a sudden episode of kidney failure or kidney damage that happens within a few hours or a few days. AKI, usually defined as an increase in the presence of the biomarker creatinine on the order of $>0.3 \frac{mg}{dL}$ to $>0.5 \frac{mg}{dL}$ from baseline. It causes a build-up of waste products in the blood and compromises the kidneys' ability to maintain the right balance of fluid in the body. AKI occurs in 20-30\% of patients with ADHF, and is associated with greater length of hospital stay, increased chance of hospital readmission, and death \citep{Goldberg2009TheInfarction,Butler2004RelationshipPatients}. 

\noindent A physician treating an ADHF patient with AKI faces a therapeutic dilemma. On the one hand, ADHF entails fluid accumulation (congestion) requiring diuretic therapy and excessive neurohormonal activation, since sustaining proper kidney function during therapeutic interventions is vital to the alleviation of congestion. On the other hand, these very treatments aimed at alleviating congestion might be a cause or an aggravating factor in AKI, which can harm the patient’s health. This requires a delicate balancing act on the side of the treating physician, a difficulty which is made worse by the fact that there are no guidelines for patients who develop AKI in the setting of ADHF.

\noindent Currently, the prevailing treatments for ADHF patients with AKI are cessation of diuretics and fluids infusion \citep{Boulos2019TreatmentInjury}. In addition, the physician may elect to reduce or discontinue neurohormonal inhibitors (e.g. beta blockers) and to initiate inotrope therapy. The latter decision, although common among physicians treating ADHF, has no proven benefit in the short term and is undoubtedly harmful in the long term \citep{Kane2017Discontinuation/doseFraction,Bohm2011BeneficialTrial}. Therefore, when AKI occurs in an ADHF patient, the treating physician is faced with multiple options to respond by modifying therapy, with no guidance as to how these decisions affect renal and overall clinical outcomes. Indeed, in a recent pilot study of 277 ADHF patients at the Rambam hospital, there was remarkable variability in the observed management of patients with AKI \citep{Boulos2019TreatmentInjury}.

\newpage
\subsection{Data description}\label{apx_sc:table1}
\begin{scriptsize}
\begin{longtable}{lllllll}
\toprule
 &  & \textbf{Missing} & \textbf{Overall} & \textbf{Test} & \textbf{Train} & \textbf{Validation} \\
\midrule
\endfirsthead
\hline
 &  & \textbf{Missing} & \textbf{Overall} & \textbf{Test} & \textbf{Train} & \textbf{Validation}\\
\midrule
\endhead
\hline
\textbf{n} &  &  & 2157 & 530 & 1305 & 322 \\
\cline{1-7}
\multirow{2}{*}{firstadm, n (\%)} & 0 & 0 & 942 (43.7) & 234 (44.2) & 565 (43.3) & 143 (44.4) \\
 & 1 &  & 1215 (56.3) & 296 (55.8) & 740 (56.7) & 179 (55.6) \\
\cline{1-7}
\multirow{9}{*}{admyear, n (\%)} & 1.0 & 0 & 198 (9.2) & 44 (8.3) & 123 (9.4) & 31 (9.6) \\
 & 2.0 &  & 205 (9.5) & 51 (9.6) & 124 (9.5) & 30 (9.3) \\
 & 3.0 &  & 256 (11.9) & 63 (11.9) & 147 (11.3) & 46 (14.3) \\
 & 4.0 &  & 292 (13.5) & 78 (14.7) & 182 (13.9) & 32 (9.9) \\
 & 5.0 &  & 322 (14.9) & 80 (15.1) & 198 (15.2) & 44 (13.7) \\
 & 6.0 &  & 312 (14.5) & 70 (13.2) & 187 (14.3) & 55 (17.1) \\
 & 7.0 &  & 297 (13.8) & 71 (13.4) & 181 (13.9) & 45 (14.0) \\
 & 8.0 &  & 273 (12.7) & 73 (13.8) & 162 (12.4) & 38 (11.8) \\
 & 0.0 &  & 2 (0.1) &  & 1 (0.1) & 1 (0.3) \\
\cline{1-7}
age, mean (SD) &  & 0 & 76.9 (11.5) & 76.2 (12.1) & 77.5 (11.1) & 75.6 (12.2) \\
\cline{1-7}
\multirow{2}{*}{gender, n (\%)} & 0 & 0 & 1140 (52.9) & 263 (49.6) & 715 (54.8) & 162 (50.3) \\
 & 1 &  & 1017 (47.1) & 267 (50.4) & 590 (45.2) & 160 (49.7) \\
\cline{1-7}
Weight, mean (SD) &  & 0 & 80.9 (14.5) & 81.3 (14.5) & 80.5 (14.8) & 82.0 (13.1) \\
\cline{1-7}
Hight, mean (SD) &  & 0 & 164.8 (6.3) & 165.1 (7.4) & 164.6 (6.0) & 165.0 (5.0) \\
\cline{1-7}
bmi, mean (SD) &  & 0 & 30.3 (17.6) & 31.3 (33.3) & 29.9 (7.7) & 30.2 (4.4) \\
\cline{1-7}
\multirow{2}{*}{htn, n (\%)} & 0 & 0 & 610 (28.3) & 149 (28.1) & 365 (28.0) & 96 (29.8) \\
 & 1 &  & 1547 (71.7) & 381 (71.9) & 940 (72.0) & 226 (70.2) \\
\cline{1-7}
firsttemp, mean (SD) &  & 0 & 37.1 (10.3) & 37.4 (14.5) & 36.8 (2.9) & 37.8 (18.4) \\
\cline{1-7}
\multirow{2}{*}{dm, n (\%)} & 0 & 0 & 963 (44.6) & 235 (44.3) & 570 (43.7) & 158 (49.1) \\
 & 1 &  & 1194 (55.4) & 295 (55.7) & 735 (56.3) & 164 (50.9) \\
\cline{1-7}
\multirow{3}{*}{smk, n (\%)} & 0 & 0 & 1637 (75.9) & 405 (76.4) & 994 (76.2) & 238 (73.9) \\
 & 1 &  & 285 (13.2) & 63 (11.9) & 171 (13.1) & 51 (15.8) \\
 & 2 &  & 235 (10.9) & 62 (11.7) & 140 (10.7) & 33 (10.2) \\
\cline{1-7}
\multirow{2}{*}{ihd, n (\%)} & 0 & 0 & 1605 (74.4) & 405 (76.4) & 955 (73.2) & 245 (76.1) \\
 & 1 &  & 552 (25.6) & 125 (23.6) & 350 (26.8) & 77 (23.9) \\
\cline{1-7}
\multirow{2}{*}{vhd, n (\%)} & 0 & 0 & 1859 (86.2) & 458 (86.4) & 1125 (86.2) & 276 (85.7) \\
 & 1 &  & 298 (13.8) & 72 (13.6) & 180 (13.8) & 46 (14.3) \\
\cline{1-7}
\multirow{2}{*}{af, n (\%)} & 0 & 0 & 1204 (55.8) & 308 (58.1) & 719 (55.1) & 177 (55.0) \\
 & 1 &  & 953 (44.2) & 222 (41.9) & 586 (44.9) & 145 (45.0) \\
\cline{1-7}
\multirow{2}{*}{Hyperlipidemia, n (\%)} & 0 & 0 & 882 (40.9) & 211 (39.8) & 542 (41.5) & 129 (40.1) \\
 & 1 &  & 1275 (59.1) & 319 (60.2) & 763 (58.5) & 193 (59.9) \\
\cline{1-7}
\multirow{2}{*}{copd, n (\%)} & 0 & 0 & 1841 (85.4) & 453 (85.5) & 1109 (85.0) & 279 (86.6) \\
 & 1 &  & 316 (14.6) & 77 (14.5) & 196 (15.0) & 43 (13.4) \\
\cline{1-7}
\multirow{2}{*}{crf, n (\%)} & 0 & 0 & 1171 (54.3) & 294 (55.5) & 697 (53.4) & 180 (55.9) \\
 & 1 &  & 986 (45.7) & 236 (44.5) & 608 (46.6) & 142 (44.1) \\
\cline{1-7}
\multirow{2}{*}{bb, n (\%)} & 0 & 0 & 547 (25.4) & 131 (24.7) & 319 (24.4) & 97 (30.1) \\
 & 1 &  & 1610 (74.6) & 399 (75.3) & 986 (75.6) & 225 (69.9) \\
\cline{1-7}
\multirow{2}{*}{acei, n (\%)} & 0 & 0 & 733 (34.0) & 179 (33.8) & 445 (34.1) & 109 (33.9) \\
 & 1 &  & 1424 (66.0) & 351 (66.2) & 860 (65.9) & 213 (66.1) \\
\cline{1-7}
\multirow{2}{*}{arf, n (\%)} & 0 & 0 & 1298 (60.2) & 312 (58.9) & 796 (61.0) & 190 (59.0) \\
 & 1 &  & 859 (39.8) & 218 (41.1) & 509 (39.0) & 132 (41.0) \\
\cline{1-7}
\multirow{2}{*}{antiplt, n (\%)} & 0 & 0 & 1315 (61.0) & 317 (59.8) & 801 (61.4) & 197 (61.2) \\
 & 1 &  & 842 (39.0) & 213 (40.2) & 504 (38.6) & 125 (38.8) \\
\cline{1-7}
\multirow{2}{*}{anticoagulants, n (\%)} & 0 & 0 & 1547 (71.7) & 394 (74.3) & 932 (71.4) & 221 (68.6) \\
 & 1 &  & 610 (28.3) & 136 (25.7) & 373 (28.6) & 101 (31.4) \\
\cline{1-7}
\multirow{2}{*}{furosemide, n (\%)} & 0 & 0 & 623 (28.9) & 159 (30.0) & 371 (28.4) & 93 (28.9) \\
 & 1 &  & 1534 (71.1) & 371 (70.0) & 934 (71.6) & 229 (71.1) \\
\cline{1-7}
\multirow{2}{*}{zaroxolin, n (\%)} & 0 & 0 & 2098 (97.3) & 511 (96.4) & 1278 (97.9) & 309 (96.0) \\
 & 1 &  & 59 (2.7) & 19 (3.6) & 27 (2.1) & 13 (4.0) \\
\cline{1-7}
\multirow{2}{*}{thiamin, n (\%)} & 0 & 0 & 2104 (97.5) & 518 (97.7) & 1272 (97.5) & 314 (97.5) \\
 & 1 &  & 53 (2.5) & 12 (2.3) & 33 (2.5) & 8 (2.5) \\
\cline{1-7}
\multirow{2}{*}{insulin, n (\%)} & 0 & 0 & 1672 (77.5) & 410 (77.4) & 1012 (77.5) & 250 (77.6) \\
 & 1 &  & 485 (22.5) & 120 (22.6) & 293 (22.5) & 72 (22.4) \\
\cline{1-7}
tn, mean (SD) &  & 0 & 0.3 (1.9) & 0.3 (1.2) & 0.4 (2.1) & 0.4 (2.1) \\
\cline{1-7}
hb, mean (SD) &  & 0 & 11.3 (1.9) & 11.4 (2.0) & 11.3 (1.9) & 11.4 (2.1) \\
\cline{1-7}
first\_mcv, mean (SD) &  & 0 & 86.6 (7.0) & 86.7 (7.4) & 86.6 (6.8) & 86.5 (6.9) \\
\cline{1-7}
first\_rdw, mean (SD) &  & 0 & 15.9 (2.0) & 15.9 (2.0) & 15.9 (2.0) & 15.8 (1.9) \\
\cline{1-7}
wbc, mean (SD) &  & 0 & 10.9 (9.2) & 10.8 (11.6) & 11.0 (8.8) & 10.6 (5.9) \\
\cline{1-7}
alb, mean (SD) &  & 0 & 3.1 (0.5) & 3.1 (0.5) & 3.1 (0.5) & 3.2 (0.5) \\
\cline{1-7}
ast, mean (SD) &  & 0 & 55.2 (359.1) & 46.4 (164.4) & 57.3 (428.6) & 61.0 (274.0) \\
\cline{1-7}
alt, mean (SD) &  & 0 & 62.9 (250.8) & 71.1 (296.6) & 60.0 (241.6) & 61.2 (201.1) \\
\cline{1-7}
ggt, mean (SD) &  & 0 & 98.5 (141.6) & 98.1 (124.8) & 100.7 (157.0) & 90.3 (93.7) \\
\cline{1-7}
alkp, mean (SD) &  & 0 & 113.5 (72.0) & 110.7 (58.8) & 115.6 (79.7) & 109.8 (56.9) \\
\cline{1-7}
first\_p, mean (SD) &  & 0 & 4.4 (1.2) & 4.4 (1.2) & 4.4 (1.2) & 4.4 (1.3) \\
\cline{1-7}
crp, mean (SD) &  & 0 & 46.8 (28.9) & 47.5 (32.0) & 46.4 (27.5) & 47.1 (28.7) \\
\cline{1-7}
first\_cl, mean (SD) &  & 0 & 102.5 (5.2) & 102.3 (5.4) & 102.6 (5.2) & 102.8 (5.0) \\
\cline{1-7}
glu, mean (SD) &  & 0 & 174.3 (83.0) & 174.0 (83.7) & 174.4 (81.8) & 174.5 (86.6) \\
\cline{1-7}
dbp\_last, mean (SD) &  & 0 & 64.4 (14.6) & 64.7 (14.7) & 64.2 (14.5) & 64.6 (15.1) \\
\cline{1-7}
dbp\_first, mean (SD) &  & 0 & 76.3 (16.9) & 76.1 (16.3) & 76.5 (17.3) & 75.9 (16.3) \\
\cline{1-7}
dbp\_mean, mean (SD) &  & 0 & 68.8 (10.9) & 68.9 (10.8) & 68.9 (10.9) & 68.5 (11.5) \\
\cline{1-7}
dbp\_median, mean (SD) &  & 0 & 68.2 (11.6) & 68.3 (11.1) & 68.2 (11.6) & 68.0 (12.2) \\
\cline{1-7}
dbp\_quantile\_90, mean (SD) &  & 0 & 80.8 (13.2) & 80.6 (12.6) & 81.1 (13.4) & 79.9 (13.6) \\
\cline{1-7}
dbp\_quantile\_10, mean (SD) &  & 0 & 57.3 (11.0) & 57.7 (11.1) & 57.2 (10.7) & 57.2 (11.6) \\
\cline{1-7}
dbp\_std, mean (SD) &  & 0 & 11.6 (4.9) & 11.3 (4.6) & 11.9 (5.1) & 11.1 (4.3) \\
\cline{1-7}
sbp\_last, mean (SD) &  & 0 & 123.0 (23.7) & 123.8 (24.7) & 123.0 (23.6) & 122.1 (22.5) \\
\cline{1-7}
sbp\_first, mean (SD) &  & 0 & 145.6 (32.4) & 145.3 (31.2) & 146.0 (33.4) & 144.6 (30.1) \\
\cline{1-7}
sbp\_mean, mean (SD) &  & 0 & 131.7 (21.2) & 132.3 (21.8) & 131.9 (21.0) & 130.0 (21.1) \\
\cline{1-7}
sbp\_median, mean (SD) &  & 0 & 130.7 (22.0) & 131.1 (22.5) & 131.0 (21.8) & 129.0 (21.9) \\
\cline{1-7}
sbp\_quantile\_90, mean (SD) &  & 0 & 151.0 (26.4) & 150.9 (26.2) & 151.7 (26.6) & 147.8 (25.8) \\
\cline{1-7}
sbp\_quantile\_10, mean (SD) &  & 0 & 113.2 (19.6) & 114.3 (20.3) & 113.0 (19.4) & 112.3 (19.6) \\
\cline{1-7}
sbp\_std, mean (SD) &  & 0 & 18.8 (9.3) & 18.1 (7.8) & 19.3 (9.1) & 18.2 (12.1) \\
\cline{1-7}
pp\_last, mean (SD) &  & 0 & 58.6 (19.8) & 59.0 (20.4) & 58.8 (19.8) & 57.4 (18.6) \\
\cline{1-7}
pp\_first, mean (SD) &  & 0 & 69.3 (24.6) & 69.1 (23.8) & 69.6 (25.3) & 68.7 (23.1) \\
\cline{1-7}
pp\_mean, mean (SD) &  & 0 & 62.9 (17.2) & 63.4 (17.5) & 63.1 (17.1) & 61.5 (16.9) \\
\cline{1-7}
pp\_median, mean (SD) &  & 0 & 62.3 (18.0) & 62.6 (18.4) & 62.5 (17.9) & 60.7 (17.4) \\
\cline{1-7}
pp\_quantile\_90, mean (SD) &  & 0 & 77.2 (20.7) & 77.3 (20.9) & 77.6 (20.8) & 75.3 (20.3) \\
\cline{1-7}
pp\_quantile\_10, mean (SD) &  & 0 & 49.2 (15.8) & 50.0 (16.1) & 49.2 (15.8) & 48.0 (15.7) \\
\cline{1-7}
pp\_std, mean (SD) &  & 0 & 14.1 (7.5) & 13.6 (5.9) & 14.2 (6.8) & 14.1 (11.4) \\
\cline{1-7}
MAP\_last, mean (SD) &  & 0 & 83.9 (15.6) & 84.4 (15.9) & 83.8 (15.4) & 83.8 (15.7) \\
\cline{1-7}
MAP\_first, mean (SD) &  & 0 & 99.4 (20.1) & 99.2 (19.4) & 99.6 (20.7) & 98.8 (19.0) \\
\cline{1-7}
MAP\_mean, mean (SD) &  & 0 & 89.8 (12.8) & 90.0 (13.0) & 89.9 (12.7) & 89.0 (13.2) \\
\cline{1-7}
MAP\_median, mean (SD) &  & 0 & 89.2 (13.4) & 89.5 (13.3) & 89.2 (13.3) & 88.5 (14.0) \\
\cline{1-7}
MAP\_quantile\_90, mean (SD) &  & 0 & 103.0 (15.9) & 103.0 (15.5) & 103.4 (16.1) & 101.4 (15.8) \\
\cline{1-7}
MAP\_quantile\_10, mean (SD) &  & 0 & 77.1 (12.3) & 77.6 (12.7) & 76.9 (12.1) & 76.7 (12.6) \\
\cline{1-7}
MAP\_std, mean (SD) &  & 0 & 12.8 (5.6) & 12.4 (5.1) & 13.2 (5.9) & 12.3 (5.5) \\
\cline{1-7}
hr\_last, mean (SD) &  & 0 & 79.3 (18.6) & 79.7 (18.6) & 78.7 (17.9) & 80.9 (21.0) \\
\cline{1-7}
hr\_first, mean (SD) &  & 0 & 85.8 (20.7) & 85.8 (20.7) & 85.4 (20.4) & 87.5 (21.7) \\
\cline{1-7}
hr\_mean, mean (SD) &  & 0 & 80.9 (15.0) & 81.1 (14.7) & 80.7 (14.9) & 81.5 (15.9) \\
\cline{1-7}
hr\_median, mean (SD) &  & 0 & 80.1 (15.9) & 80.4 (15.7) & 79.8 (15.8) & 81.0 (16.8) \\
\cline{1-7}
hr\_quantile\_90, mean (SD) &  & 0 & 92.2 (19.0) & 92.6 (18.9) & 91.9 (18.5) & 93.3 (20.8) \\
\cline{1-7}
hr\_quantile\_10, mean (SD) &  & 0 & 70.3 (13.8) & 70.4 (13.3) & 70.3 (13.8) & 70.5 (14.5) \\
\cline{1-7}
hr\_std, mean (SD) &  & 0 & 11.1 (6.6) & 11.3 (6.6) & 11.1 (6.3) & 11.2 (7.5) \\
\cline{1-7}
bnp\_last, mean (SD) &  & 0 & 1224.2 (882.1) & 1251.6 (930.5) & 1219.5 (868.7) & 1198.4 (855.5) \\
\cline{1-7}
bnp\_first, mean (SD) &  & 0 & 1224.7 (878.8) & 1244.7 (921.9) & 1222.7 (866.0) & 1199.6 (859.5) \\
\cline{1-7}
bnp\_mean, mean (SD) &  & 0 & 1222.5 (879.3) & 1246.3 (925.4) & 1219.1 (865.7) & 1197.1 (857.3) \\
\cline{1-7}
bnp\_median, mean (SD) &  & 0 & 1222.5 (879.3) & 1246.3 (925.4) & 1219.1 (865.7) & 1197.1 (857.3) \\
\cline{1-7}
bnp\_std, mean (SD) &  & 0 & 246.9 (59.2) & 246.5 (50.6) & 247.4 (67.9) & 245.5 (23.9) \\
\cline{1-7}
bnp\_quantile\_90, mean (SD) &  & 0 & 1231.8 (881.9) & 1255.2 (928.6) & 1228.8 (868.8) & 1205.6 (857.1) \\
\cline{1-7}
bnp\_quantile\_10, mean (SD) &  & 0 & 1216.4 (878.0) & 1240.6 (923.0) & 1212.7 (864.5) & 1191.8 (857.5) \\
\cline{1-7}
bun\_last, mean (SD) &  & 0 & 41.0 (23.7) & 40.8 (23.7) & 40.7 (23.3) & 42.2 (25.5) \\
\cline{1-7}
bun\_first, mean (SD) &  & 0 & 38.7 (23.4) & 38.2 (23.2) & 38.8 (23.2) & 39.2 (24.7) \\
\cline{1-7}
bun\_mean, mean (SD) &  & 0 & 39.6 (23.1) & 39.2 (23.0) & 39.5 (22.8) & 40.5 (24.7) \\
\cline{1-7}
bun\_median, mean (SD) &  & 0 & 39.4 (23.1) & 39.0 (23.1) & 39.4 (22.8) & 40.3 (24.7) \\
\cline{1-7}
bun\_std, mean (SD) &  & 0 & 4.0 (3.1) & 4.1 (3.1) & 4.0 (3.2) & 4.0 (2.8) \\
\cline{1-7}
bun\_quantile\_90, mean (SD) &  & 0 & 41.8 (23.8) & 41.4 (23.6) & 41.6 (23.5) & 42.7 (25.3) \\
\cline{1-7}
bun\_quantile\_10, mean (SD) &  & 0 & 37.5 (22.9) & 37.0 (22.8) & 37.5 (22.5) & 38.3 (24.5) \\
\cline{1-7}
crea\_last, mean (SD) &  & 0 & 1.8 (1.0) & 1.8 (1.0) & 1.8 (1.0) & 1.8 (1.2) \\
\cline{1-7}
crea\_first, mean (SD) &  & 0 & 1.8 (1.0) & 1.8 (1.1) & 1.7 (1.0) & 1.8 (1.1) \\
\cline{1-7}
crea\_mean, mean (SD) &  & 0 & 1.8 (1.0) & 1.8 (1.0) & 1.8 (1.0) & 1.8 (1.1) \\
\cline{1-7}
crea\_median, mean (SD) &  & 0 & 1.8 (1.0) & 1.8 (1.0) & 1.8 (1.0) & 1.8 (1.1) \\
\cline{1-7}
crea\_std, mean (SD) &  & 0 & 0.1 (0.1) & 0.1 (0.1) & 0.1 (0.1) & 0.1 (0.1) \\
\cline{1-7}
crea\_quantile\_90, mean (SD) &  & 0 & 1.8 (1.0) & 1.8 (1.1) & 1.8 (1.0) & 1.9 (1.1) \\
\cline{1-7}
crea\_quantile\_10, mean (SD) &  & 0 & 1.7 (1.0) & 1.7 (1.0) & 1.7 (1.0) & 1.7 (1.1) \\
\cline{1-7}
hct\_last, mean (SD) &  & 0 & 34.2 (5.7) & 34.4 (6.1) & 34.0 (5.5) & 34.5 (6.2) \\
\cline{1-7}
hct\_first, mean (SD) &  & 0 & 34.4 (5.7) & 34.6 (6.0) & 34.3 (5.5) & 34.7 (6.3) \\
\cline{1-7}
hct\_mean, mean (SD) &  & 0 & 34.2 (5.5) & 34.5 (5.8) & 34.1 (5.2) & 34.5 (5.9) \\
\cline{1-7}
hct\_median, mean (SD) &  & 0 & 34.2 (5.5) & 34.4 (5.8) & 34.1 (5.3) & 34.4 (6.0) \\
\cline{1-7}
hct\_std, mean (SD) &  & 0 & 1.9 (1.3) & 1.8 (1.2) & 1.9 (1.2) & 1.9 (1.4) \\
\cline{1-7}
hct\_quantile\_90, mean (SD) &  & 0 & 35.3 (5.5) & 35.5 (5.8) & 35.1 (5.2) & 35.6 (5.9) \\
\cline{1-7}
hct\_quantile\_10, mean (SD) &  & 0 & 33.2 (5.8) & 33.5 (6.0) & 33.0 (5.5) & 33.3 (6.3) \\
\cline{1-7}
k\_last, mean (SD) &  & 0 & 4.2 (0.6) & 4.2 (0.6) & 4.2 (0.7) & 4.1 (0.6) \\
\cline{1-7}
k\_first, mean (SD) &  & 0 & 4.4 (0.7) & 4.4 (0.7) & 4.4 (0.6) & 4.3 (0.6) \\
\cline{1-7}
k\_mean, mean (SD) &  & 0 & 4.3 (0.6) & 4.3 (0.6) & 4.3 (0.6) & 4.2 (0.6) \\
\cline{1-7}
k\_median, mean (SD) &  & 0 & 4.2 (0.6) & 4.3 (0.6) & 4.2 (0.6) & 4.2 (0.6) \\
\cline{1-7}
k\_std, mean (SD) &  & 0 & 0.3 (0.2) & 0.3 (0.2) & 0.3 (0.2) & 0.3 (0.2) \\
\cline{1-7}
k\_quantile\_90, mean (SD) &  & 0 & 4.5 (0.6) & 4.5 (0.7) & 4.5 (0.6) & 4.5 (0.6) \\
\cline{1-7}
k\_quantile\_10, mean (SD) &  & 0 & 4.0 (0.6) & 4.0 (0.6) & 4.0 (0.6) & 4.0 (0.6) \\
\cline{1-7}
na\_last, mean (SD) &  & 0 & 138.0 (4.8) & 137.7 (4.7) & 138.0 (4.8) & 138.3 (5.0) \\
\cline{1-7}
na\_first, mean (SD) &  & 0 & 136.9 (4.9) & 136.8 (4.9) & 136.9 (4.9) & 137.0 (4.6) \\
\cline{1-7}
na\_mean, mean (SD) &  & 0 & 137.5 (4.3) & 137.4 (4.2) & 137.6 (4.3) & 137.7 (4.3) \\
\cline{1-7}
na\_median, mean (SD) &  & 0 & 137.6 (4.3) & 137.4 (4.3) & 137.6 (4.4) & 137.7 (4.3) \\
\cline{1-7}
na\_std, mean (SD) &  & 0 & 2.2 (1.3) & 2.2 (1.3) & 2.2 (1.3) & 2.2 (1.4) \\
\cline{1-7}
na\_quantile\_90, mean (SD) &  & 0 & 139.3 (4.4) & 139.1 (4.3) & 139.3 (4.4) & 139.5 (4.6) \\
\cline{1-7}
na\_quantile\_10, mean (SD) &  & 0 & 135.8 (4.6) & 135.6 (4.7) & 135.8 (4.6) & 136.0 (4.6) \\
\cline{1-7}
EGFR\_last, mean (SD) &  & 0 & 40.7 (21.4) & 40.6 (20.9) & 40.3 (21.0) & 42.3 (23.4) \\
\cline{1-7}
EGFR\_first, mean (SD) &  & 0 & 42.5 (22.4) & 42.8 (22.2) & 42.0 (21.9) & 44.2 (24.2) \\
\cline{1-7}
EGFR\_mean, mean (SD) &  & 0 & 42.0 (21.8) & 42.3 (21.7) & 41.5 (21.3) & 43.7 (23.6) \\
\cline{1-7}
EGFR\_median, mean (SD) &  & 0 & 41.9 (21.8) & 42.3 (21.8) & 41.4 (21.3) & 43.5 (23.6) \\
\cline{1-7}
EGFR\_quantile\_90, mean (SD) &  & 0 & 44.1 (23.2) & 44.5 (23.5) & 43.5 (22.6) & 45.9 (25.0) \\
\cline{1-7}
EGFR\_quantile\_10, mean (SD) &  & 0 & 40.0 (20.7) & 40.1 (20.5) & 39.5 (20.4) & 41.5 (22.4) \\
\cline{1-7}
EGFR\_std, mean (SD) &  & 0 & 3.8 (3.3) & 4.0 (4.0) & 3.8 (3.0) & 3.9 (3.4) \\
\cline{1-7}
crea\_d\_point, mean (SD) &  & 0 & 2.3 (1.1) & 2.4 (1.2) & 2.3 (1.1) & 2.4 (1.2) \\
\cline{1-7}
crea\_diff, mean (SD) &  & 0 & 0.6 (0.4) & 0.6 (0.4) & 0.6 (0.4) & 0.6 (0.4) \\
\cline{1-7}
time\_to\_treat, mean (SD) &  & 0 & 7.7 (9.7) & 8.1 (9.9) & 7.4 (9.7) & 8.1 (9.6) \\
\cline{1-7}
\multirow{2}{*}{1\_ind, n (\%)} & 1.0 & 0 & 2156 (100.0) & 530 (100.0) & 1304 (99.9) & 322 (100.0) \\
 & 0.0 &  & 1 (0.0) &  & 1 (0.1) &  \\
\cline{1-7}
\multirow{2}{*}{2\_ind, n (\%)} & 0.0 & 0 & 449 (20.8) & 100 (18.9) & 281 (21.5) & 68 (21.1) \\
 & 1.0 &  & 1708 (79.2) & 430 (81.1) & 1024 (78.5) & 254 (78.9) \\
\cline{1-7}
\multirow{2}{*}{3\_ind, n (\%)} & 0.0 & 0 & 1198 (55.5) & 290 (54.7) & 729 (55.9) & 179 (55.6) \\
 & 1.0 &  & 959 (44.5) & 240 (45.3) & 576 (44.1) & 143 (44.4) \\
\cline{1-7}
\multirow{2}{*}{4\_ind, n (\%)} & 0.0 & 0 & 1782 (82.6) & 440 (83.0) & 1071 (82.1) & 271 (84.2) \\
 & 1.0 &  & 375 (17.4) & 90 (17.0) & 234 (17.9) & 51 (15.8) \\
\cline{1-7}
\multirow{2}{*}{5\_ind, n (\%)} & 0.0 & 0 & 1736 (80.5) & 419 (79.1) & 1069 (81.9) & 248 (77.0) \\
 & 1.0 &  & 421 (19.5) & 111 (20.9) & 236 (18.1) & 74 (23.0) \\
\cline{1-7}
\multirow{2}{*}{6\_ind, n (\%)} & 0.0 & 0 & 1996 (92.5) & 490 (92.5) & 1211 (92.8) & 295 (91.6) \\
 & 1.0 &  & 161 (7.5) & 40 (7.5) & 94 (7.2) & 27 (8.4) \\
\cline{1-7}
\multirow{2}{*}{7\_ind, n (\%)} & 0.0 & 0 & 2143 (99.4) & 528 (99.6) & 1298 (99.5) & 317 (98.4) \\
 & 1.0 &  & 14 (0.6) & 2 (0.4) & 7 (0.5) & 5 (1.6) \\
\cline{1-7}
\multirow{2}{*}{8\_ind, n (\%)} & 0.0 & 0 & 2148 (99.6) & 529 (99.8) & 1299 (99.5) & 320 (99.4) \\
 & 1.0 &  & 9 (0.4) & 1 (0.2) & 6 (0.5) & 2 (0.6) \\
\cline{1-7}
\multirow{2}{*}{1\_ind\_hosp, n (\%)} & 0.0 & 0 & 524 (24.3) & 124 (23.4) & 320 (24.5) & 80 (24.8) \\
 & 1.0 &  & 1633 (75.7) & 406 (76.6) & 985 (75.5) & 242 (75.2) \\
\cline{1-7}
\multirow{2}{*}{2\_ind\_hosp, n (\%)} & 0.0 & 0 & 524 (24.3) & 125 (23.6) & 304 (23.3) & 95 (29.5) \\
 & 1.0 &  & 1633 (75.7) & 405 (76.4) & 1001 (76.7) & 227 (70.5) \\
\cline{1-7}
\multirow{2}{*}{3\_ind\_hosp, n (\%)} & 0.0 & 0 & 1226 (56.8) & 296 (55.8) & 745 (57.1) & 185 (57.5) \\
 & 1.0 &  & 931 (43.2) & 234 (44.2) & 560 (42.9) & 137 (42.5) \\
\cline{1-7}
\multirow{2}{*}{4\_ind\_hosp, n (\%)} & 0.0 & 0 & 1674 (77.6) & 409 (77.2) & 1008 (77.2) & 257 (79.8) \\
 & 1.0 &  & 483 (22.4) & 121 (22.8) & 297 (22.8) & 65 (20.2) \\
\cline{1-7}
\multirow{2}{*}{5\_ind\_hosp, n (\%)} & 0.0 & 0 & 1781 (82.6) & 441 (83.2) & 1083 (83.0) & 257 (79.8) \\
 & 1.0 &  & 376 (17.4) & 89 (16.8) & 222 (17.0) & 65 (20.2) \\
\cline{1-7}
\multirow{2}{*}{6\_ind\_hosp, n (\%)} & 0.0 & 0 & 2153 (99.8) & 528 (99.6) & 1303 (99.8) & 322 (100.0) \\
 & 1.0 &  & 4 (0.2) & 2 (0.4) & 2 (0.2) &  \\
\cline{1-7}
\multirow{2}{*}{7\_ind\_hosp, n (\%)} & 0.0 & 0 & 2122 (98.4) & 526 (99.2) & 1287 (98.6) & 309 (96.0) \\
 & 1.0 &  & 35 (1.6) & 4 (0.8) & 18 (1.4) & 13 (4.0) \\
\cline{1-7}
\multirow{2}{*}{8\_ind\_hosp, n (\%)} & 0.0 & 0 & 2140 (99.2) & 527 (99.4) & 1294 (99.2) & 319 (99.1) \\
 & 1.0 &  & 17 (0.8) & 3 (0.6) & 11 (0.8) & 3 (0.9) \\
\cline{1-7}
\multirow{4}{*}{returning\_patient, n (\%)} & 0.0 & 0 & 1215 (56.3) & 296 (55.8) & 740 (56.7) & 179 (55.6) \\
 & 1.0 &  & 756 (35.0) & 192 (36.2) & 453 (34.7) & 111 (34.5) \\
 & 2.0 &  & 157 (7.3) & 35 (6.6) & 97 (7.4) & 25 (7.8) \\
 & 3.0 &  & 29 (1.3) & 7 (1.3) & 15 (1.1) & 7 (2.2) \\
\cline{1-7}
relative\_date\_first\_tn, mean (SD) &  & 0 & 0.6 (2.2) & 0.6 (2.3) & 0.5 (2.0) & 0.7 (3.0) \\
\cline{1-7}
\multirow{7}{*}{relative\_date\_first\_hb, n (\%)} & 1.0 & 0 & 90 (4.2) & 22 (4.2) & 49 (3.8) & 19 (5.9) \\
 & 0 &  & 2044 (94.8) & 501 (94.5) & 1242 (95.2) & 301 (93.5) \\
 & 2.0 &  & 16 (0.7) & 6 (1.1) & 9 (0.7) & 1 (0.3) \\
 & 3.0 &  & 4 (0.2) & 1 (0.2) & 3 (0.2) &  \\
 & 12.0 &  & 1 (0.0) &  & 1 (0.1) &  \\
 & 8.0 &  & 1 (0.0) &  & 1 (0.1) &  \\
 & 6.0 &  & 1 (0.0) &  &  & 1 (0.3) \\
\cline{1-7}
\multirow{7}{*}{relative\_date\_first\_mcv, n (\%)} & 0.0 & 0 & 2041 (94.6) & 499 (94.2) & 1241 (95.1) & 301 (93.5) \\
 & 1.0 &  & 90 (4.2) & 22 (4.2) & 49 (3.8) & 19 (5.9) \\
 & 2.0 &  & 19 (0.9) & 8 (1.5) & 10 (0.8) & 1 (0.3) \\
 & 3.0 &  & 4 (0.2) & 1 (0.2) & 3 (0.2) &  \\
 & 12.0 &  & 1 (0.0) &  & 1 (0.1) &  \\
 & 8.0 &  & 1 (0.0) &  & 1 (0.1) &  \\
 & 6.0 &  & 1 (0.0) &  &  & 1 (0.3) \\
\cline{1-7}
relative\_date\_first\_rdw, mean (SD) &  & 0 & 1.5 (1.8) & 1.5 (1.7) & 1.6 (1.8) & 1.5 (1.6) \\
\cline{1-7}
\multirow{7}{*}{relative\_date\_first\_wbc, n (\%)} & 0.0 & 0 & 2041 (94.6) & 499 (94.2) & 1241 (95.1) & 301 (93.5) \\
 & 1.0 &  & 90 (4.2) & 22 (4.2) & 49 (3.8) & 19 (5.9) \\
 & 2.0 &  & 19 (0.9) & 8 (1.5) & 10 (0.8) & 1 (0.3) \\
 & 3.0 &  & 4 (0.2) & 1 (0.2) & 3 (0.2) &  \\
 & 12.0 &  & 1 (0.0) &  & 1 (0.1) &  \\
 & 8.0 &  & 1 (0.0) &  & 1 (0.1) &  \\
 & 6.0 &  & 1 (0.0) &  &  & 1 (0.3) \\
\cline{1-7}
relative\_date\_first\_alb, mean (SD) &  & 0 & 2.0 (1.6) & 1.9 (1.3) & 2.1 (1.7) & 2.0 (1.4) \\
\cline{1-7}
relative\_date\_first\_ast, mean (SD) &  & 0 & 1.6 (1.6) & 1.7 (1.7) & 1.7 (1.6) & 1.5 (1.4) \\
\cline{1-7}
relative\_date\_first\_alt, mean (SD) &  & 0 & 2.1 (1.8) & 2.1 (1.8) & 2.2 (2.0) & 2.0 (1.4) \\
\cline{1-7}
relative\_date\_first\_ggt, mean (SD) &  & 0 & 2.2 (1.9) & 2.2 (2.0) & 2.2 (2.0) & 2.0 (1.4) \\
\cline{1-7}
relative\_date\_first\_alkp, mean (SD) &  & 0 & 2.1 (1.8) & 2.1 (1.8) & 2.2 (2.0) & 2.1 (1.4) \\
\cline{1-7}
relative\_date\_first\_p, mean (SD) &  & 0 & 2.1 (1.7) & 2.1 (1.6) & 2.1 (1.7) & 2.2 (1.7) \\
\cline{1-7}
relative\_date\_first\_crp, mean (SD) &  & 0 & 3.5 (4.9) & 3.7 (6.8) & 3.4 (3.9) & 3.6 (5.0) \\
\cline{1-7}
relative\_date\_first\_cl, mean (SD) &  & 0 & 0.6 (1.1) & 0.6 (1.2) & 0.6 (1.1) & 0.6 (1.1) \\
\cline{1-7}
\multirow{6}{*}{relative\_date\_first\_glu, n (\%)} & 0.0 & 0 & 2052 (95.1) & 503 (94.9) & 1246 (95.5) & 303 (94.1) \\
 & 1.0 &  & 86 (4.0) & 23 (4.3) & 46 (3.5) & 17 (5.3) \\
 & 2.0 &  & 12 (0.6) & 3 (0.6) & 8 (0.6) & 1 (0.3) \\
 & 3.0 &  & 4 (0.2) & 1 (0.2) & 3 (0.2) &  \\
 & 6.0 &  & 2 (0.1) &  & 1 (0.1) & 1 (0.3) \\
 & 8.0 &  & 1 (0.0) &  & 1 (0.1) &  \\
\cline{1-7}
d\_point\_relative\_date, mean (SD) &  & 0 & 3.9 (4.3) & 4.0 (4.7) & 3.8 (3.8) & 4.2 (5.1) \\
\cline{1-7}
\multirow{2}{*}{Weight\_missing, n (\%)} & 0 & 0 & 1041 (48.3) & 270 (50.9) & 616 (47.2) & 155 (48.1) \\
 & 1 &  & 1116 (51.7) & 260 (49.1) & 689 (52.8) & 167 (51.9) \\
\cline{1-7}
\multirow{2}{*}{Hight\_missing, n (\%)} & 0 & 0 & 497 (23.0) & 122 (23.0) & 293 (22.5) & 82 (25.5) \\
 & 1 &  & 1660 (77.0) & 408 (77.0) & 1012 (77.5) & 240 (74.5) \\
\cline{1-7}
\multirow{2}{*}{bmi\_missing, n (\%)} & 0 & 0 & 496 (23.0) & 121 (22.8) & 293 (22.5) & 82 (25.5) \\
 & 1 &  & 1661 (77.0) & 409 (77.2) & 1012 (77.5) & 240 (74.5) \\
\cline{1-7}
\multirow{2}{*}{firsttemp\_missing, n (\%)} & 0 & 0 & 2153 (99.8) & 529 (99.8) & 1302 (99.8) & 322 (100.0) \\
 & 1 &  & 4 (0.2) & 1 (0.2) & 3 (0.2) &  \\
\cline{1-7}
\multirow{2}{*}{tn\_missing, n (\%)} & 0 & 0 & 1862 (86.3) & 454 (85.7) & 1131 (86.7) & 277 (86.0) \\
 & 1 &  & 295 (13.7) & 76 (14.3) & 174 (13.3) & 45 (14.0) \\
\cline{1-7}
\multirow{2}{*}{hb\_missing, n (\%)} & 0 & 0 & 2156 (100.0) & 530 (100.0) & 1304 (99.9) & 322 (100.0) \\
 & 1 &  & 1 (0.0) &  & 1 (0.1) &  \\
\cline{1-7}
\multirow{2}{*}{first\_mcv\_missing, n (\%)} & 0 & 0 & 2156 (100.0) & 530 (100.0) & 1304 (99.9) & 322 (100.0) \\
 & 1 &  & 1 (0.0) &  & 1 (0.1) &  \\
\cline{1-7}
\multirow{2}{*}{first\_rdw\_missing, n (\%)} & 0 & 0 & 2080 (96.4) & 516 (97.4) & 1254 (96.1) & 310 (96.3) \\
 & 1 &  & 77 (3.6) & 14 (2.6) & 51 (3.9) & 12 (3.7) \\
\cline{1-7}
\multirow{2}{*}{wbc\_missing, n (\%)} & 0 & 0 & 2156 (100.0) & 530 (100.0) & 1304 (99.9) & 322 (100.0) \\
 & 1 &  & 1 (0.0) &  & 1 (0.1) &  \\
\cline{1-7}
\multirow{2}{*}{alb\_missing, n (\%)} & 0 & 0 & 2084 (96.6) & 510 (96.2) & 1263 (96.8) & 311 (96.6) \\
 & 1 &  & 73 (3.4) & 20 (3.8) & 42 (3.2) & 11 (3.4) \\
\cline{1-7}
\multirow{2}{*}{ast\_missing, n (\%)} & 0 & 0 & 2073 (96.1) & 508 (95.8) & 1254 (96.1) & 311 (96.6) \\
 & 1 &  & 84 (3.9) & 22 (4.2) & 51 (3.9) & 11 (3.4) \\
\cline{1-7}
\multirow{2}{*}{alt\_missing, n (\%)} & 0 & 0 & 2048 (94.9) & 507 (95.7) & 1236 (94.7) & 305 (94.7) \\
 & 1 &  & 109 (5.1) & 23 (4.3) & 69 (5.3) & 17 (5.3) \\
\cline{1-7}
\multirow{2}{*}{ggt\_missing, n (\%)} & 0 & 0 & 2040 (94.6) & 503 (94.9) & 1232 (94.4) & 305 (94.7) \\
 & 1 &  & 117 (5.4) & 27 (5.1) & 73 (5.6) & 17 (5.3) \\
\cline{1-7}
\multirow{2}{*}{alkp\_missing, n (\%)} & 0 & 0 & 2048 (94.9) & 506 (95.5) & 1237 (94.8) & 305 (94.7) \\
 & 1 &  & 109 (5.1) & 24 (4.5) & 68 (5.2) & 17 (5.3) \\
\cline{1-7}
\multirow{2}{*}{first\_p\_missing, n (\%)} & 0 & 0 & 2076 (96.2) & 517 (97.5) & 1253 (96.0) & 306 (95.0) \\
 & 1 &  & 81 (3.8) & 13 (2.5) & 52 (4.0) & 16 (5.0) \\
\cline{1-7}
\multirow{2}{*}{crp\_missing, n (\%)} & 0 & 0 & 327 (15.2) & 79 (14.9) & 203 (15.6) & 45 (14.0) \\
 & 1 &  & 1830 (84.8) & 451 (85.1) & 1102 (84.4) & 277 (86.0) \\
\cline{1-7}
\multirow{2}{*}{first\_cl\_missing, n (\%)} & 0 & 0 & 2049 (95.0) & 502 (94.7) & 1237 (94.8) & 310 (96.3) \\
 & 1 &  & 108 (5.0) & 28 (5.3) & 68 (5.2) & 12 (3.7) \\
\cline{1-7}
glu\_missing, n (\%) & 0 & 0 & 2157 (100.0) & 530 (100.0) & 1305 (100.0) & 322 (100.0) \\
\cline{1-7}
\multirow{2}{*}{dbp\_ts\_missing, n (\%)} & 0 & 0 & 2145 (99.4) & 525 (99.1) & 1298 (99.5) & 322 (100.0) \\
 & 1 &  & 12 (0.6) & 5 (0.9) & 7 (0.5) &  \\
\cline{1-7}
\multirow{2}{*}{sbp\_ts\_missing, n (\%)} & 0 & 0 & 2145 (99.4) & 525 (99.1) & 1298 (99.5) & 322 (100.0) \\
 & 1 &  & 12 (0.6) & 5 (0.9) & 7 (0.5) &  \\
\cline{1-7}
\multirow{2}{*}{pp\_ts\_missing, n (\%)} & 0 & 0 & 2145 (99.4) & 525 (99.1) & 1298 (99.5) & 322 (100.0) \\
 & 1 &  & 12 (0.6) & 5 (0.9) & 7 (0.5) &  \\
\cline{1-7}
\multirow{2}{*}{MAP\_ts\_missing, n (\%)} & 0 & 0 & 2145 (99.4) & 525 (99.1) & 1298 (99.5) & 322 (100.0) \\
 & 1 &  & 12 (0.6) & 5 (0.9) & 7 (0.5) &  \\
\cline{1-7}
\multirow{2}{*}{hr\_ts\_missing, n (\%)} & 0 & 0 & 2146 (99.5) & 526 (99.2) & 1298 (99.5) & 322 (100.0) \\
 & 1 &  & 11 (0.5) & 4 (0.8) & 7 (0.5) &  \\
\cline{1-7}
\multirow{2}{*}{bnp\_ts\_missing, n (\%)} & 0 & 0 & 1214 (56.3) & 307 (57.9) & 727 (55.7) & 180 (55.9) \\
 & 1 &  & 943 (43.7) & 223 (42.1) & 578 (44.3) & 142 (44.1) \\
\cline{1-7}
bun\_ts\_missing, n (\%) & 0 & 0 & 2157 (100.0) & 530 (100.0) & 1305 (100.0) & 322 (100.0) \\
\cline{1-7}
\multirow{2}{*}{hct\_ts\_missing, n (\%)} & 0 & 0 & 2153 (99.8) & 529 (99.8) & 1302 (99.8) & 322 (100.0) \\
 & 1 &  & 4 (0.2) & 1 (0.2) & 3 (0.2) &  \\
\cline{1-7}
\multirow{2}{*}{k\_ts\_missing, n (\%)} & 0 & 0 & 2139 (99.2) & 524 (98.9) & 1296 (99.3) & 319 (99.1) \\
 & 1 &  & 18 (0.8) & 6 (1.1) & 9 (0.7) & 3 (0.9) \\
\cline{1-7}
\multirow{2}{*}{na\_ts\_missing, n (\%)} & 0 & 0 & 2156 (100.0) & 530 (100.0) & 1304 (99.9) & 322 (100.0) \\
 & 1 &  & 1 (0.0) &  & 1 (0.1) &  \\
\cline{1-7}
\multirow{2}{*}{relative\_date\_first\_tn\_missing, n (\%)} & 0 & 0 & 1873 (86.8) & 457 (86.2) & 1137 (87.1) & 279 (86.6) \\
 & 1 &  & 284 (13.2) & 73 (13.8) & 168 (12.9) & 43 (13.4) \\
\cline{1-7}
\multirow{2}{*}{relative\_date\_first\_hb\_missing, n (\%)} & 0 & 0 & 2156 (100.0) & 530 (100.0) & 1304 (99.9) & 322 (100.0) \\
 & 1 &  & 1 (0.0) &  & 1 (0.1) &  \\
\cline{1-7}
\multirow{2}{*}{relative\_date\_first\_mcv\_missing, n (\%)} & 0 & 0 & 2156 (100.0) & 530 (100.0) & 1304 (99.9) & 322 (100.0) \\
 & 1 &  & 1 (0.0) &  & 1 (0.1) &  \\
\cline{1-7}
\multirow{2}{*}{relative\_date\_first\_rdw\_missing, n (\%)} & 0 & 0 & 2080 (96.4) & 516 (97.4) & 1254 (96.1) & 310 (96.3) \\
 & 1 &  & 77 (3.6) & 14 (2.6) & 51 (3.9) & 12 (3.7) \\
\cline{1-7}
\multirow{2}{*}{relative\_date\_first\_wbc\_missing, n (\%)} & 0 & 0 & 2156 (100.0) & 530 (100.0) & 1304 (99.9) & 322 (100.0) \\
 & 1 &  & 1 (0.0) &  & 1 (0.1) &  \\
\cline{1-7}
\multirow{2}{*}{relative\_date\_first\_alb\_missing, n (\%)} & 0 & 0 & 2084 (96.6) & 510 (96.2) & 1263 (96.8) & 311 (96.6) \\
 & 1 &  & 73 (3.4) & 20 (3.8) & 42 (3.2) & 11 (3.4) \\
\cline{1-7}
\multirow{2}{*}{relative\_date\_first\_ast\_missing, n (\%)} & 0 & 0 & 2074 (96.2) & 508 (95.8) & 1254 (96.1) & 312 (96.9) \\
 & 1 &  & 83 (3.8) & 22 (4.2) & 51 (3.9) & 10 (3.1) \\
\cline{1-7}
\multirow{2}{*}{relative\_date\_first\_alt\_missing, n (\%)} & 0 & 0 & 2048 (94.9) & 507 (95.7) & 1236 (94.7) & 305 (94.7) \\
 & 1 &  & 109 (5.1) & 23 (4.3) & 69 (5.3) & 17 (5.3) \\
\cline{1-7}
\multirow{2}{*}{relative\_date\_first\_ggt\_missing, n (\%)} & 0 & 0 & 2040 (94.6) & 503 (94.9) & 1232 (94.4) & 305 (94.7) \\
 & 1 &  & 117 (5.4) & 27 (5.1) & 73 (5.6) & 17 (5.3) \\
\cline{1-7}
\multirow{2}{*}{relative\_date\_first\_alkp\_missing, n (\%)} & 0 & 0 & 2048 (94.9) & 506 (95.5) & 1237 (94.8) & 305 (94.7) \\
 & 1 &  & 109 (5.1) & 24 (4.5) & 68 (5.2) & 17 (5.3) \\
\cline{1-7}
\multirow{2}{*}{relative\_date\_first\_p\_missing, n (\%)} & 0 & 0 & 2076 (96.2) & 517 (97.5) & 1253 (96.0) & 306 (95.0) \\
 & 1 &  & 81 (3.8) & 13 (2.5) & 52 (4.0) & 16 (5.0) \\
\cline{1-7}
\multirow{2}{*}{relative\_date\_first\_crp\_missing, n (\%)} & 0 & 0 & 327 (15.2) & 79 (14.9) & 203 (15.6) & 45 (14.0) \\
 & 1 &  & 1830 (84.8) & 451 (85.1) & 1102 (84.4) & 277 (86.0) \\
\cline{1-7}
\multirow{2}{*}{relative\_date\_first\_cl\_missing, n (\%)} & 0 & 0 & 2150 (99.7) & 528 (99.6) & 1300 (99.6) & 322 (100.0) \\
 & 1 &  & 7 (0.3) & 2 (0.4) & 5 (0.4) &  \\
\cline{1-7}
outcome, mean (SD) &  & 0 & 0.2 (1.5) & 0.2 (1.6) & 0.2 (1.5) & 0.2 (1.7) \\
\cline{1-7}
crea\_outcome, mean (SD) &  & 0 & 2.2 (1.3) & 2.2 (1.4) & 2.2 (1.2) & 2.2 (1.4) \\
\cline{1-7}
\multirow{2}{*}{c1, n (\%)} & 0.0 & 0 & 1065 (49.4) & 252 (47.5) & 651 (49.9) & 162 (50.3) \\
 & 1.0 &  & 1092 (50.6) & 278 (52.5) & 654 (50.1) & 160 (49.7) \\
\cline{1-7}
\bottomrule
    \caption{Covariants characteristics, used for policy construction. See \cref{tab:abbrev_dictionary} for descriptions of feature name abbreviations.}
    \label{tab:acute_table_one}
\end{longtable}
\end{scriptsize}

\scriptsize
\begin{longtable}{|l|l|}
\toprule
\textbf{Abbreviation} & \textbf{Description} \\ 
\midrule
\endfirsthead
\hline
\textbf{Abbreviation} & \textbf{Description} \\ 
\midrule
\endhead 
\hline
firstadm & Indicator of first admission (0/1) \\ \hline
admyear & Year of admission \\ \hline
age & Age in years (mean and standard deviation) \\ \hline
gender & Gender (0: Male, 1: Female) \\ \hline
Weight, Hight & Patient weight (kg) and height (cm) \\ \hline
bmi & Body Mass Index \\ \hline
htn & Hypertension indicator \\ \hline
firsttemp & First recorded temperature \\ \hline
dm & Diabetes mellitus indicator \\ \hline
smk & Smoking status (0: non-smoker, 1: former, 2: current) \\ \hline
ihd & Ischemic heart disease \\ \hline
vhd & Valvular heart disease \\ \hline
af & Atrial fibrillation \\ \hline
Hyperlipidemia & High lipid levels \\ \hline
copd & Chronic obstructive pulmonary disease \\ \hline
crf & Chronic renal failure \\ \hline
bb & Beta Blockers \\ \hline
acei & ACE Inhibitors \\ \hline
arf & Angiotensin Receptor Blockers \\ \hline
antiplt & Antiplatelet agents \\ \hline
anticoagulants & Anticoagulant use \\ \hline
furosemide & Use of furosemide \\ \hline
zaroxolin & Zaroxolin administration \\ \hline
thiamin & Thiamin (Vitamin B1) administration \\ \hline
insulin & Insulin use \\ \hline
tn, hb & Laboratory tests (e.g., total nitrogen, hemoglobin) \\ \hline
first\_mcv, rdw & Mean corpuscular volume and red cell distribution width \\ \hline
wbc & White blood cell count \\ \hline
alb & Albumin \\ \hline
ast, alt, ggt, alkp & Liver function tests (AST, ALT, GGT, ALKP) \\ \hline
first\_p & First phosphorus level \\ \hline
crp & C-reactive protein \\ \hline
first\_cl & First chloride level \\ \hline
glu & Glucose \\ \hline
dbp, sbp, pp & Blood pressure and pulse pressure (various measures) \\ \hline
MAP & Mean arterial pressure \\ \hline
hr & Heart rate \\ \hline
bnp & Brain natriuretic peptide \\ \hline
bun & Blood urea nitrogen \\ \hline
crea & Creatinine \\ \hline
hct & Hematocrit \\ \hline
k, na & Potassium, Sodium \\ \hline
EGFR & Estimated glomerular filtration rate \\ \hline
crea\_d\_point/diff & Measures related to creatinine changes \\ \hline
time\_to\_treat & Time (in days) to treatment \\ \hline
\multicolumn{2}{|l|}{\textbf{Drug Indicators:}} \\ \hline
1\_ind & Diuretics (administered during hospitalization) \\ \hline
2\_ind & Beta Blockers \\ \hline
3\_ind & ACE Inhibitors \\ \hline
4\_ind & Angiotensin Receptor Blockers \\ \hline
5\_ind & MRA (Mineralocorticoid Receptor Antagonists) \\ \hline
6\_ind & Fluids \\ \hline
7\_ind & Diuretics and ARB \\ \hline
8\_ind & Diuretics and ACE \\ \hline
N\_ind\_hosp & Indicates whether the patient had the drug N at the time of admission \\ \hline
returning\_patient & 
Indicator for returning patients, derived from the admission serial number. \\
& Assigns: 0 for (0,1] (first admission), 1 for (1,4] (1-4 previous admission), \\
& 2 for (4,8] (4-8 previous admissions), and 3 for (8,$\infty$) (8+ previous admissions).\\
 \hline
relative\_date\_first\_* & Relative time (in days) from admission to the first measurement of various tests \\ \hline
Weight\_missing, Hight\_missing, bmi\_missing, etc. & Indicators for missing data in the respective variables \\ \hline
crea\_outcome & Last creatinine value recorded within 7 days from the decision point \\ \hline
outcome (RTB) & Indicator of return to baseline creatinine level \\ \hline
c1 & Treatment adjustment for diuretics (increase or decrease) \\ \hline
\caption{Dictionary of Abbreviations for Data Description.}
\label{tab:abbrev_dictionary}
\end{longtable}

\subsection{Case study: propensity score model}\label{appendix_propensity}

\begin{figure}[H]
  \centering
      \includegraphics[width=0.6\textwidth]{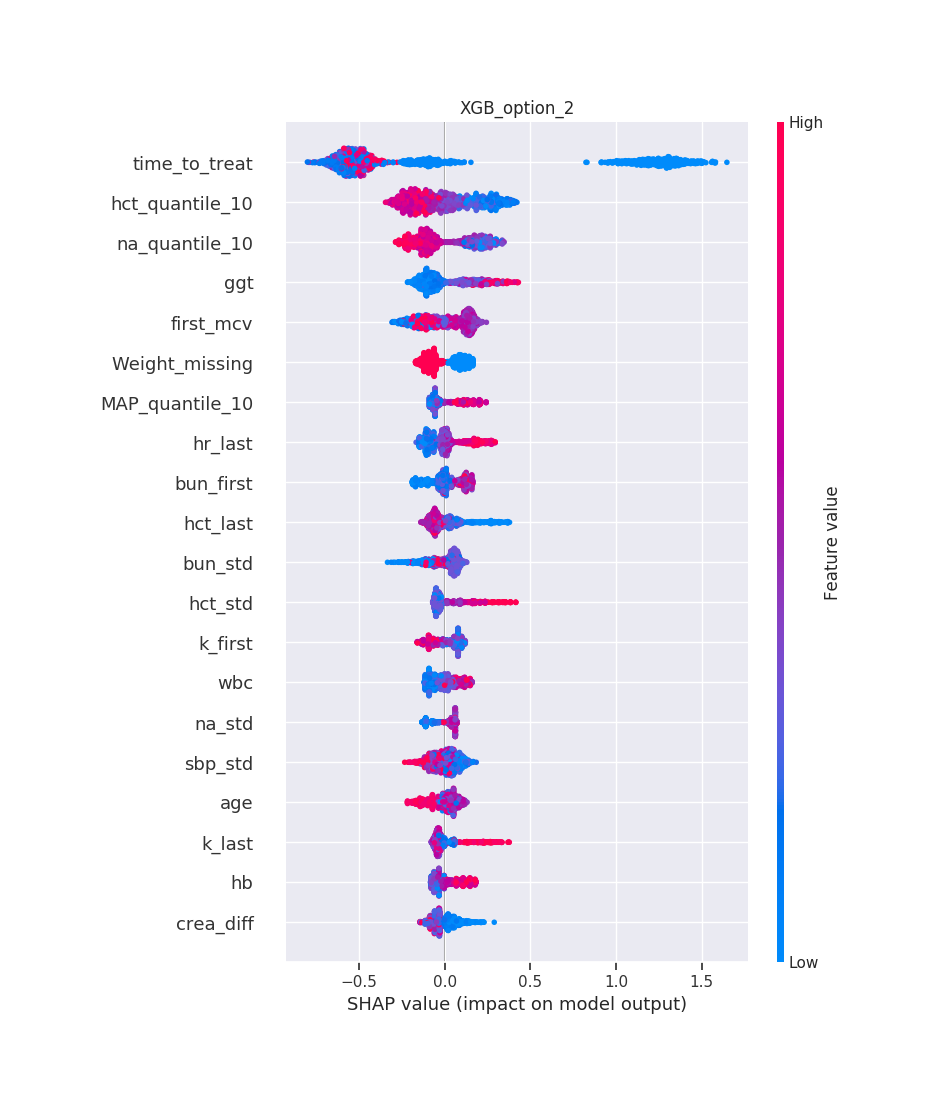}
  \caption{The features with the highest SHAP values in the XGBoost propensity score model.}
  \label{fig:acute_prop_features}
\end{figure}

\begin{figure}[H]
    \centering
    \begin{subfigure}{0.45\textwidth}
        \includegraphics[width=1.\textwidth]{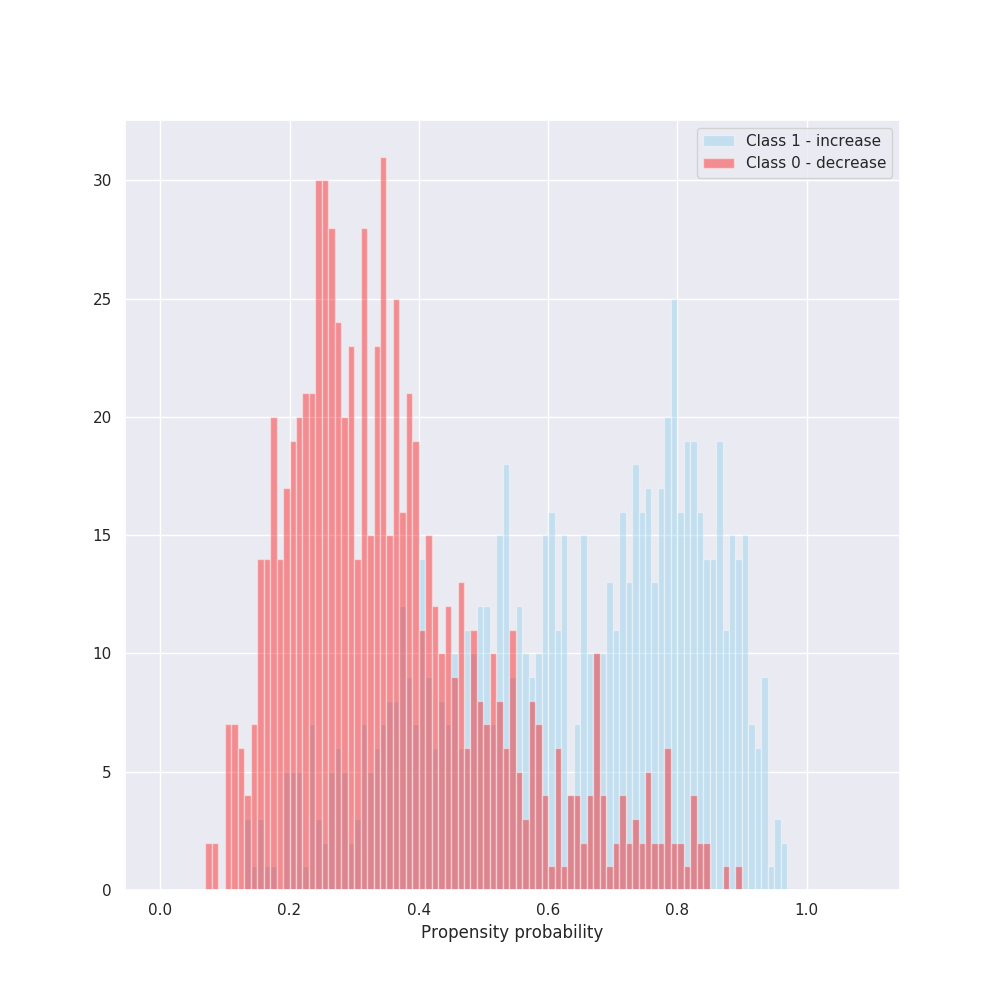}
        \caption{Propensity distribution before trimming}
        \label{fig:acture_before_prop_trim}
    \end{subfigure}
    \hfill
    \begin{subfigure}{0.45\textwidth}
        \includegraphics[width=1.\textwidth]{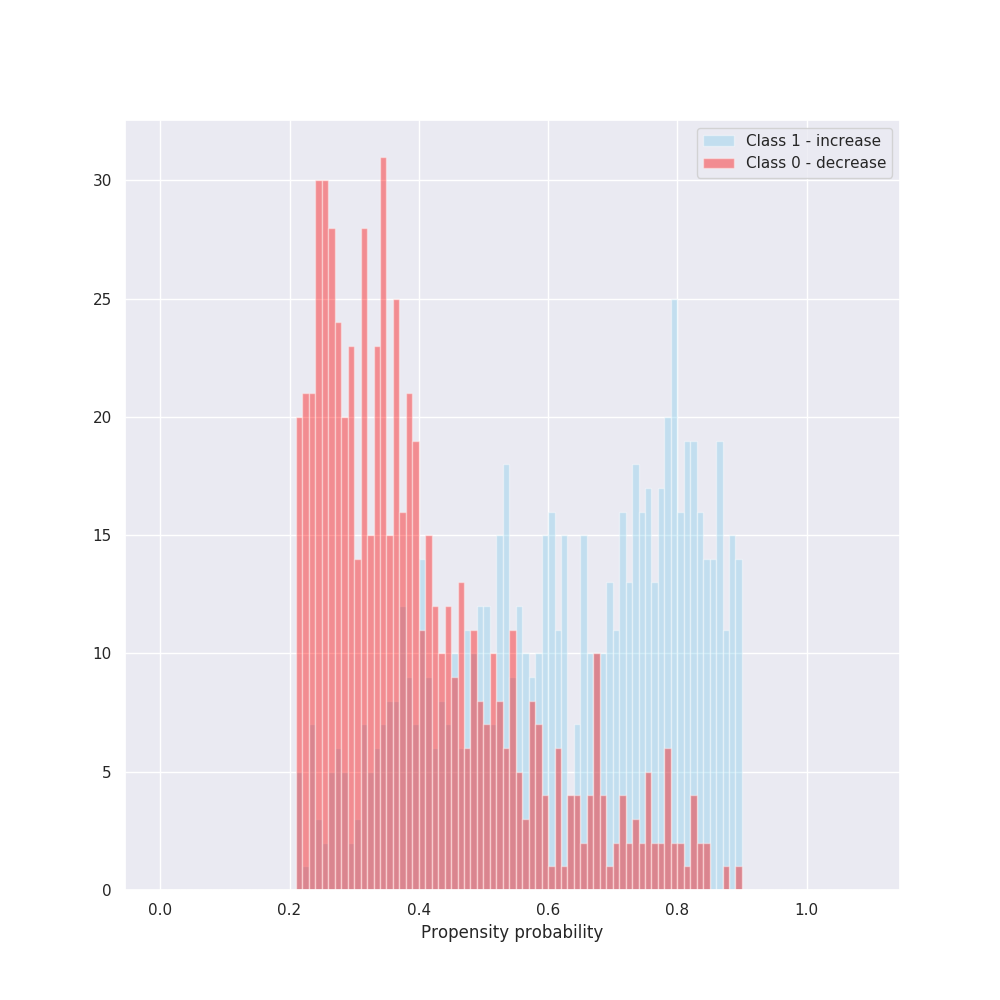}
        \caption{Propensity distribution after trimming}
        \label{fig:acture_after_prop_trim}
    \end{subfigure}
    \caption[trimming]{}
    \label{fig:acute_prop_triming}
\end{figure}

\begin{table}
    \centering
    \begin{tabular}{c|cccccc}
         & \textbf{Brier} & \textbf{AUROC} & \textbf{Accuracy} & \textbf{Precision} & \textbf{Recall} & \textbf{Fscore}\\ \toprule
         Train      & 0.153 & 0.878 & 0.798 & 0.8 & 0.799 & 0.798 \\
         Validation & 0.220 & 0.698 & 0.671 & 0.674 & 0.670 & 0.669\\
    \end{tabular}
    \caption{Goodness-of-fit metrics of the XGBoost model used for propensity score estimation. The results on the train and validation set. }
    \label{tab:acute_prop_metrics}
\end{table}

\subsection{Simulation results for case study}\label{sp:acute_policy_detalied}

\begin{figure}[H]
\centering
\includegraphics[width=0.5\textwidth]{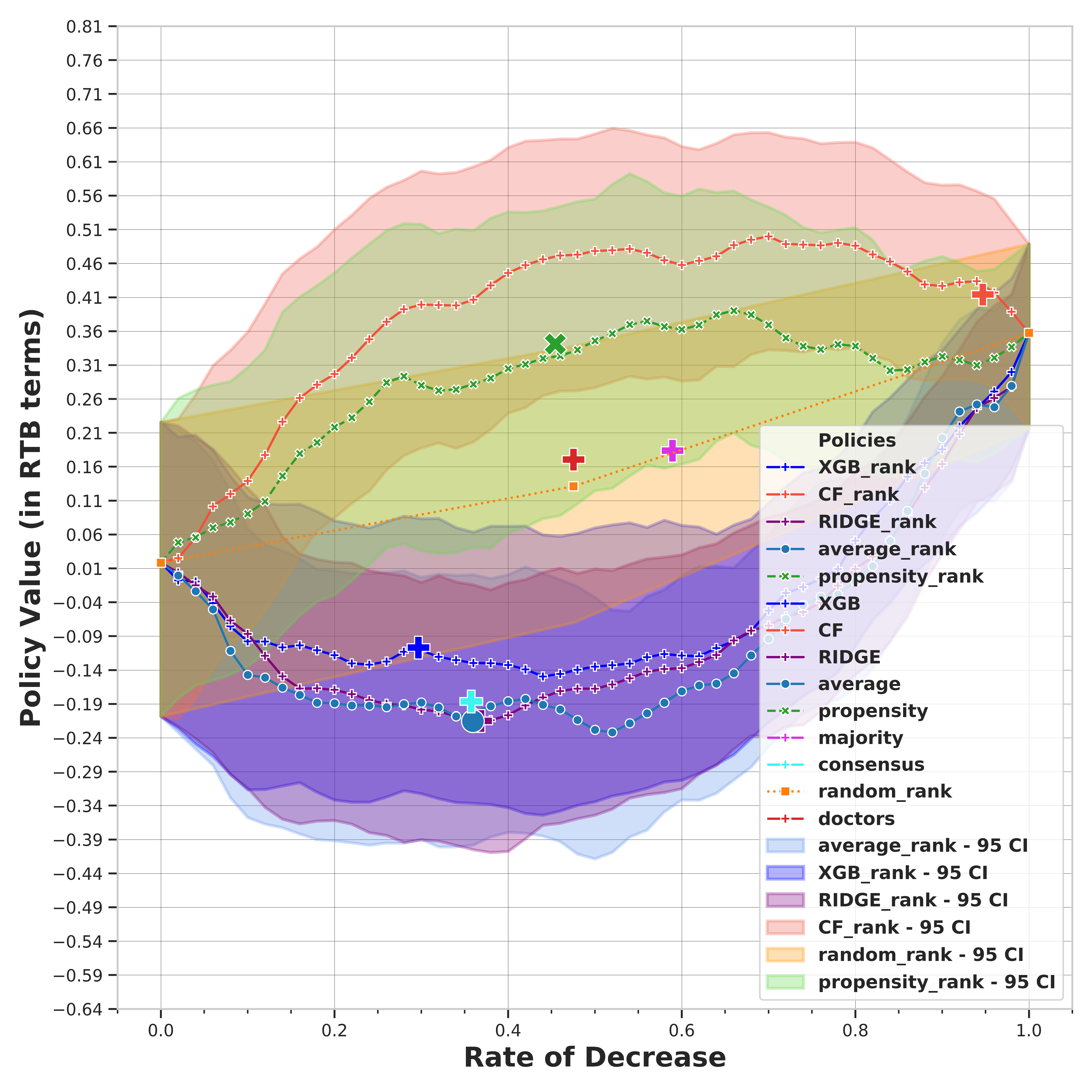}
\caption{Policy value rank graph for simulation data. Lower values are better.}
\label{fig:acute_p_val_th_sim}
\end{figure}

\begin{figure}[H]
\centering
\includegraphics[width=0.8\textwidth]{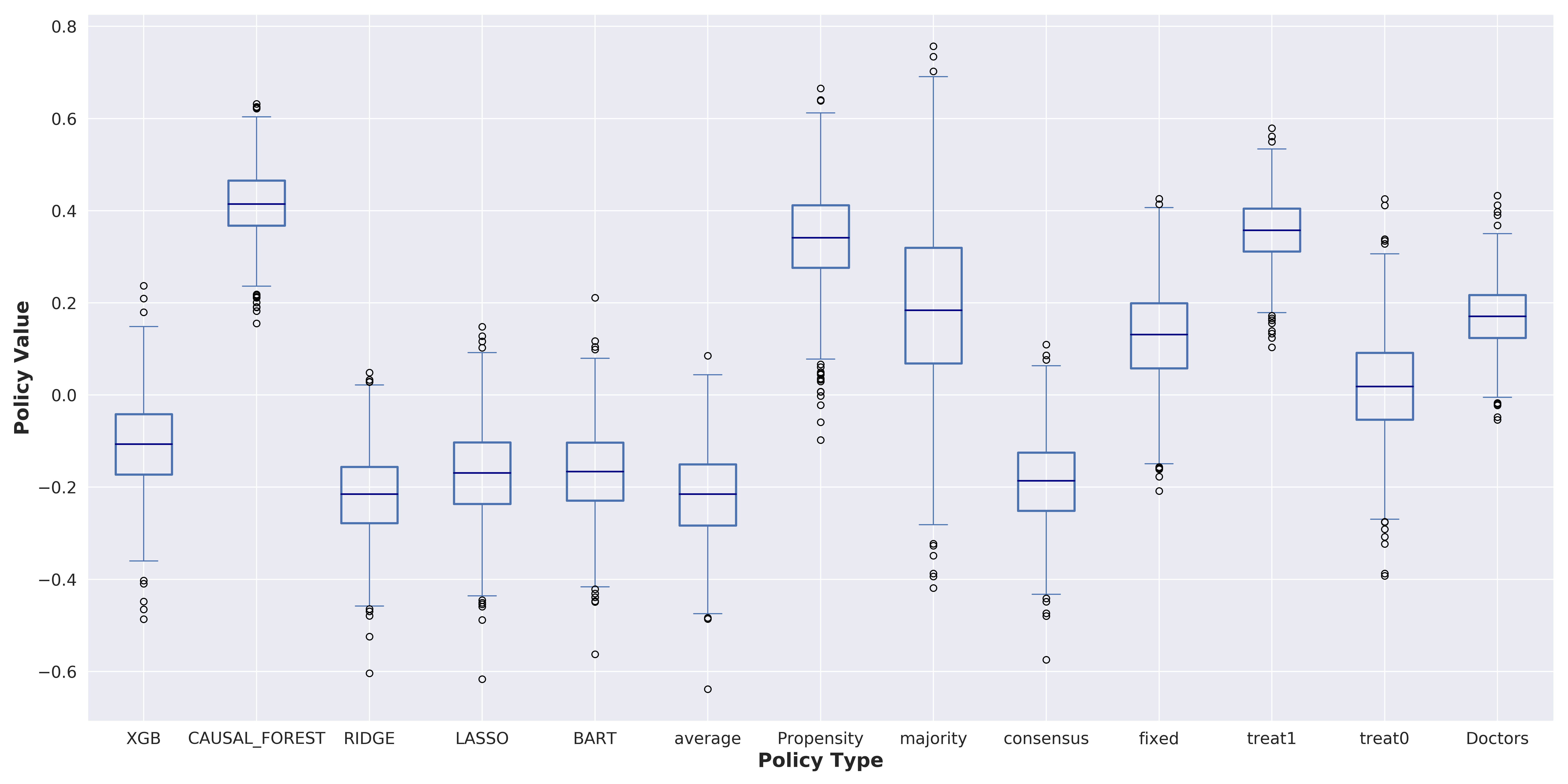}
\caption{Box-plots of simulation policy values, per policy. Lower values are better. }
\label{fig:acute_boxplot_sim}
\end{figure}

Following \cref{acute_simulation}, \cref{fig:acute_p_val_th_sim} and \cref{fig:acute_boxplot_sim} show a rank-graph and a box-plot graph of the policy values of various policies on the simulated data. For both policy value estimations, we use plain Logistic Regression (with L2 regularization) to estimate the ``true'' propensity score on all the data ($p^{*}(t=\pi(x)|x)$). In DR evaluation the XGB T-learner predication was used as the plug-in estimator $\hat{y}^{\pi(x)}$.

We see that we have a clear separation between the policy value of the ``Doctors'' treatment assignment, the policy values of two policies that performed poorly (Causal Forest and propensity-score based policy), and the other learned policies, which all performed well (note that lower is better in this case).

\subsection{Outcome models Results}\label{apx_sec:acute_outcome_models}

\begin{figure}[H]
    \centering
    \begin{subfigure}{0.45\textwidth}
        \includegraphics[width=1.\textwidth]{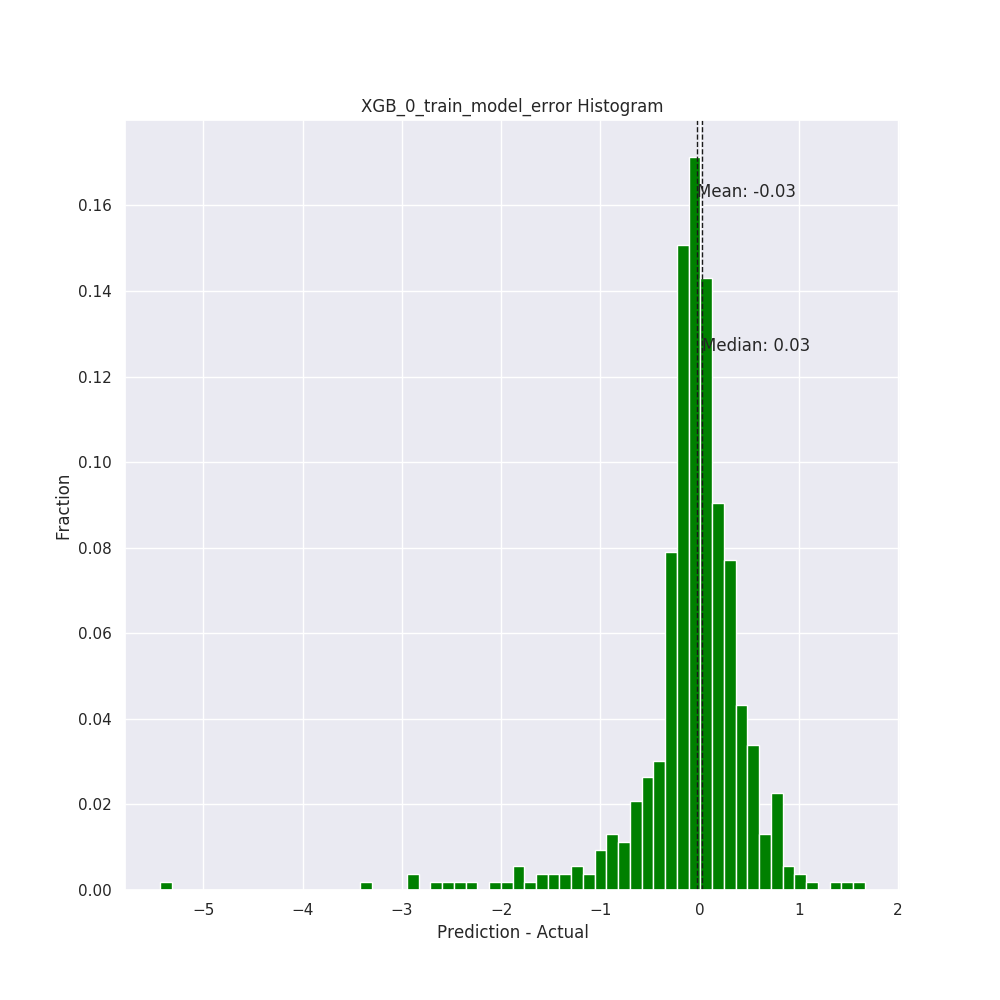}
        \caption{Error distributions of XGBoost model that predicts creatinine in the ``Decrease'' population.}
        \label{fig:xgb_0_error_hist}
    \end{subfigure}
    \hfill
    \begin{subfigure}{0.45\textwidth}
        \includegraphics[width=1.\textwidth]{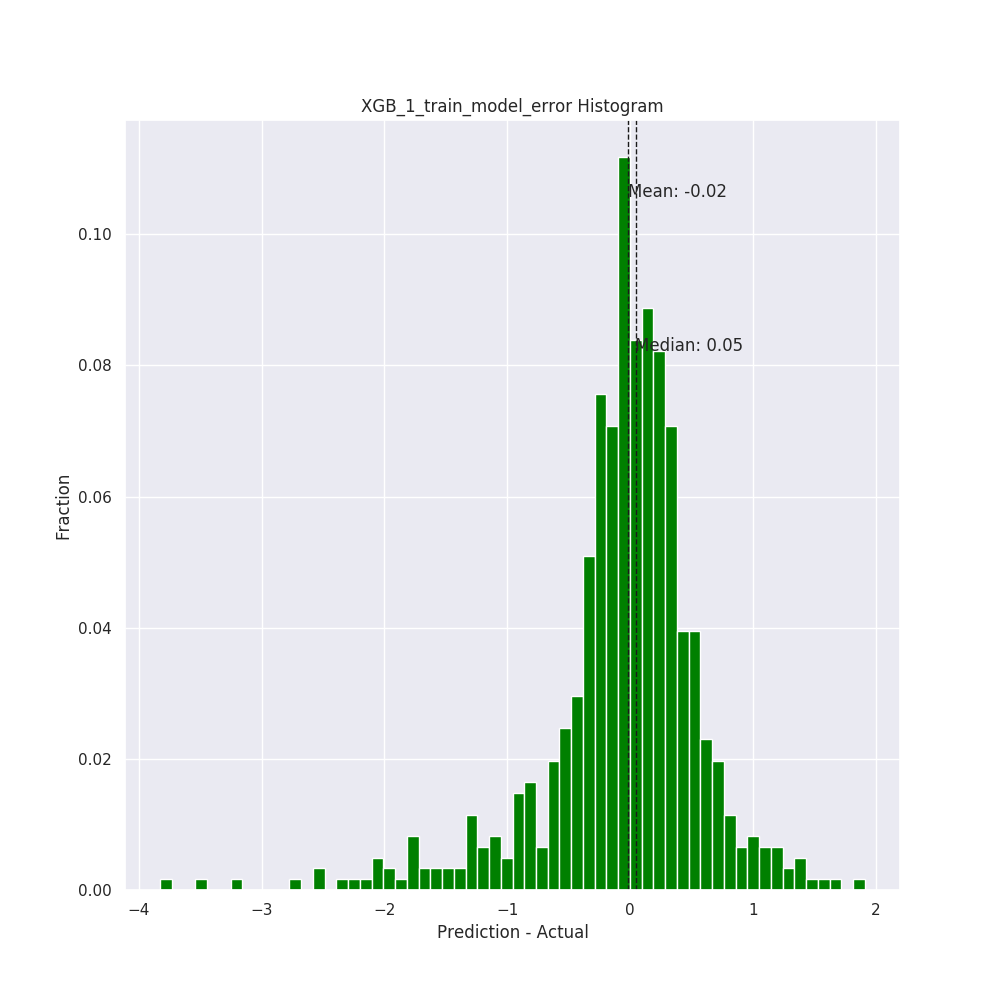}
        \caption{Error distributions of XGBoost model that predicts creatinine in the ``Increase'' population.}
        \label{fig:xgb_1_error_hist}
    \end{subfigure}
    \caption[XGBhist]{}
    \label{fig:xgb_error_hist}
\end{figure}

In \cref{fig:xgb_error_hist} we present the error distribution of the XGBoost model on the train set. The results indicate that the model mostly predicts well.

\begin{figure}[H]
    \centering
    \begin{subfigure}{0.45\textwidth}
        \includegraphics[width=1.\textwidth]{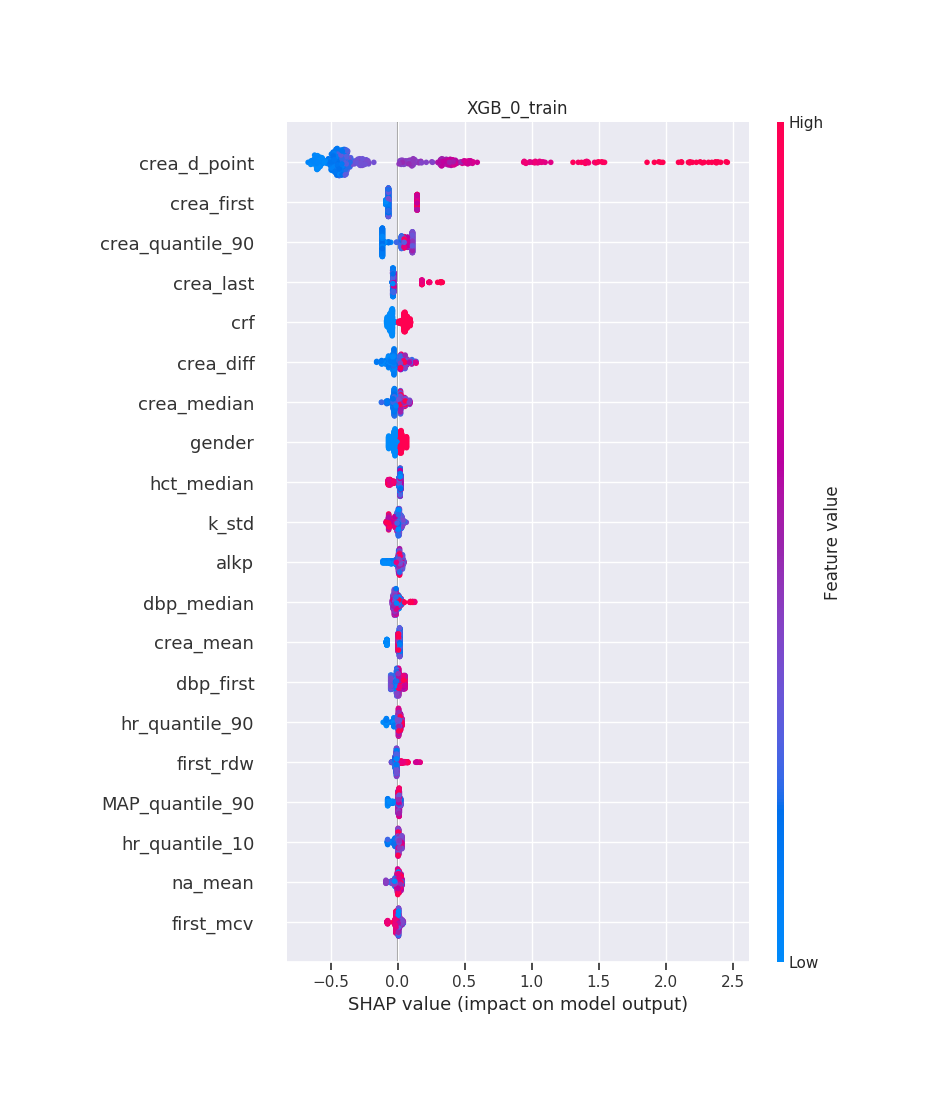}
        \caption{Important features, according to SHAP values of XGB model that predicts creatinine in the ``Decrease'' population.}
        \label{fig:xgb_0_features}
    \end{subfigure}
    \hfill
    \begin{subfigure}{0.45\textwidth}
        \includegraphics[width=1.\textwidth]{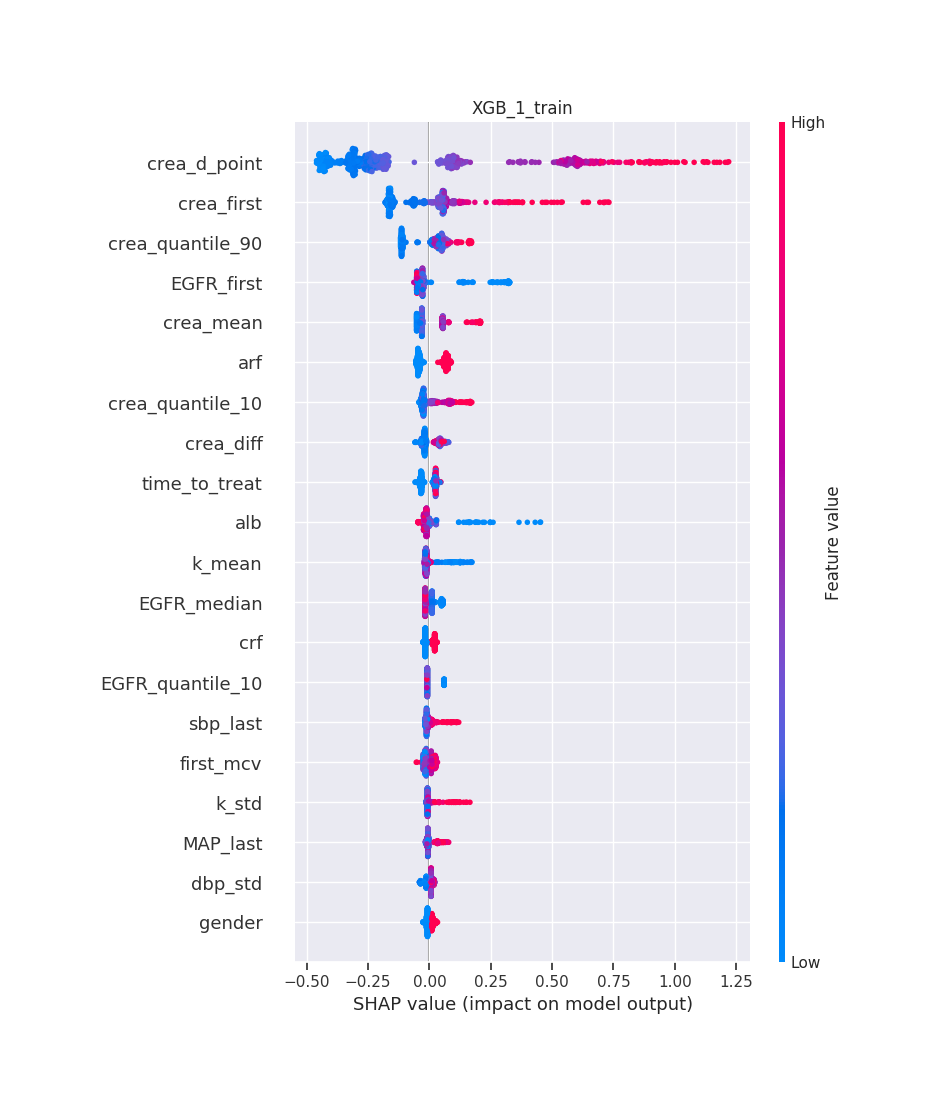}
        \caption{Important features, according to SHAP values of XGB model that predicts creatinine in the ``Increase'' population.}
        \label{fig:xgb_1_features}
    \end{subfigure}
    \caption[XgbFeatureImportants]{}
    \label{fig:xgb_features}
\end{figure}

In \cref{fig:xgb_features} we present the most important features, according to SHAP, for predicting creatinine in the underline XGBoost outcome models.

\begin{table}[H]
    \centering
    \scriptsize
    \begin{adjustwidth}{-2cm}{-1cm}
    \begin{tabular}{lcccccccccc}
        & \textbf{BART} & \textbf{XGBoost} & \textbf{Ridge} & \textbf{Lasso} & \textbf{Average} & \textbf{Causal Forest} & \textbf{Majority} & \textbf{Consensus} & \textbf{Propensity} & \textbf{DragoNet} \\
        \hline
        \textbf{BART} &  1.000 & 0.583 & 0.375 & 0.246 & 0.836 & 0.413 & 0.300 & 0.515 & 0.124 & 0.344 \\
        \textbf{XGBoost} & 0.583 & 1.000 & 0.221 & 0.370 & 0.753 & 0.510 & 0.284 & 0.542 & 0.042 & 0.305 \\
        \textbf{Ridge} & 0.375 & 0.221 & 1.000 & 0.104 & 0.686 & 0.165 & 0.179 & 0.353 & -0.004 & 0.259 \\
        \textbf{Lasso} & 0.246 & 0.370 & 0.104 & 1.000 & 0.470 & 0.623 & 0.216 & 0.513 & 0.060 & 0.250 \\
        \textbf{Average} & 0.836 & 0.753 & 0.686 & 0.470 & 1.000 & 0.557 & 0.346 & 0.657 & 0.079 & 0.412 \\
        \textbf{Causal Forest}& 0.413 & 0.510 & 0.165 & 0.623 & 0.557 & 1.000 & 0.279 & 0.520 & 0.117 & 0.388 \\
        \textbf{Majority} & 0.300 & 0.284 & 0.179 & 0.216 & 0.346 & 0.279 & 1.000 & 0.263 & 0.474 & 0.181 \\
        \textbf{Consensus} & 0.515 & 0.542 & 0.353 & 0.513 & 0.657 & 0.520 & 0.263 & 1.000 & 0.067 & 0.317  \\
        \textbf{Propensity} & 0.124 & 0.042 & -0.004 & 0.060 & 0.079 & 0.117 & 0.474 & 0.067 & 1.000 & 0.020 \\
        \textbf{DragoNet} & 0.344 & 0.305 & 0.259 & 0.250 & 0.412 & 0.388 & 0.181 & 0.317 & 0.020 & 1.000  \\
        \end{tabular}
        \end{adjustwidth}
        \caption{Pearson correlation between the CATE estimates on the train set}
        \label{tab:pearson_outcome_models}
\end{table}

\begin{table}[H]
    \centering
    \scriptsize
    \begin{adjustwidth}{-2cm}{-1cm}
    \begin{tabular}{lcccccccccc}
        & \textbf{BART} & \textbf{XGBoost} & \textbf{Ridge} & \textbf{Lasso} & \textbf{Average} & \textbf{Causal Forest} & \textbf{Majority} & \textbf{Consensus} & \textbf{Propensity} & \textbf{DragoNet} \\
        \hline
        \textbf{BART} &  1.000 & 0.368 & 0.274 & 0.170 & 0.632 & 0.239 & 0.251 & 0.456 & 0.081 & 0.284 \\
        \textbf{XGBoost} & 0.368 & 1.000 & 0.169 & 0.270 & 0.534 & 0.307 & 0.250 & 0.503 & 0.022 & 0.252 \\
        \textbf{Ridge} & 0.274 & 0.169 & 1.000 & 0.079 & 0.494 & 0.134 & 0.183 & 0.349 & -0.003 & 0.249 \\
        \textbf{Lasso} & 0.170 & 0.270 & 0.079 & 1.000 & 0.346 & 0.373 & 0.181 & 0.456 & 0.020 & 0.176 \\
        \textbf{Average} & 0.632 & 0.534 & 0.494 & 0.346 & 1.000 & 0.359 & 0.286 & 0.613 & 0.047 & 0.353 \\
        \textbf{Causal Forest}& 0.239 & 0.307 & 0.134 & 0.373 & 0.359 & 1.000 & 0.248 & 0.464 & 0.071 & 0.305 \\
        \textbf{Majority} & 0.251 & 0.250 & 0.183 & 0.181 & 0.286 & 0.248 & 1.000 & 0.263 & 0.386 & 0.181 \\
        \textbf{Consensus} & 0.456 & 0.503 & 0.349 & 0.456 & 0.613 & 0.464 & 0.263 & 1.000 & 0.052 & 0.317  \\
        \textbf{Propensity} & 0.081 & 0.022 & -0.003 & 0.020 & 0.047 & 0.071 & 0.386 & 0.052 & 1.000 & 0.014 \\
        \textbf{DragoNet} & 0.284 & 0.252 & 0.249 & 0.176 & 0.353 & 0.305 & 0.181 & 0.317 & 0.014 & 1.000  \\
        \end{tabular}
        \end{adjustwidth}
        \caption{Kendall correlation between the CATE estimates on the train set}
        \label{tab:kendall_outcome_models}

\end{table}

\begin{figure}[H]
\centering
\includegraphics[width=0.9\textwidth]{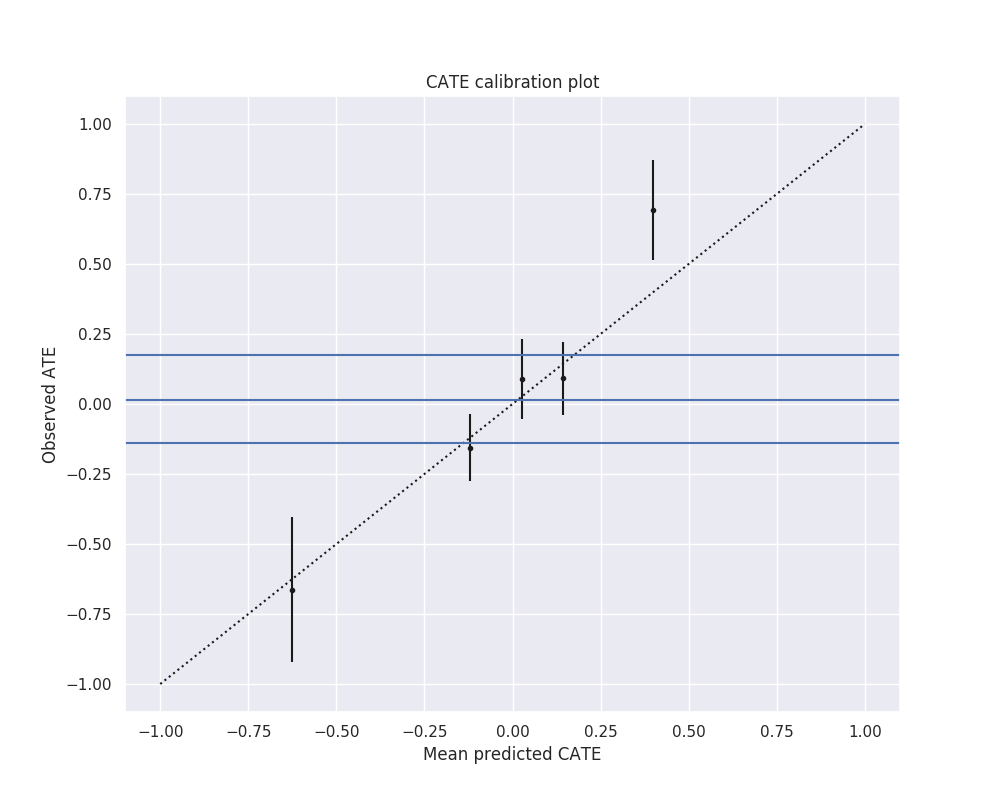}
\caption{CATE calibration graph of XGBoost T-learner model.}
\label{fig:acute_xgb_cate_clib}
\end{figure}

\subsection{Deferral Insights}\label{apx_sec:acute_deferral_inisghts}

\begin{figure}[htbp]
    \centering
    \includegraphics[width=.65\textwidth]{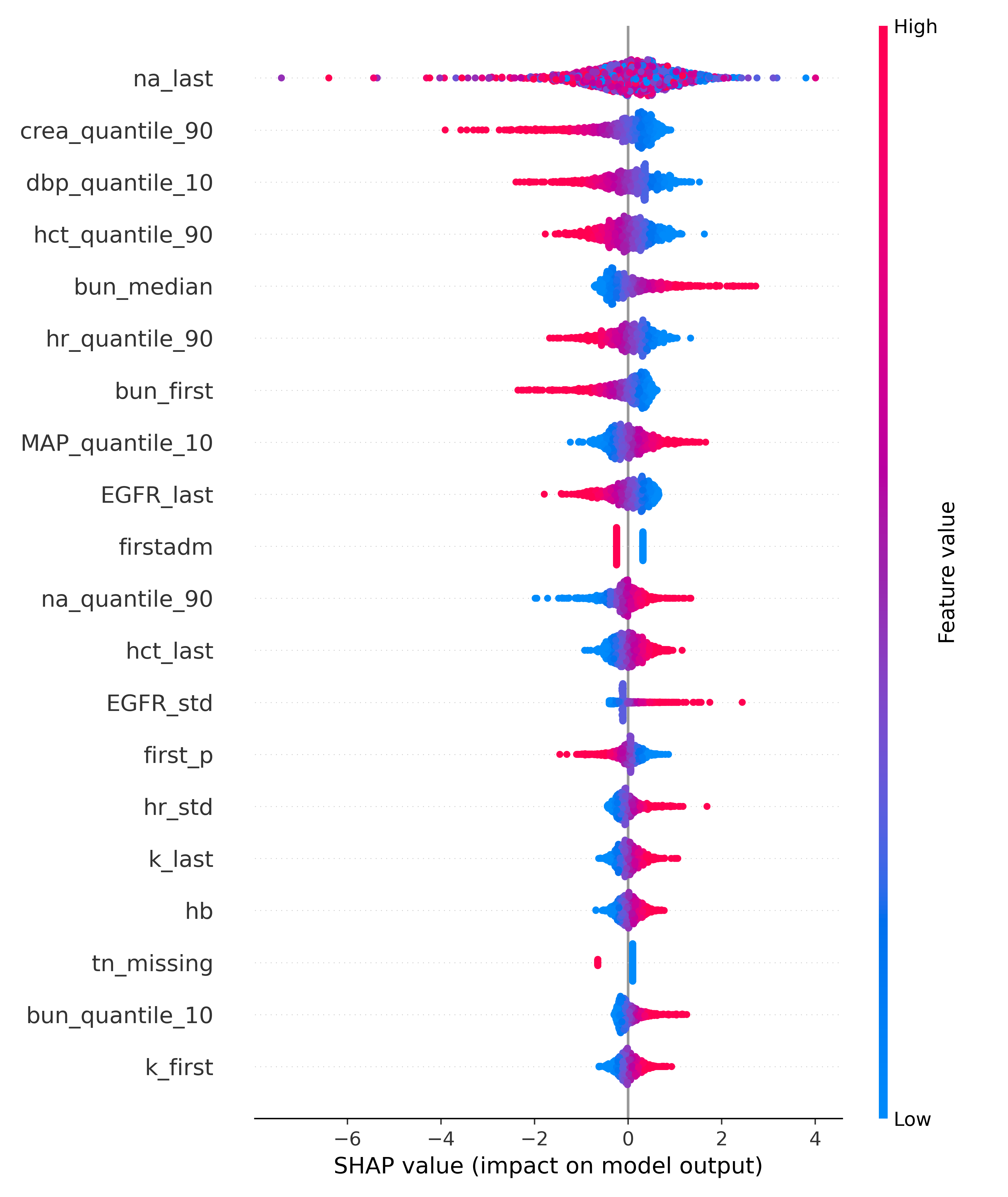}
    \caption{SHAP summary plot of the most important features of a logistic regression model trained on the deferral assignment. }
    \label{apx_fig:deferral_shap_values}
\end{figure}

To further understand the population that got deferral, in \cref{acute_deferal}, a policy summarization is required. 
Towards this goal, we train a $L_1$ regularized logistic regression model, on the label whether the patient was part of the deferral set. 
In \cref{apx_fig:deferral_shap_values} we present the SHAP summary plot \citep{lundberg2017unified}, with the 20 most important values. Then, in \cref{apx_tab:deferral_tab_one} we present the group statistics difference, based on the features in the summary plot. The analysis suggests that deferred patients have, among other things, lower values in the 90th quantile of creatinine ('crea\_quantile\_90'),  lower 10th quantile of diastolic blood pressure ('dbp\_quantile\_10'), and a lower proportion of first-time admissions (51.7 vs. 58.4). Conversely, the standard deviation of estimated glomerular filtration rate ('EGFR\_std') is higher in the deferred group, suggesting that greater variability in kidney function contribute to increased uncertainty in treatment recommendations.

\begin{table}[!htbp]
\begin{tabular}{llllll}
\toprule
 &  & \multicolumn{4}{r}{Grouped by deferred} \\
 &  & Missing & Overall & Rec & Def \\
\midrule
n &  &  & 1139 & 822 & 317 \\
\cline{1-6}
na\_last, mean (SD) &  & 0 & 137.8 (4.8) & 137.8 (4.9) & 137.8 (4.6) \\
\cline{1-6}
crea\_quantile\_90, mean (SD) &  & 0 & 1.8 (1.0) & 1.9 (1.1) & 1.7 (0.7) \\
\cline{1-6}
dbp\_quantile\_10, mean (SD) &  & 0 & 57.2 (10.9) & 57.4 (11.1) & 56.8 (10.6) \\
\cline{1-6}
hct\_quantile\_90, mean (SD) &  & 0 & 35.0 (5.3) & 35.0 (5.5) & 35.0 (5.0) \\
\cline{1-6}
bun\_median, mean (SD) &  & 0 & 40.0 (23.1) & 40.6 (23.7) & 38.6 (21.3) \\
\cline{1-6}
hr\_quantile\_90, mean (SD) &  & 0 & 91.6 (18.5) & 92.0 (18.6) & 90.3 (18.4) \\
\cline{1-6}
bun\_first, mean (SD) &  & 0 & 39.5 (23.5) & 40.2 (24.1) & 37.8 (21.8) \\
\cline{1-6}
MAP\_quantile\_10, mean (SD) &  & 0 & 77.0 (12.3) & 77.1 (12.5) & 76.6 (11.8) \\
\cline{1-6}
EGFR\_last, mean (SD) &  & 0 & 40.0 (21.1) & 39.5 (21.8) & 41.3 (19.0) \\
\cline{1-6}
\multirow{2}{*}{firstadm, n (\%)} & 0 & 0 & 495 (43.5) & 342 (41.6) & 153 (48.3) \\
 & 1 &  & 644 (56.5) & 480 (58.4) & 164 (51.7) \\
\cline{1-6}
na\_quantile\_90, mean (SD) &  & 0 & 139.2 (4.6) & 139.1 (4.7) & 139.3 (4.3) \\
\cline{1-6}
hct\_last, mean (SD) &  & 0 & 33.9 (5.6) & 33.9 (5.7) & 34.1 (5.3) \\
\cline{1-6}
EGFR\_std, mean (SD) &  & 0 & 3.8 (3.0) & 3.6 (3.0) & 4.1 (3.0) \\
\cline{1-6}
first\_p, mean (SD) &  & 0 & 4.4 (1.2) & 4.5 (1.2) & 4.2 (1.1) \\
\cline{1-6}
hr\_std, mean (SD) &  & 0 & 10.9 (6.1) & 10.8 (5.9) & 11.1 (6.7) \\
\cline{1-6}
k\_last, mean (SD) &  & 0 & 4.2 (0.7) & 4.2 (0.7) & 4.2 (0.7) \\
\cline{1-6}
hb, mean (SD) &  & 0 & 11.2 (1.9) & 11.2 (1.9) & 11.3 (1.8) \\
\cline{1-6}
\multirow{2}{*}{tn\_missing, n (\%)} & 0 & 0 & 981 (86.1) & 695 (84.5) & 286 (90.2) \\
 & 1 &  & 158 (13.9) & 127 (15.5) & 31 (9.8) \\
\cline{1-6}
bun\_quantile\_10, mean (SD) &  & 0 & 38.1 (22.8) & 38.6 (23.4) & 36.9 (21.3) \\
\cline{1-6}
k\_first, mean (SD) &  & 0 & 4.4 (0.6) & 4.4 (0.7) & 4.4 (0.6) \\
\cline{1-6}
\bottomrule
\end{tabular}
\caption{The group differences between the deferred (``Def'') group and the vs the patients that would get a recommendation (``Rec'') and the union of both (``Overall''), on the train set. The chosen features are based on the largest SHAP values of a logistic regression model.}
\label{apx_tab:deferral_tab_one}
\end{table}

\subsection{Policy value results}\label{apx_sec:acute_policy_val}
As noted in \cref{acute_target_trail}, the aim of this test case is constructing individual-level treatment policy. This recommendation is based on the CATE value, to determine which treatment would better affect the patient state,  which in this case means having a higher predicted RTB ratio.
We examined 12 such policies -- 5 model-based (XGBoost, Causal Forest, Ridge, Lasso, BART), 3 ensemble policies (Average, Majority, Consensus based on the 5 aforementioned models), 2 based on fixed rules (either Increase or Decrease dosage to all), a policy based on the propensity score, and current treatment policy (i.e. the policy actually observed in the data).
As detailed in \cref{pipeline}, we estimated the policy value for each rule in various methods on the held out data. 

\begin{figure}[ht]
  \centering
      \includegraphics[width=\textwidth]{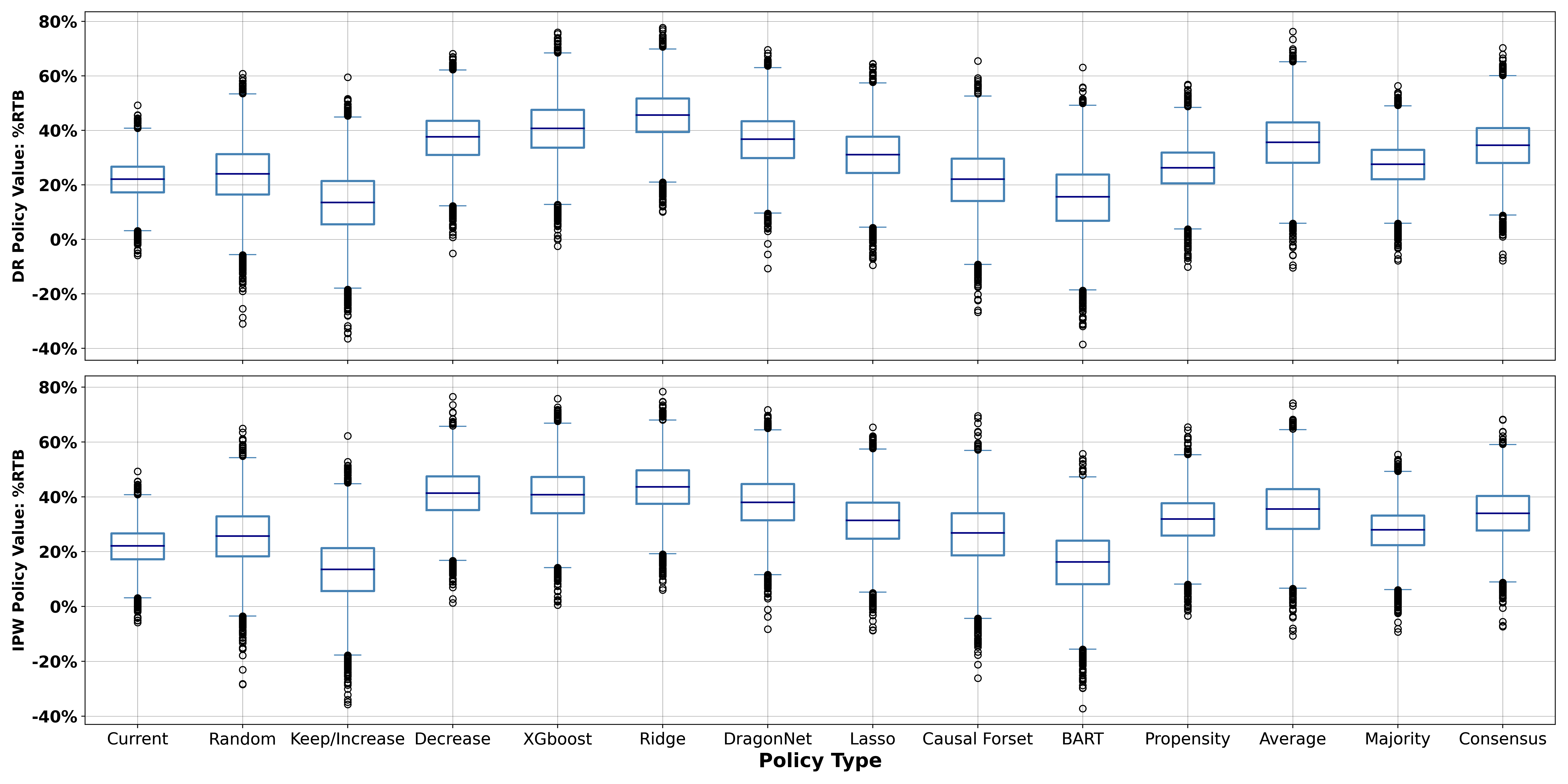}
  \caption{Policy value box-plot, result of running 10K bootstraps evaluation on held-out data, showing all policies: \emph{Keep/Increase}: all patients given ``increase'', \emph{Current}: current treatment by doctors, \emph{Random}: randomly assigning treatment at the same proportion as current treatment, \emph{Decrease}:  all patient given ``decrease'', \emph{XGBoost}: T-learner of XGBoost models, \emph{Ridge}: T-learner of Ridge models, \emph{DrangoNet}: A policy based on DragoNet estimator, \emph{Causal Forest}: A policy based on Causal forest estimator, \emph{BART}: T-learner of BART model, \emph{Lasso}: T-learner of Lasso models, \emph{Propensity}: a policy based on patient's propensity score. Ensemble models -- \emph{Average}: a policy based on the average of the above models, \emph{Majority}: a policy based on majority vote of the above models, and \emph{Consensus}: a policy based on consensus between the above models}
  \label{fig:ipw_dr_policy_values}
\end{figure}

\begin{figure}[ht]
  \centering
      \includegraphics[width=\textwidth]{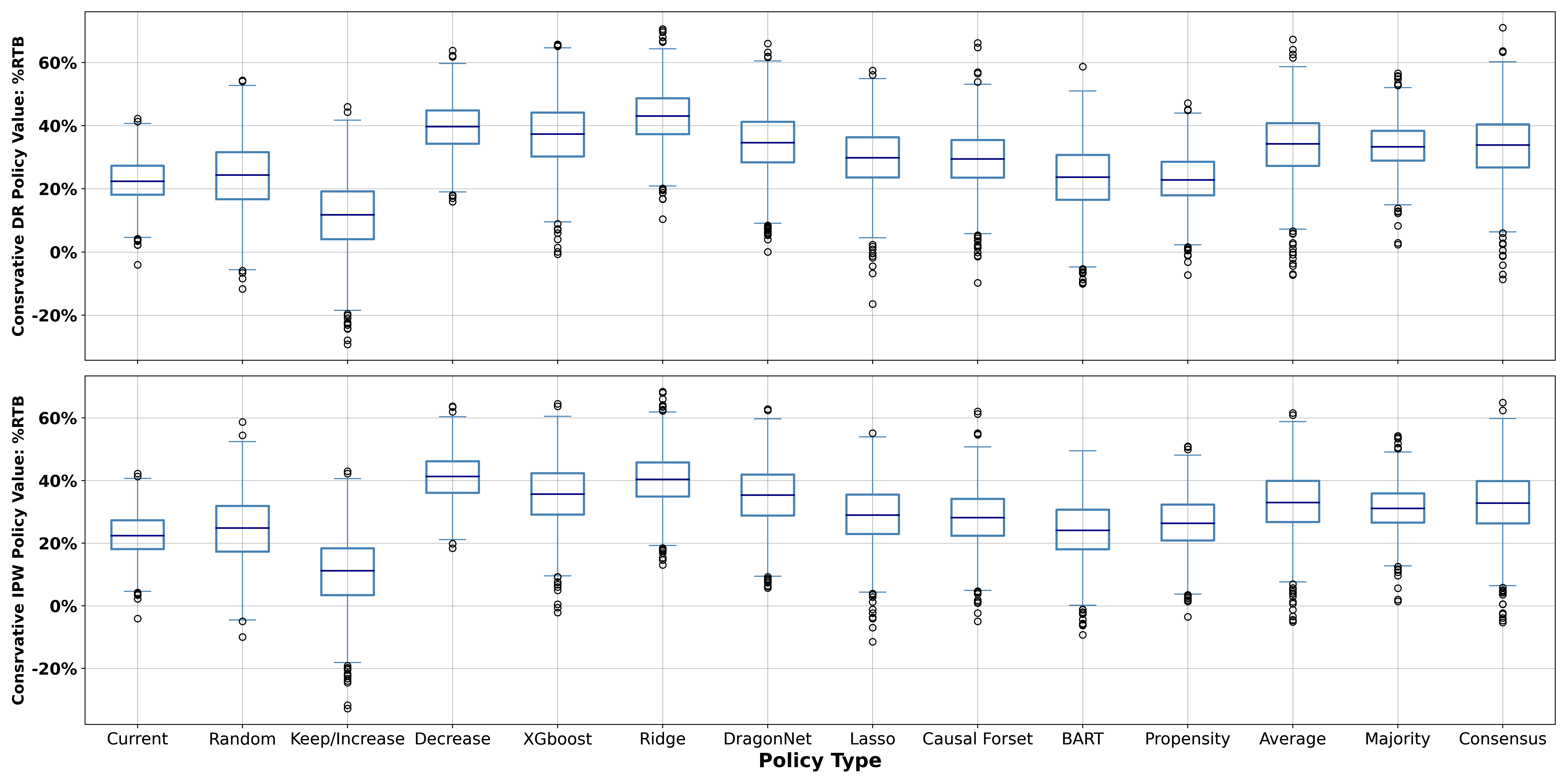}
  \caption{Policy value box-plot, result of running 10K bootstraps evaluation on held-out data, showing all policies, under conservative deferral set: \emph{Keep/Increase}: all patients given ``increase'', \emph{Current}: current treatment by doctors, \emph{Random}: randomly assigning treatment at the same proportion as current treatment, \emph{Decrease}:  all patient given ``decrease'', \emph{XGBoost}: T-learner of XGBoost models, \emph{Ridge}: T-learner of Ridge models, \emph{DrangoNet}: A policy based on DragoNet estimator, \emph{Causal Forest}: A policy based on Causal forest estimator, \emph{BART}: T-learner of BART model, \emph{Propensity}: a policy based on patient's propensity score, \emph{Average}: a policy based on the average of the above models. 
  }
  \label{fig:ipw_dr_policy_values_defer}
\end{figure}

For each policy we performed a 10K bootstrap where for each round we randomly select patients (with replacements) and evaluate the policy value with both the IPW and DR estimators (see \cref{subsec:polval} and \cref{policy_val_eq_appendix}). We use logistic regression with $L_2$ regularization to estimate the propensity score on all the data. In DR evaluation the XGBoost T-learner predictions were used as the plug-in estimator $\hat{y}^{\pi(x)}$. We present the results in \cref{fig:ipw_dr_policy_values} ("Inclusive" deferral set) and \cref{fig:ipw_dr_policy_values_defer} (“Conservative” deferral set).

\begin{table}[ht]
\begin{tabular}{l|c|rrrrrrr}
\toprule
{} &  $n$ &      Mean &       Std &       Min &       25\% &       50\% &       75\% &       Max \\
\midrule
\textbf{Current}       &  10000 &  22.00\% &  6.94\% & -5.87\% &  17.27\% &  22.07\% &  26.68\% &  49.24\% \\
\textbf{Random}        &  10000 &  23.72\% &  11.05\% & -31.00\% &  16.51\% &  24.02\% &  31.32\% &  60.83\% \\
\textbf{Keep/Increase} &  10000 &  13.35\% &  11.91\% & -36.48\% &  5.54\% &  13.56\% &  21.43\% &  59.53\% \\
\textbf{Decrease}      &  10000 &  37.22\% &  9.32\% & -5.19\% &  31.01\% &  37.61\% &  43.53\% &  68.16\% \\
\textbf{XGBoost}       &  10000 &  40.53\% &  10.37\% & -2.50\% &  33.67\% &  40.70\% &  47.60\% &  76.01\% \\
\textbf{Ridge}         &  10000 &  45.45\% &  9.19\% &  10.11\% &  39.42\% &  45.61\% &  51.72\% &  77.73\% \\
\textbf{DragonNet}     &  10000 &  36.59\% &  9.86\% & -10.71\% &  29.89\% &  36.74\% &  43.37\% &  69.59\% \\
\textbf{Lasso}         &  10000 &  30.88\% &  10.02\% & -9.47\% &  24.39\% &  31.08\% &  37.70\% &  64.49\% \\
\textbf{Causal Forset} &  10000 &  21.63\% &  11.53\% & -26.73\% &  14.10\% &  22.10\% &  29.63\% &  65.46\% \\
\textbf{BART}          &  10000 &  15.21\% &  12.34\% & -38.57\% &  6.87\% &  15.61\% &  23.83\% &  63.08\% \\
\textbf{Propensity}    &  10000 &  26.20\% &  8.43\% & -10.10\% &  20.62\% &  26.27\% &  31.86\% &  56.83\% \\
\textbf{Average}       &  10000 &  35.40\% &  10.90\% & -10.44\% &  28.15\% &  35.64\% &  43.00\% &  76.30\% \\
\textbf{Majority}      &  10000 &  27.38\% &  8.14\% & -7.80\% &  22.08\% &  27.54\% &  32.87\% &  56.31\% \\
\textbf{Consensus}     &  10000 &  34.38\% &  9.58\% & -7.81\% &  28.08\% &  34.56\% &  40.91\% &  70.33\% \\
\bottomrule
\end{tabular}
\caption{Doubly robust (DR) estimates of policy values over 10K bootstraps rounds on held-out data, showing all policies: \emph{Keep/Increase}: all patients given ``increase'', \emph{Current}: current treatment by doctors, \emph{Random}: randomly assigning treatment at the same proportion as current treatment, \emph{Decrease}:  all patient given ``decrease'', \emph{XGBoost}: T-learner of XGBoost models, \emph{Ridge}: T-learner of Ridge models, \emph{DrangoNet}: A policy based on DragoNet estimator, \emph{Causal Forest}: A policy based on Causal forest estimator, \emph{BART}: T-learner of BART model, \emph{Lasso}: T-learner of Lasso models, \emph{Propensity}: a policy based on patient's propensity score. Ensemble models -- \emph{Average}: a policy based on the average of the above models, \emph{Majority}: a policy based on majority vote of the above models, and \emph{Consensus}: a policy based on consensus between the above models}
\label{tab:full_dr_policy_values}
\end{table}

\begin{table}[ht]
\begin{tabular}{l|c|rrrrrrr}
\toprule
{} &  $n$ &      Mean &       Std &       Min &       25\% &       50\% &       75\% &       Max \\
\midrule
\textbf{Current}       &  10000 &  22.00\% &  6.94\% & -5.87\% &  17.27\% &  22.07\% &  26.68\% &  49.24\% \\
\textbf{Random}        &  10000 &  25.41\% &  10.95\% & -28.42\% &  18.29\% &  25.70\% &  32.88\% &  64.96\% \\
\textbf{Keep/Increase} &  10000 &  13.32\% &  11.87\% & -35.76\% &  5.66\% &  13.51\% &  21.30\% &  62.24\% \\
\textbf{Decrease}      &  10000 &  41.15\% &  9.09\% &  1.29\% &  35.18\% &  41.33\% &  47.46\% &  76.50\% \\
\textbf{XGBoost}       &  10000 &  40.60\% &  9.96\% &  0.55\% &  34.05\% &  40.77\% &  47.29\% &  75.83\% \\
\textbf{Ridge}         &  10000 &  43.47\% &  9.19\% &  5.99\% &  37.48\% &  43.60\% &  49.71\% &  78.35\% \\
\textbf{DragonNet}     &  10000 &  37.97\% &  9.88\% & -8.29\% &  31.44\% &  38.01\% &  44.67\% &  71.74\% \\
\textbf{Lasso}         &  10000 &  31.26\% &  9.91\% & -8.81\% &  24.75\% &  31.42\% &  37.89\% &  65.33\% \\
\textbf{Causal Forset} &  10000 &  26.12\% &  11.63\% & -26.12\% &  18.70\% &  26.79\% &  34.04\% &  69.50\% \\
\textbf{BART}          &  10000 &  15.91\% &  11.60\% & -37.21\% &  8.17\% &  16.26\% &  24.05\% &  55.68\% \\
\textbf{Propensity}    &  10000 &  31.81\% &  8.77\% & -3.41\% &  25.87\% &  31.88\% &  37.71\% &  65.44\% \\
\textbf{Average}       &  10000 &  35.45\% &  10.75\% & -10.72\% &  28.35\% &  35.55\% &  42.85\% &  74.17\% \\
\textbf{Majority}      &  10000 &  27.72\% &  8.06\% & -9.30\% &  22.36\% &  27.94\% &  33.16\% &  55.45\% \\
\textbf{Consensus}     &   10000 &  33.92\% &  9.32\% & -7.32\% &  27.74\% &  34.00\% &  40.33\% &  68.25\% \\
\bottomrule
\end{tabular}
\caption{Inverse Propensity Weighting (IPW) estimates of policy values over 10K bootstraps rounds on held-out data, showing all policies: \emph{Keep/Increase}: all patients given ``increase'', \emph{Current}: current treatment by doctors, \emph{Random}: randomly assigning treatment at the same proportion as current treatment, \emph{Decrease}:  all patient given ``decrease'', \emph{XGBoost}: T-learner of XGBoost models, \emph{Ridge}: T-learner of Ridge models, \emph{DrangoNet}: A policy based on DragoNet estimator, \emph{Causal Forest}: A policy based on Causal forest estimator, \emph{BART}: T-learner of BART model, \emph{Lasso}: T-learner of Lasso models, \emph{Propensity}: a policy based on patient's propensity score. Ensemble models -- \emph{Average}: a policy based on the average of the above models, \emph{Majority}: a policy based on majority vote of the above models, and \emph{Consensus}: a policy based on consensus between the above models}
\label{tab:full_ipw_policy_values}
\end{table}

\begin{figure}[ht]
\centering
\includegraphics[width=0.8\textwidth]{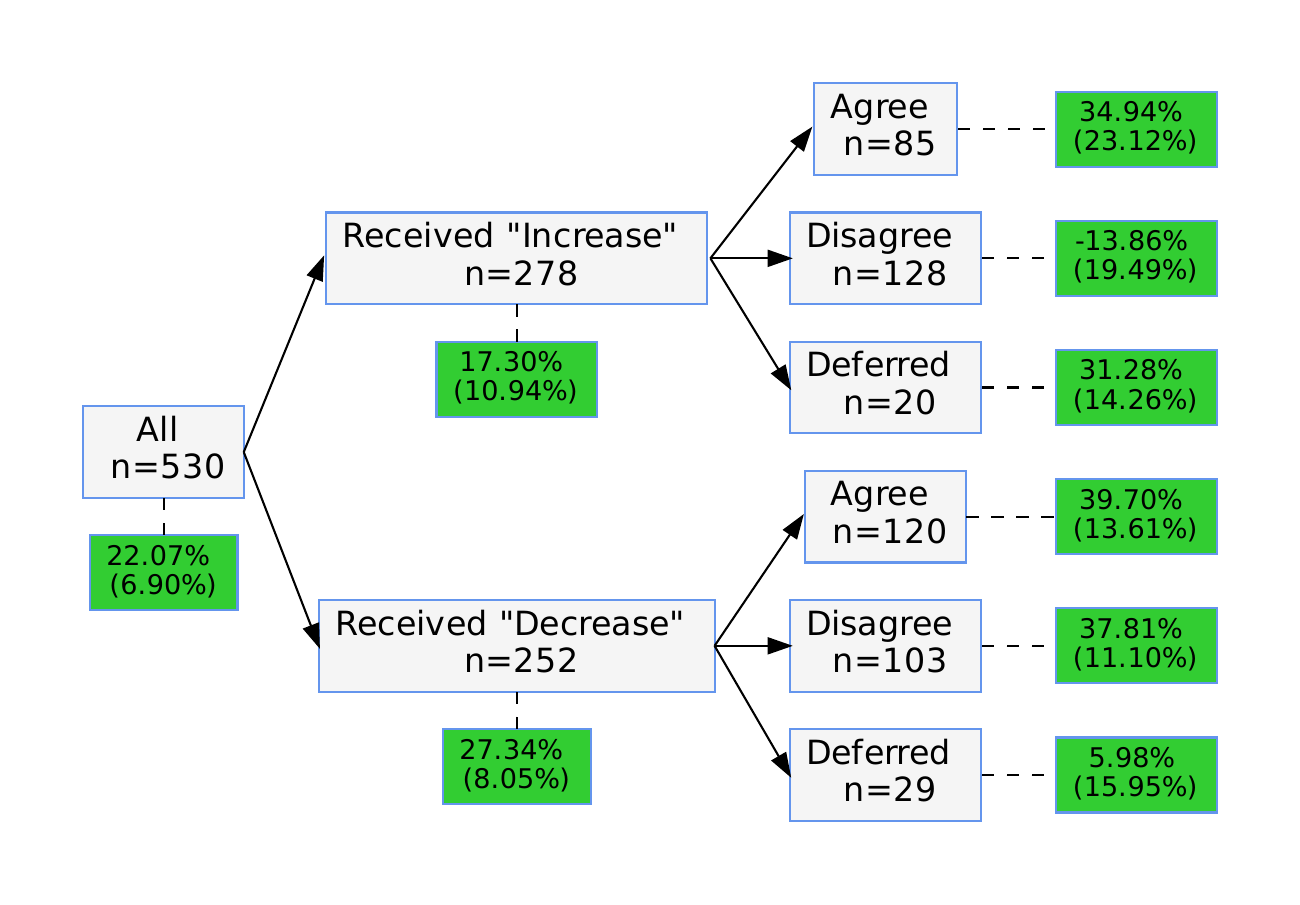}
\caption{``Outcome Tree'' under conservative deferral set: A graph representing the mean RTB value for each policy group, using XGBoost T-learner as the policy. Results on the ``Conservative'' set.
The gray boxes represent the number of patients in each subgroup, and the green boxes represent their mean RTB value and SEM in parentheses.}
\label{fig:acute_outcome_tree_conservative}
\end{figure}

%% file: new_main.bbl
\begin{thebibliography}{149}
\providecommand{\natexlab}[1]{#1}
\providecommand{\url}[1]{\texttt{#1}}
\expandafter\ifx\csname urlstyle\endcsname\relax
  \providecommand{\doi}[1]{doi: #1}\else
  \providecommand{\doi}{doi: \begingroup \urlstyle{rm}\Url}\fi

\bibitem[Hern{\'{a}}n and Robins(2016)]{Hernan2016UsingAvailable}
Miguel~A. Hern{\'{a}}n and James~M. Robins.
\newblock {Using Big Data to Emulate a Target Trial When a Randomized Trial Is Not Available}.
\newblock \emph{American Journal of Epidemiology}, 183\penalty0 (8):\penalty0 758--764, 2016.
\newblock ISSN 14766256.
\newblock \doi{10.1093/aje/kwv254}.

\bibitem[Bica et~al.(2021)Bica, Alaa, Lambert, and Schaar]{bica2021fromchallenges}
Ioana Bica, Ahmed~M. Alaa, Craig Lambert, and Mihaela van~der Schaar.
\newblock {From Real‐World Patient Data to Individualized Treatment Effects Using Machine Learning: Current and Future Methods to Address Underlying Challenges}.
\newblock \emph{Clinical Pharmacology {\&} Therapeutics}, 109\penalty0 (1):\penalty0 87--100, 1 2021.
\newblock ISSN 1532-6535.
\newblock \doi{10.1002/CPT.1907}.

\bibitem[Powell et~al.(2021)Powell, Koenecke, Byrd, Nishimura, Konig, Xiong, Mahmood, Mucaj, Bettegowda, Rose, Tamang, Sacarny, Caffo, Athey, Stuart, and Vogelstein]{Powell2021TenStudy}
Michael Powell, Allison Koenecke, James~Brian Byrd, Akihiko Nishimura, Maximilian~F. Konig, Ruoxuan Xiong, Sadiqa Mahmood, Vera Mucaj, Chetan Bettegowda, Liam Rose, Suzanne Tamang, Adam Sacarny, Brian Caffo, Susan Athey, Elizabeth~A. Stuart, and Joshua~T. Vogelstein.
\newblock {Ten Rules for Conducting Retrospective Pharmacoepidemiological Analyses: Example COVID-19 Study}.
\newblock \emph{Frontiers in Pharmacology}, 12:\penalty0 1799, 7 2021.
\newblock ISSN 16639812.
\newblock \doi{10.3389/fphar.2021.700776}.

\bibitem[Meid et~al.(2020)Meid, Ruff, Wirbka, Stoll, Seidling, Groll, and Haefeli]{meid2020usingdata}
Andreas~D. Meid, Carmen Ruff, Lucas Wirbka, Felicitas Stoll, Hanna~M. Seidling, Andreas Groll, and Walter~E. Haefeli.
\newblock {Using the causal inference framework to support individualized drug treatment decisions based on observational healthcare data}.
\newblock \emph{Clinical Epidemiology}, 12:\penalty0 1223--1234, 2020.
\newblock ISSN 11791349.
\newblock \doi{10.2147/CLEP.S274466}.

\bibitem[Kent et~al.(2018)Kent, Steyerberg, and Van~Klaveren]{kent2018personalizedeffects}
David~M. Kent, Ewout Steyerberg, and David Van~Klaveren.
\newblock {Personalized evidence based medicine: Predictive approaches to heterogeneous treatment effects}.
\newblock \emph{BMJ (Online)}, 363, 2018.
\newblock ISSN 17561833.
\newblock \doi{10.1136/bmj.k4245}.

\bibitem[Eini-Porat et~al.(2022)Eini-Porat, Amir, Eytan, and Shalit]{eini2022tell}
Bar Eini-Porat, Ofra Amir, Danny Eytan, and Uri Shalit.
\newblock Tell me something interesting: Clinical utility of machine learning prediction models in the icu.
\newblock \emph{Journal of Biomedical Informatics}, page 104107, 2022.

\bibitem[Prosperi et~al.(2020)Prosperi, Guo, Sperrin, Koopman, Min, He, Rich, Wang, Buchan, and Bian]{Prosperi2020CausalHealthcare}
Mattia Prosperi, Yi~Guo, Matt Sperrin, James~S. Koopman, Jae~S. Min, Xing He, Shannan Rich, Mo~Wang, Iain~E. Buchan, and Jiang Bian.
\newblock {Causal inference and counterfactual prediction in machine learning for actionable healthcare}.
\newblock \emph{Nature Machine Intelligence}, 2\penalty0 (7):\penalty0 369--375, 7 2020.
\newblock ISSN 25225839.
\newblock \doi{10.1038/s42256-020-0197-y}.

\bibitem[Boulos et~al.(2019)Boulos, Darawsha, Abassi, Azzam, and Aronson]{Boulos2019TreatmentInjury}
Jubran Boulos, Wisam Darawsha, Zaid~A. Abassi, Zaher~S. Azzam, and Doron Aronson.
\newblock {Treatment patterns of patients with acute heart failure who develop acute kidney injury}.
\newblock \emph{ESC Heart Failure}, 6\penalty0 (1):\penalty0 45--52, 2 2019.
\newblock ISSN 20555822.
\newblock \doi{10.1002/ehf2.12364}.

\bibitem[Damman et~al.(2014)Damman, Valente, Voors, O'Connor, Van~Veldhuisen, and Hillege]{damman2014renalmeta-analysis}
Kevin Damman, Mattia~A.E. Valente, Adriaan~A. Voors, Christopher~M. O'Connor, Dirk~J. Van~Veldhuisen, and Hans~L. Hillege.
\newblock {Renal impairment, worsening renal function, and outcome in patients with heart failure: An updated meta-analysis}.
\newblock \emph{European Heart Journal}, 35\penalty0 (7):\penalty0 455--469, 2 2014.
\newblock ISSN 0195668X.
\newblock \doi{10.1093/eurheartj/eht386}.

\bibitem[Tang and Mullens(2010)]{tang2010cardiorenal}
WH~Wilson Tang and Wilfried Mullens.
\newblock Cardiorenal syndrome in decompensated heart failure.
\newblock \emph{Heart}, 96\penalty0 (4):\penalty0 255--260, 2010.

\bibitem[Rubin(1974)]{Rubin1974EstimatingStudies}
Donald~B. Rubin.
\newblock {Estimating causal effects of treatments in randomized and nonrandomized studies}.
\newblock \emph{Journal of Educational Psychology}, 66\penalty0 (5):\penalty0 688--701, 1974.
\newblock ISSN 00220663.
\newblock \doi{10.1037/h0037350}.

\bibitem[Holland and Rubin(1987)]{holland1987causalstudies}
Paul~W. Holland and Donald~B. Rubin.
\newblock {Causal inference in retrospective studies}.
\newblock \emph{ETS Research Report Series}, 1987\penalty0 (1):\penalty0 203--231, 6 1987.
\newblock ISSN 2330-8516.
\newblock \doi{10.1002/j.2330-8516.1987.tb00211.x}.

\bibitem[Pearl(2009)]{pearl2009causality}
Judea Pearl.
\newblock \emph{{Causality}}.
\newblock Cambridge university press, illustrate edition, 2009.
\newblock ISBN 9780521895606.

\bibitem[Hern{\'{a}}n(2011)]{hernan2011withepidemiology}
Miguel~A Hern{\'{a}}n.
\newblock {With great data comes great responsibility: Publishing comparative effectiveness research in epidemiology}.
\newblock \emph{Epidemiology}, 22\penalty0 (3):\penalty0 290--291, 2011.
\newblock ISSN 10443983.
\newblock \doi{10.1097/EDE.0b013e3182114039}.

\bibitem[Shalit(2019)]{shalit2019candata}
Uri Shalit.
\newblock Can we learn individual-level treatment policies from clinical data?
\newblock \emph{Biostatistics}, 21\penalty0 (2):\penalty0 359--362, 11 2019.
\newblock ISSN 1465-4644.
\newblock \doi{10.1093/biostatistics/kxz043}.

\bibitem[Shi and Norgeot(2022)]{shi2022learning}
Jingpu Shi and Beau Norgeot.
\newblock Learning causal effects from observational data in healthcare: A review and summary.
\newblock \emph{Frontiers in Medicine}, page 2027, 2022.

\bibitem[Fern{\'{a}}ndez-Lor{\'{i}}a and Provost(2021)]{fernandezloria2021causalmatters}
Carlos Fern{\'{a}}ndez-Lor{\'{i}}a and Foster Provost.
\newblock {Causal Decision Making and Causal Effect Estimation Are Not the Same... and Why It Matters}.
\newblock \emph{arXiv preprint arXiv:2104.04103}, 2021.

\bibitem[Hernan and Robins(2020)]{hernan2020causal}
MA~Hernan and J~Robins.
\newblock \emph{Causal Inference: What if.}
\newblock Chapman \& Hill/CRC, Boca Raton, 2020.

\bibitem[Chernozhukov et~al.(2024)Chernozhukov, Hansen, Kallus, Spindler, and Syrgkanis]{chernozhukov2024applied}
Victor Chernozhukov, Christian Hansen, Nathan Kallus, Martin Spindler, and Vasilis Syrgkanis.
\newblock Applied causal inference powered by ml and ai.
\newblock \emph{arXiv preprint arXiv:2403.02467}, 2024.

\bibitem[Wager(2024)]{wager2024causal}
Stefan Wager.
\newblock Causal inference: A statistical learning approach, 2024.

\bibitem[Imbens(2004)]{Imbens2004NonparametricReview}
Guido~W Imbens.
\newblock {Nonparametric estimation of average treatment effects under exogeneity: A review}.
\newblock \emph{Review of Economics and Statistics}, 86\penalty0 (1):\penalty0 4--29, 2004.
\newblock ISSN 00346535.
\newblock \doi{10.1162/003465304323023651}.

\bibitem[Rosenbaum and Rubin(1983)]{ROSENBAUM1983TheEffects}
Paul~R. Rosenbaum and Donald~B. Rubin.
\newblock {The central role of the propensity score in observational studies for causal effects}.
\newblock \emph{Biometrika}, 70\penalty0 (1):\penalty0 41--55, 4 1983.
\newblock ISSN 00063444.
\newblock \doi{10.1093/biomet/70.1.41}.

\bibitem[Rubin(1980)]{Rubin1980RandomizationComment}
Donald~B Rubin.
\newblock {Randomization Analysis of Experimental Data: The Fisher Randomization Test Comment}.
\newblock \emph{Journal of the American Statistical Association}, 75\penalty0 (371):\penalty0 591--593, 1980.

\bibitem[Cole and Hern{\'{a}}n(2008)]{Cole2008ConstructingModels}
Stephen~R. Cole and Miguel~A. Hern{\'{a}}n.
\newblock {Constructing inverse probability weights for marginal structural models}.
\newblock \emph{American Journal of Epidemiology}, 168\penalty0 (6):\penalty0 656--664, 7 2008.
\newblock ISSN 00029262.
\newblock \doi{10.1093/aje/kwn164}.

\bibitem[Tan(2006)]{tan2006}
Zhiqiang Tan.
\newblock A distributional approach for causal inference using propensity scores.
\newblock \emph{Journal of the American Statistical Association}, 101\penalty0 (476):\penalty0 1619--1637, dec 2006.
\newblock \doi{10.1198/016214506000000023}.

\bibitem[Yadlowsky et~al.(2022)Yadlowsky, Namkoong, Basu, Duchi, and Tian]{yadlowsky2022bounds}
Steve Yadlowsky, Hongseok Namkoong, Sanjay Basu, John Duchi, and Lu~Tian.
\newblock Bounds on the conditional and average treatment effect with unobserved confounding factors.
\newblock \emph{The Annals of Statistics}, 50\penalty0 (5):\penalty0 2587--2615, 2022.

\bibitem[Kallus et~al.(2019)Kallus, Mao, and Zhou]{kallus2019interval}
Nathan Kallus, Xiaojie Mao, and Angela Zhou.
\newblock Interval estimation of individual-level causal effects under unobserved confounding.
\newblock In \emph{The 22nd international conference on artificial intelligence and statistics}, pages 2281--2290. PMLR, 2019.

\bibitem[Jesson et~al.(2021)Jesson, Mindermann, Gal, and Shalit]{jesson2021quantifying}
Andrew Jesson, S{\"o}ren Mindermann, Yarin Gal, and Uri Shalit.
\newblock Quantifying ignorance in individual-level causal-effect estimates under hidden confounding.
\newblock In \emph{International Conference on Machine Learning}, pages 4829--4838. PMLR, 2021.

\bibitem[Jin et~al.(2021)Jin, Ren, and Cand{\`e}s]{jin2021sensitivity}
Ying Jin, Zhimei Ren, and Emmanuel~J Cand{\`e}s.
\newblock Sensitivity analysis of individual treatment effects: A robust conformal inference approach.
\newblock \emph{arXiv preprint arXiv:2111.12161}, 2021.

\bibitem[Yin et~al.(2022)Yin, Shi, Wang, and Blei]{yin2022conformal}
Mingzhang Yin, Claudia Shi, Yixin Wang, and David~M Blei.
\newblock Conformal sensitivity analysis for individual treatment effects.
\newblock \emph{Journal of the American Statistical Association}, pages 1--14, 2022.

\bibitem[Oprescu et~al.(2023)Oprescu, Dorn, Ghoummaid, Jesson, Kallus, and Shalit]{oprescu2023b}
Miruna Oprescu, Jacob Dorn, Marah Ghoummaid, Andrew Jesson, Nathan Kallus, and Uri Shalit.
\newblock B-learner: Quasi-oracle bounds on heterogeneous causal effects under hidden confounding.
\newblock \emph{arXiv preprint arXiv:2304.10577}, 2023.

\bibitem[Austin(2011)]{austin2011anstudies}
Peter~C. Austin.
\newblock {An introduction to propensity score methods for reducing the effects of confounding in observational studies}.
\newblock \emph{Multivariate Behavioral Research}, 46\penalty0 (3):\penalty0 399--424, 5 2011.
\newblock ISSN 00273171.
\newblock \doi{10.1080/00273171.2011.568786}.

\bibitem[K{\"{u}}nzel et~al.(2019)K{\"{u}}nzel, Sekhon, Bickel, and Yu]{Kunzel2019MetalearnersLearning}
S\"{o}ren~R. K{\"{u}}nzel, Jasjeet~S. Sekhon, Peter~J. Bickel, and Bin Yu.
\newblock {Metalearners for estimating heterogeneous treatment effects using machine learning}.
\newblock \emph{Proceedings of the National Academy of Sciences of the United States of America}, 116\penalty0 (10):\penalty0 4156--4165, 2019.
\newblock ISSN 10916490.
\newblock \doi{10.1073/pnas.1804597116}.

\bibitem[Nie and Wager(2021)]{nie2021quasi}
Xinkun Nie and Stefan Wager.
\newblock Quasi-oracle estimation of heterogeneous treatment effects.
\newblock \emph{Biometrika}, 108\penalty0 (2):\penalty0 299--319, 5 2021.
\newblock ISSN 0006-3444.

\bibitem[Kennedy(2020)]{kennedy2020towards}
Edward~H Kennedy.
\newblock Towards optimal doubly robust estimation of heterogeneous causal effects.
\newblock \emph{arXiv preprint arXiv:2004.14497}, 2020.

\bibitem[Okasa(2022)]{okasa2022meta}
Gabriel Okasa.
\newblock Meta-learners for estimation of causal effects: Finite sample cross-fit performance.
\newblock \emph{arXiv preprint arXiv:2201.12692}, 2022.

\bibitem[Wager and Athey(2018)]{Wager2018EstimationForests}
Stefan Wager and Susan Athey.
\newblock {Estimation and Inference of Heterogeneous Treatment Effects using Random Forests}.
\newblock \emph{Journal of the American Statistical Association}, 113\penalty0 (523):\penalty0 1228--1242, 7 2018.
\newblock ISSN 1537274X.
\newblock \doi{10.1080/01621459.2017.1319839}.

\bibitem[Breiman(2001)]{Breiman2001RandomForests}
Leo Breiman.
\newblock {Random Forests}.
\newblock \emph{Machine Learning}, 45\penalty0 (1):\penalty0 5--32, 2001.
\newblock ISSN 08856125.
\newblock \doi{10.1023/A:1010933404324}.

\bibitem[Powers et~al.(2018)Powers, Qian, Jung, Schuler, Shah, Hastie, and Tibshirani]{powers2018somedimensions}
Scott Powers, Junyang Qian, Kenneth Jung, Alejandro Schuler, Nigam~H. Shah, Trevor Hastie, and Robert Tibshirani.
\newblock {Some methods for heterogeneous treatment effect estimation in high dimensions}.
\newblock \emph{Statistics in Medicine}, 37\penalty0 (11):\penalty0 1767--1787, 5 2018.
\newblock ISSN 10970258.
\newblock \doi{10.1002/sim.7623}.

\bibitem[Gruber and van~der Laan(2010)]{Gruber2010AOutcome.}
Susan Gruber and Mark~J van~der Laan.
\newblock {A targeted maximum likelihood estimator of a causal effect on a bounded continuous outcome.}
\newblock \emph{The international journal of biostatistics}, 6\penalty0 (1):\penalty0 Article 26, 2010.
\newblock ISSN 1557-4679.
\newblock \doi{10.2202/1557-4679.1260}.

\bibitem[van~der Laan and Luedtke(2015)]{vanderLaan2015TargetedRule}
Mark~J. van~der Laan and Alexander~R. Luedtke.
\newblock {Targeted Learning of the Mean Outcome under an Optimal Dynamic Treatment Rule}.
\newblock \emph{Journal of Causal Inference}, 3\penalty0 (1), 10 2015.
\newblock ISSN 2193-3677.
\newblock \doi{10.1515/jci-2013-0022}.

\bibitem[Cheng et~al.(2019)Cheng, Dumitrascu, Zhang, Chivers, Draugelis, Li, and Engelhardt]{Cheng2019Patient-SpecificProcesses}
Li-Fang Cheng, Bianca Dumitrascu, Michael Zhang, Corey Chivers, Michael Draugelis, Kai Li, and Barbara~E. Engelhardt.
\newblock {Patient-Specific Effects of Medication Using Latent Force Models with Gaussian Processes}.
\newblock \emph{arXiv preprint arXiv:1906.00226}, 2019.

\bibitem[Shalit et~al.(2017)Shalit, Johansson, and Sontag]{Shalit2017EstimatingAlgorithms}
Uri Shalit, Fredrik~D Johansson, and David Sontag.
\newblock {Estimating individual treatment effect: Generalization bounds and algorithms}.
\newblock In \emph{34th International Conference on Machine Learning, ICML 2017}, volume~6, pages 4709--4718, 2017.
\newblock ISBN 9781510855144.

\bibitem[Johansson et~al.(2018)Johansson, Kallus, Shalit, and Sontag]{Johansson2018LearningDesigns}
Fredrik~D. Johansson, Nathan Kallus, Uri Shalit, and David Sontag.
\newblock {Learning Weighted Representations for Generalization Across Designs}.
\newblock \emph{arXiv preprint arXiv:1802.08598}, 2 2018.

\bibitem[Louizos et~al.(2017)Louizos, Shalit, Mooij, Sontag, Zemel, and Welling]{Louizos2017CausalModels}
Christos Louizos, Uri Shalit, Joris~M Mooij, David Sontag, Richard Zemel, and Max Welling.
\newblock {Causal Effect Inference with Deep Latent-Variable Models}.
\newblock In I~Guyon, U~V Luxburg, S~Bengio, H~Wallach, R~Fergus, S~Vishwanathan, and R~Garnett, editors, \emph{Advances in Neural Information Processing Systems}, volume~30, pages 6446--6456. Curran Associates, Inc., 5 2017.

\bibitem[Shi et~al.(2019)Shi, Blei, and Veitch]{shi2019adaptingeffects}
Claudia Shi, David~M Blei, and Victor Veitch.
\newblock {Adapting neural networks for the estimation of treatment effects}.
\newblock \emph{arXiv preprint arXiv:1906.02120}, 2019.
\newblock ISSN 23318422.

\bibitem[Chipman et~al.(2010)Chipman, George, and McCulloch]{Chipman2010BART:Trees}
Hugh~A Chipman, Edward~I George, and Robert~E. McCulloch.
\newblock {BART: Bayesian additive regression trees}.
\newblock \emph{Annals of Applied Statistics}, 4\penalty0 (1):\penalty0 266--298, 2010.
\newblock ISSN 19326157.
\newblock \doi{10.1214/09-AOAS285}.

\bibitem[Hill(2011)]{Hill2011BayesianInference}
Jennifer~L Hill.
\newblock {Bayesian nonparametric modeling for causal inference}.
\newblock \emph{Journal of Computational and Graphical Statistics}, 20\penalty0 (1):\penalty0 217--240, 2011.
\newblock ISSN 10618600.
\newblock \doi{10.1198/jcgs.2010.08162}.

\bibitem[Rasmussen(2003)]{rasmussen2003gaussian}
Carl~Edward Rasmussen.
\newblock Gaussian processes in machine learning.
\newblock In \emph{Summer school on machine learning}, pages 63--71. Springer, 2003.

\bibitem[Shafer and Vovk(2008)]{shafer2008tutorial}
Glenn Shafer and Vladimir Vovk.
\newblock A tutorial on conformal prediction.
\newblock \emph{Journal of Machine Learning Research}, 9\penalty0 (3), 2008.

\bibitem[Lei and Cand{\`e}s(2021)]{lei2021conformal}
Lihua Lei and Emmanuel~J Cand{\`e}s.
\newblock Conformal inference of counterfactuals and individual treatment effects.
\newblock \emph{Journal of the Royal Statistical Society Series B: Statistical Methodology}, 83\penalty0 (5):\penalty0 911--938, 2021.

\bibitem[Yadlowsky et~al.(2018)Yadlowsky, Namkoong, Basu, Duchi, and Tian]{Yadlowsky2018BoundsFactors}
Steve Yadlowsky, Hongseok Namkoong, Sanjay Basu, John Duchi, and Lu~Tian.
\newblock {Bounds on the conditional and average treatment effect with unobserved confounding factors}.
\newblock \emph{arXiv preprint arXiv:1808.09521}, 8 2018.

\bibitem[Jesson et~al.(2020)Jesson, Mindermann, Shalit, and Gal]{jesson2020identifying}
Andrew Jesson, S{\"o}ren Mindermann, Uri Shalit, and Yarin Gal.
\newblock Identifying causal-effect inference failure with uncertainty-aware models.
\newblock \emph{Advances in Neural Information Processing Systems}, 33, 2020.

\bibitem[Qian and Murphy(2011)]{Qian2011PerformanceRules}
Min Qian and Susan~A. Murphy.
\newblock {Performance guarantees for individualized treatment rules}.
\newblock \emph{The Annals of Statistics}, 39\penalty0 (2):\penalty0 1180--1210, 4 2011.
\newblock ISSN 0090-5364.
\newblock \doi{10.1214/10-aos864}.

\bibitem[Dud{\'\i}k et~al.(2014)Dud{\'\i}k, Erhan, Langford, and Li]{dudik2014doubly}
Miroslav Dud{\'\i}k, Dumitru Erhan, John Langford, and Lihong Li.
\newblock Doubly robust policy evaluation and optimization.
\newblock \emph{Statistical Science}, 29\penalty0 (4):\penalty0 485--511, 2014.

\bibitem[Montoya et~al.(2022{\natexlab{a}})Montoya, van~der Laan, Skeem, and Petersen]{montoya2022estimators}
Lina~M Montoya, Mark~J van~der Laan, Jennifer~L Skeem, and Maya~L Petersen.
\newblock Estimators for the value of the optimal dynamic treatment rule with application to criminal justice interventions.
\newblock \emph{The International Journal of Biostatistics}, 2022{\natexlab{a}}.

\bibitem[Bertsimas et~al.(2019)Bertsimas, Orfanoudaki, and Weiner]{bertsimas2019personalizedapproach}
Dimitris Bertsimas, Agni Orfanoudaki, and Rory~B. Weiner.
\newblock {Personalized Treatment for Coronary Artery Disease Patients: A Machine Learning Approach}.
\newblock \emph{arXiv preprint arXiv:1910.08483}, 10 2019.

\bibitem[Athey and Wager(2017)]{Athey2017EfficientLearning}
Susan Athey and Stefan Wager.
\newblock {Efficient Policy Learning}.
\newblock \emph{arXiv preprint arXiv:1702.02896}, 2 2017.

\bibitem[Athey and Wager(2021)]{athey2021policy}
Susan Athey and Stefan Wager.
\newblock Policy learning with observational data.
\newblock \emph{Econometrica}, 89\penalty0 (1):\penalty0 133--161, 2021.

\bibitem[Kallus(2018)]{Kallus2018BalancedLearning}
Nathan Kallus.
\newblock {Balanced policy evaluation and learning}.
\newblock In \emph{Advances in Neural Information Processing Systems}, volume 2018-Decem, pages 8895--8906, 2018.

\bibitem[Montoya et~al.(2022{\natexlab{b}})Montoya, van~der Laan, Luedtke, Skeem, Coyle, and Petersen]{montoya2022optimal}
Lina~M Montoya, Mark~J van~der Laan, Alexander~R Luedtke, Jennifer~L Skeem, Jeremy~R Coyle, and Maya~L Petersen.
\newblock The optimal dynamic treatment rule superlearner: considerations, performance, and application to criminal justice interventions.
\newblock \emph{The International Journal of Biostatistics}, 2022{\natexlab{b}}.

\bibitem[Rubin(2004)]{rubin2004teaching}
Donald~B Rubin.
\newblock Teaching statistical inference for causal effects in experiments and observational studies.
\newblock \emph{Journal of Educational and Behavioral Statistics}, 29\penalty0 (3):\penalty0 343--367, 2004.

\bibitem[Dickerman et~al.(2019)Dickerman, Garc{\'\i}a-Alb{\'e}niz, Logan, Denaxas, and Hern{\'a}n]{dickerman2019avoidable}
Barbra~A Dickerman, Xabier Garc{\'\i}a-Alb{\'e}niz, Roger~W Logan, Spiros Denaxas, and Miguel~A Hern{\'a}n.
\newblock Avoidable flaws in observational analyses: an application to statins and cancer.
\newblock \emph{Nature medicine}, 25\penalty0 (10):\penalty0 1601--1606, 2019.

\bibitem[Sendak et~al.(2020)Sendak, D’Arcy, Kashyap, Gao, Nichols, Corey, Ratliff, and Balu]{sendak2020adelivery}
Mark~P. Sendak, Joshua D’Arcy, Sehj Kashyap, Michael Gao, Marshall Nichols, Kristin Corey, William Ratliff, and Suresh Balu.
\newblock {A Path for Translation of Machine Learning Products into Healthcare Delivery}.
\newblock \emph{EMJ Innovations}, 1 2020.
\newblock ISSN 2513-8634.
\newblock \doi{10.33590/emjinnov/19-00172}.

\bibitem[Tennant et~al.(2021)Tennant, Murray, Arnold, Berrie, Fox, Gadd, Harrison, Keeble, Ranker, Textor, et~al.]{tennant2021use}
Peter~WG Tennant, Eleanor~J Murray, Kellyn~F Arnold, Laurie Berrie, Matthew~P Fox, Sarah~C Gadd, Wendy~J Harrison, Claire Keeble, Lynsie~R Ranker, Johannes Textor, et~al.
\newblock Use of directed acyclic graphs (dags) to identify confounders in applied health research: review and recommendations.
\newblock \emph{International journal of epidemiology}, 50\penalty0 (2):\penalty0 620--632, 2021.

\bibitem[Simon(1955)]{simon1955behavioral}
Herbert~A Simon.
\newblock A behavioral model of rational choice.
\newblock \emph{The quarterly journal of economics}, pages 99--118, 1955.

\bibitem[Kushniruk(2001)]{kushniruk2001analysis}
Andre~W Kushniruk.
\newblock Analysis of complex decision-making processes in health care: cognitive approaches to health informatics.
\newblock \emph{Journal of biomedical informatics}, 34\penalty0 (5):\penalty0 365--376, 2001.

\bibitem[Patel et~al.(2002)Patel, Kaufman, and Arocha]{patel2002emerging}
Vimla~L Patel, David~R Kaufman, and Jose~F Arocha.
\newblock Emerging paradigms of cognition in medical decision-making.
\newblock \emph{Journal of biomedical informatics}, 35\penalty0 (1):\penalty0 52--75, 2002.

\bibitem[Pearl(1995)]{pearl1995causal}
Judea Pearl.
\newblock Causal diagrams for empirical research.
\newblock \emph{Biometrika}, 82\penalty0 (4):\penalty0 669--688, 1995.

\bibitem[Duarte et~al.(2015)Duarte, Monnez, Albuisson, Pitt, Zannad, and Rossignol]{duarte2015prognostic}
K{\'e}vin Duarte, Jean-Marie Monnez, Eliane Albuisson, Bertram Pitt, Faiez Zannad, and Patrick Rossignol.
\newblock Prognostic value of estimated plasma volume in heart failure.
\newblock \emph{JACC: Heart Failure}, 3\penalty0 (11):\penalty0 886--893, 2015.

\bibitem[Ding et~al.(2017)Ding, VanderWeele, and Robins]{ding2017instrumental}
Peng Ding, TJ~VanderWeele, and James~M Robins.
\newblock Instrumental variables as bias amplifiers with general outcome and confounding.
\newblock \emph{Biometrika}, 104\penalty0 (2):\penalty0 291--302, 2017.

\bibitem[Sauer et~al.(2013)Sauer, Brookhart, Roy, and VanderWeele]{sauer2013covariate}
Brian Sauer, M~Alan Brookhart, Jason~A Roy, and Tyler~J VanderWeele.
\newblock Covariate selection.
\newblock In \emph{Developing a protocol for observational comparative effectiveness research: a user's guide}. Agency for Healthcare Research and Quality (US), 2013.

\bibitem[Huenermund et~al.(2022)Huenermund, Louw, and R{\"o}nkk{\"o}]{huenermund2022choice}
Paul Huenermund, Beyers Louw, and Mikko R{\"o}nkk{\"o}.
\newblock The choice of control variables: How causal graphs can inform the decision.
\newblock In \emph{Academy of Management Proceedings}, page 294. Academy of Management Briarcliff Manor, NY 10510, 2022.
\newblock \doi{10.5465/AMBPP.2022.294}.

\bibitem[Chen and Guestrin(2016)]{Chen2016XGBoost}
Tianqi Chen and Carlos Guestrin.
\newblock {XGBoost}.
\newblock In \emph{Proceedings of the 22nd ACM SIGKDD International Conference on Knowledge Discovery and Data Mining - KDD '16}, pages 785--794, 3 2016.
\newblock ISBN 9781450342322.
\newblock \doi{10.1145/2939672.2939785}.

\bibitem[Susan M~Shortreed(2017)]{Shortreed2017}
Ashkan~Ertefaie Susan M~Shortreed.
\newblock Outcome adaptive lasso: Variable selection for causal inference.
\newblock \emph{Biometrics}, 73\penalty0 (4):\penalty0 1111--1122, dec 2017.
\newblock ISSN 0006-341X.
\newblock \doi{10.1111/biom.12679}.

\bibitem[Gutman et~al.(2022{\natexlab{a}})Gutman, Karavani, and Shimoni]{gutman2022propensity}
Rom Gutman, Ehud Karavani, and Yishai Shimoni.
\newblock Propensity score models are better when post-calibrated.
\newblock \emph{arXiv preprint arXiv:2211.01221}, 2022{\natexlab{a}}.

\bibitem[Platt et~al.(1999)]{platt1999probabilistic}
John Platt et~al.
\newblock Probabilistic outputs for support vector machines and comparisons to regularized likelihood methods.
\newblock \emph{Advances in large margin classifiers}, 10\penalty0 (3):\penalty0 61--74, 1999.

\bibitem[Zadrozny and Elkan(2001)]{zadrozny2001obtaining}
Bianca Zadrozny and Charles Elkan.
\newblock Obtaining calibrated probability estimates from decision trees and naive bayesian classifiers.
\newblock In \emph{Icml}, volume~1, pages 609--616, 2001.

\bibitem[Lundberg and Lee(2017{\natexlab{a}})]{lundberg2017unified}
Scott~M Lundberg and Su-In Lee.
\newblock A unified approach to interpreting model predictions.
\newblock \emph{Advances in neural information processing systems}, 30, 2017{\natexlab{a}}.

\bibitem[Dehejia and Wahba(1999)]{dehejia1999causal}
Rajeev~H Dehejia and Sadek Wahba.
\newblock Causal effects in nonexperimental studies: Reevaluating the evaluation of training programs.
\newblock \emph{Journal of the American statistical Association}, 94\penalty0 (448):\penalty0 1053--1062, 1999.

\bibitem[Fogarty et~al.(2016)Fogarty, Mikkelsen, Gaieski, and Small]{fogarty2016discrete}
Colin~B Fogarty, Mark~E Mikkelsen, David~F Gaieski, and Dylan~S Small.
\newblock Discrete optimization for interpretable study populations and randomization inference in an observational study of severe sepsis mortality.
\newblock \emph{Journal of the American Statistical Association}, 111\penalty0 (514):\penalty0 447--458, 2016.

\bibitem[Crump et~al.(2009)Crump, Hotz, Imbens, and Mitnik]{crump2009dealing}
Richard~K Crump, V~Joseph Hotz, Guido~W Imbens, and Oscar~A Mitnik.
\newblock Dealing with limited overlap in estimation of average treatment effects.
\newblock \emph{Biometrika}, 96\penalty0 (1):\penalty0 187--199, 2009.

\bibitem[D’Amour et~al.(2021)D’Amour, Ding, Feller, Lei, and Sekhon]{d2021overlap}
Alexander D’Amour, Peng Ding, Avi Feller, Lihua Lei, and Jasjeet Sekhon.
\newblock Overlap in observational studies with high-dimensional covariates.
\newblock \emph{Journal of Econometrics}, 221\penalty0 (2):\penalty0 644--654, 2021.

\bibitem[Curth and Van Der~Schaar(2023)]{curth2023magic}
Alicia Curth and Mihaela Van Der~Schaar.
\newblock In search of insights, not magic bullets: Towards demystification of the model selection dilemma in heterogeneous treatment effect estimation.
\newblock In Andreas Krause, Emma Brunskill, Kyunghyun Cho, Barbara Engelhardt, Sivan Sabato, and Jonathan Scarlett, editors, \emph{Proceedings of the 40th International Conference on Machine Learning}, volume 202 of \emph{Proceedings of Machine Learning Research}, pages 6623--6642. PMLR, 23--29 Jul 2023.

\bibitem[Curth et~al.(2021)Curth, Svensson, Weatherall, and van~der Schaar]{curth2021really}
Alicia Curth, David Svensson, Jim Weatherall, and Mihaela van~der Schaar.
\newblock Really doing great at estimating cate? a critical look at ml benchmarking practices in treatment effect estimation.
\newblock In \emph{Thirty-fifth conference on neural information processing systems datasets and benchmarks track (round 2)}, 2021.

\bibitem[Agniel et~al.(2018)Agniel, Kohane, and Weber]{agniel2018biases}
Denis Agniel, Isaac~S Kohane, and Griffin~M Weber.
\newblock Biases in electronic health record data due to processes within the healthcare system: retrospective observational study.
\newblock \emph{Bmj}, 361, 2018.

\bibitem[Gutman et~al.(2022{\natexlab{b}})Gutman, Aronson, Caspi, and Shalit]{Gutman2022WhatOutcome}
Rom Gutman, Doron Aronson, Oren Caspi, and Uri Shalit.
\newblock {What drives performance in machine learning models for predicting heart failure outcome?}
\newblock \emph{European Heart Journal - Digital Health}, 9 2022{\natexlab{b}}.
\newblock \doi{10.1093/EHJDH/ZTAC054}.

\bibitem[Athey and Wager(2019)]{athey2019estimating}
Susan Athey and Stefan Wager.
\newblock Estimating treatment effects with causal forests: An application.
\newblock \emph{Observational Studies}, 5\penalty0 (2):\penalty0 37--51, 2019.

\bibitem[Glynn and Quinn(2010)]{aipw}
Adam~N Glynn and Kevin~M Quinn.
\newblock An introduction to the augmented inverse propensity weighted estimator.
\newblock \emph{Political analysis}, pages 36--56, 2010.

\bibitem[Leete et~al.(2019)Leete, Kallus, Hudgens, Napravnik, and Kosorok]{leete2019balanceddata}
Owen~E Leete, Nathan Kallus, Michael~G Hudgens, Sonia Napravnik, and Michael~R Kosorok.
\newblock Balanced policy evaluation and learning for right censored data.
\newblock \emph{arXiv preprint arXiv:1911.05728}, 2019.

\bibitem[Imai and Li(2023)]{imai2023experimental}
Kosuke Imai and Michael~Lingzhi Li.
\newblock Experimental evaluation of individualized treatment rules.
\newblock \emph{Journal of the American Statistical Association}, 118\penalty0 (541):\penalty0 242--256, 2023.

\bibitem[Lunceford and Davidian(2004)]{lunceford2004stratificationstudy}
Jared~K Lunceford and Marie Davidian.
\newblock {Stratification and weighting via the propensity score in estimation of causal treatment effects: A comparative study}.
\newblock \emph{Statistics in Medicine}, 23\penalty0 (19):\penalty0 2937--2960, 2004.
\newblock ISSN 02776715.
\newblock \doi{10.1002/sim.1903}.

\bibitem[Lage et~al.(2019)Lage, Lifschitz, Doshi-Velez, and Amir]{lage2019toward}
Isaac Lage, Daphna Lifschitz, Finale Doshi-Velez, and Ofra Amir.
\newblock Toward robust policy summarization.
\newblock \emph{Autonomous agents and multi-agent systems}, 2019:\penalty0 2081, 2019.

\bibitem[Matsson and Johansson(2022)]{matsson2022case}
Anton Matsson and Fredrik~D Johansson.
\newblock Case-based off-policy evaluation using prototype learning.
\newblock In \emph{Uncertainty in Artificial Intelligence}, pages 1339--1349. PMLR, 2022.

\bibitem[McDonagh et~al.(2021)McDonagh, Metra, Adamo, Gardner, Baumbach, Böhm, Burri, Butler, Čelutkienė, Chioncel, Cleland, Coats, Crespo-Leiro, Farmakis, Gilard, Heymans, Hoes, Jaarsma, Jankowska, Lainscak, Lam, Lyon, McMurray, Mebazaa, Mindham, Muneretto, Francesco~Piepoli, Price, Rosano, Ruschitzka, Kathrine~Skibelund, and Group]{mcdonagh2021}
Theresa~A McDonagh, Marco Metra, Marianna Adamo, Roy~S Gardner, Andreas Baumbach, Michael Böhm, Haran Burri, Javed Butler, Jelena Čelutkienė, Ovidiu Chioncel, John G~F Cleland, Andrew J~S Coats, Maria~G Crespo-Leiro, Dimitrios Farmakis, Martine Gilard, Stephane Heymans, Arno~W Hoes, Tiny Jaarsma, Ewa~A Jankowska, Mitja Lainscak, Carolyn S~P Lam, Alexander~R Lyon, John J~V McMurray, Alexandre Mebazaa, Richard Mindham, Claudio Muneretto, Massimo Francesco~Piepoli, Susanna Price, Giuseppe M~C Rosano, Frank Ruschitzka, Anne Kathrine~Skibelund, and ESC Scientific~Document Group.
\newblock {2021 ESC Guidelines for the diagnosis and treatment of acute and chronic heart failure: Developed by the Task Force for the diagnosis and treatment of acute and chronic heart failure of the European Society of Cardiology (ESC) With the special contribution of the Heart Failure Association (HFA) of the ESC}.
\newblock \emph{European Heart Journal}, 42\penalty0 (36):\penalty0 3599--3726, 08 2021.
\newblock ISSN 0195-668X.
\newblock \doi{10.1093/eurheartj/ehab368}.

\bibitem[Mutlak et~al.(2018)Mutlak, Lessick, Khalil, Yalonetsky, Agmon, and Aronson]{Mutlak2018TricuspidRisk}
Diab Mutlak, Jonathan Lessick, Shehrban Khalil, Sergey Yalonetsky, Yoram Agmon, and Doron Aronson.
\newblock {Tricuspid regurgitation in acute heart failure: is there any incremental risk?}
\newblock \emph{European Heart Journal - Cardiovascular Imaging}, 19\penalty0 (9):\penalty0 993--1001, 9 2018.
\newblock ISSN 2047-2404.
\newblock \doi{10.1093/ehjci/jex343}.

\bibitem[Aronson et~al.(2013)Aronson, Darawsha, Atamna, Kaplan, Makhoul, Mutlak, Lessick, Carasso, Reisner, Agmon, Dragu, and Azzam]{Aronson2013PulmonaryFailure}
Doron Aronson, Wisam Darawsha, Aula Atamna, Marielle Kaplan, Badira~F. Makhoul, Diab Mutlak, Jonathan Lessick, Shemy Carasso, Shimon Reisner, Yoram Agmon, Robert Dragu, and Zaher~S. Azzam.
\newblock {Pulmonary Hypertension, Right Ventricular Function, and Clinical Outcome in Acute Decompensated Heart Failure}.
\newblock \emph{Journal of Cardiac Failure}, 19\penalty0 (10):\penalty0 665--671, 10 2013.
\newblock ISSN 10719164.
\newblock \doi{10.1016/j.cardfail.2013.08.007}.

\bibitem[Ponikowski et~al.(2016)Ponikowski, Voors, Anker, Bueno, Cleland, Coats, Falk, Gonz{\'{a}}lez-Juanatey, Harjola, Jankowska, Jessup, Linde, Nihoyannopoulos, Parissis, Pieske, Riley, Rosano, Ruilope, Ruschitzka, Rutten, and van~der Meer]{Ponikowski20162016Failure}
Piotr Ponikowski, Adriaan~A. Voors, Stefan~D. Anker, Héctor Bueno, John G.~F. Cleland, Andrew J.~S. Coats, Volkmar Falk, José~Ramón Gonz{\'{a}}lez-Juanatey, Veli-Pekka Harjola, Ewa~A. Jankowska, Mariell Jessup, Cecilia Linde, Petros Nihoyannopoulos, John~T. Parissis, Burkert Pieske, Jillian~P. Riley, Giuseppe M.~C. Rosano, Luis~M. Ruilope, Frank Ruschitzka, Frans~H. Rutten, and Peter van~der Meer.
\newblock {2016 ESC Guidelines for the diagnosis and treatment of acute and chronic heart failure}.
\newblock \emph{European Journal of Heart Failure}, 18\penalty0 (8):\penalty0 891--975, 8 2016.
\newblock ISSN 13889842.
\newblock \doi{10.1002/ejhf.592}.

\bibitem[Lundberg and Lee(2017{\natexlab{b}})]{Lundberg2017APredictions}
Scott~M Lundberg and Su~In Lee.
\newblock {A unified approach to interpreting model predictions}.
\newblock In \emph{Advances in Neural Information Processing Systems}, volume 2017-Decem, pages 4766--4775, 5 2017{\natexlab{b}}.

\bibitem[Dang et~al.(2023)Dang, Gruber, Lee, Dahabreh, Stuart, Williamson, Wyss, D{\'\i}az, Ghosh, K{\i}c{\i}man, et~al.]{dang2023causal}
Lauren~E Dang, Susan Gruber, Hana Lee, Issa Dahabreh, Elizabeth~A Stuart, Brian~D Williamson, Richard Wyss, Iv{\'a}n D{\'\i}az, Debashis Ghosh, Emre K{\i}c{\i}man, et~al.
\newblock A causal roadmap for generating high-quality real-world evidence.
\newblock \emph{arXiv preprint arXiv:2305.06850}, 2023.

\bibitem[Kravitz et~al.(2004)Kravitz, Duan, and Braslow]{kravitz2004evidence}
Richard~L. Kravitz, Naihua Duan, and Joel Braslow.
\newblock {Evidence-based medicine, heterogeneity of treatment effects, and the trouble with averages}.
\newblock \emph{Milbank Quarterly}, 82\penalty0 (4):\penalty0 661--687, 12 2004.
\newblock ISSN 0887378X.
\newblock \doi{10.1111/j.0887-378X.2004.00327.x}.

\bibitem[Hayward et~al.(2005)Hayward, Kent, Vijan, and Hofer]{hayward2005reporting}
Rodney~A Hayward, David~M Kent, Sandeep Vijan, and Timothy~P Hofer.
\newblock Reporting clinical trial results to inform providers, payers, and consumers.
\newblock \emph{Health Affairs}, 24\penalty0 (6):\penalty0 1571--1581, 2005.

\bibitem[Dahabreh et~al.(2016)Dahabreh, Hayward, and Kent]{Dahabreh2016UsingEvidence}
Issa~J. Dahabreh, Rodney Hayward, and David~M. Kent.
\newblock {Using group data to treat individuals: Understanding heterogeneous treatment effects in the age of precision medicine and patient-centred evidence}.
\newblock \emph{International Journal of Epidemiology}, 45\penalty0 (6):\penalty0 2184--2193, 12 2016.
\newblock ISSN 14643685.
\newblock \doi{10.1093/ije/dyw125}.

\bibitem[Chen et~al.(2017)Chen, Tian, Cai, and Yu]{Chen2017AScoring}
Shuai Chen, Lu~Tian, Tianxi Cai, and Menggang Yu.
\newblock {A general statistical framework for subgroup identification and comparative treatment scoring}.
\newblock \emph{Biometrics}, 73\penalty0 (4):\penalty0 1199--1209, 2017.
\newblock ISSN 15410420.
\newblock \doi{10.1111/biom.12676}.

\bibitem[Anoke et~al.(2019)Anoke, Normand, and Zigler]{anoke2019approachesconfounding}
Sarah~C. Anoke, Sharon~Lise Normand, and Corwin~M. Zigler.
\newblock {Approaches to treatment effect heterogeneity in the presence of confounding}.
\newblock \emph{Statistics in Medicine}, 38\penalty0 (15):\penalty0 2797--2815, 7 2019.
\newblock ISSN 10970258.
\newblock \doi{10.1002/sim.8143}.

\bibitem[Luedtke and Van Der~Laan(2017)]{Luedtke2017EvaluatingSubgroup}
Alexander~R. Luedtke and Mark~J. Van Der~Laan.
\newblock {Evaluating the impact of treating the optimal subgroup}.
\newblock \emph{Statistical Methods in Medical Research}, 26\penalty0 (4):\penalty0 1630--1640, 8 2017.
\newblock ISSN 14770334.
\newblock \doi{10.1177/0962280217708664}.

\bibitem[Ling et~al.(2023)Ling, Upadhyaya, Chen, Jiang, and Kim]{ling2023emulate}
Yaobin Ling, Pulakesh Upadhyaya, Luyao Chen, Xiaoqian Jiang, and Yejin Kim.
\newblock Emulate randomized clinical trials using heterogeneous treatment effect estimation for personalized treatments: methodology review and benchmark.
\newblock \emph{Journal of Biomedical Informatics}, 137:\penalty0 104256, 2023.

\bibitem[Kent et~al.(2020)Kent, Paulus, Van~Klaveren, D'Agostino, Goodman, Hayward, Ioannidis, Patrick-Lake, Morton, Pencina, et~al.]{kent2020predictive}
David~M Kent, Jessica~K Paulus, David Van~Klaveren, Ralph D'Agostino, Steve Goodman, Rodney Hayward, John~PA Ioannidis, Bray Patrick-Lake, Sally Morton, Michael Pencina, et~al.
\newblock The predictive approaches to treatment effect heterogeneity (path) statement.
\newblock \emph{Annals of internal medicine}, 172\penalty0 (1):\penalty0 35--45, 2020.

\bibitem[Sharma and K{\i}c{\i}man(2020)]{sharma2020dowhy}
Amit Sharma and Emre K{\i}c{\i}man.
\newblock Dowhy: An end-to-end library for causal inference.
\newblock \emph{arXiv preprint arXiv:2011.04216}, 2020.
\newblock \doi{10.48550/arxiv.2011.04216}.

\bibitem[Boominathan et~al.(2020)Boominathan, Oberst, Zhou, Kanjilal, and Sontag]{boominathan2020treatment}
Soorajnath Boominathan, Michael Oberst, Helen Zhou, Sanjat Kanjilal, and David Sontag.
\newblock {Treatment Policy Learning in Multiobjective Settings with Fully Observed Outcomes}.
\newblock In \emph{Proceedings of the ACM SIGKDD International Conference on Knowledge Discovery and Data Mining}, pages 1937--1947, 2020.
\newblock ISBN 9781450379984.
\newblock \doi{10.1145/3394486.3403245}.

\bibitem[Meid et~al.(2022)Meid, Wirbka, Group, Groll, and Haefeli]{meid2022can}
Andreas~D Meid, Lucas Wirbka, ARMIN~Study Group, Andreas Groll, and Walter~E Haefeli.
\newblock Can machine learning from real-world data support drug treatment decisions? a prediction modeling case for direct oral anticoagulants.
\newblock \emph{Medical Decision Making}, 42\penalty0 (5):\penalty0 587--598, 2022.

\bibitem[Petersen and van~der Laan(2014)]{petersen2014causal}
Maya~L Petersen and Mark~J van~der Laan.
\newblock Causal models and learning from data: integrating causal modeling and statistical estimation.
\newblock \emph{Epidemiology (Cambridge, Mass.)}, 25\penalty0 (3):\penalty0 418, 2014.

\bibitem[Bates et~al.(2020)Bates, Auerbach, Schulam, Wright, and Saria]{Bates2020ReportingIntelligence}
David~W. Bates, Andrew Auerbach, Peter Schulam, Adam Wright, and Suchi Saria.
\newblock {Reporting and Implementing Interventions Involving Machine Learning and Artificial Intelligence}.
\newblock \emph{Annals of internal medicine}, 172\penalty0 (11):\penalty0 S137--S144, 6 2020.
\newblock ISSN 15393704.
\newblock \doi{10.7326/M19-0872}.

\bibitem[Rajpurkar et~al.(2022)Rajpurkar, Chen, Banerjee, and Topol]{rajpurkar2022ai}
Pranav Rajpurkar, Emma Chen, Oishi Banerjee, and Eric~J Topol.
\newblock {AI} in health and medicine.
\newblock \emph{Nature medicine}, 28\penalty0 (1):\penalty0 31--38, 2022.

\bibitem[Harris et~al.(2022)Harris, Bonnici, Keen, Lilaonitkul, White, and Swanepoel]{harris2022clinical}
Steve Harris, Tim Bonnici, Thomas Keen, Watjana Lilaonitkul, Mark~J White, and Nel Swanepoel.
\newblock Clinical deployment environments: Five pillars of translational machine learning for health.
\newblock \emph{Frontiers in Digital Health}, 4:\penalty0 939292, 2022.

\bibitem[Zhang et~al.(2022)Zhang, Xing, Zou, and Wu]{zhang2022shifting}
Angela Zhang, Lei Xing, James Zou, and Joseph~C Wu.
\newblock Shifting machine learning for healthcare from development to deployment and from models to data.
\newblock \emph{Nature Biomedical Engineering}, 6\penalty0 (12):\penalty0 1330--1345, 2022.

\bibitem[Heaven(2021)]{heaven2021hundreds}
Will~Douglas Heaven.
\newblock Hundreds of {AI} tools have been built to catch {C}ovid. none of them helped.
\newblock \emph{MIT Technology Review. Retrieved October}, 6:\penalty0 2021, 2021.

\bibitem[Smit et~al.(2023)Smit, Krijthe, Kant, Labrecque, Komorowski, Gommers, van Bommel, Reinders, and van Genderen]{smit2023causal}
JM~Smit, JH~Krijthe, WMR Kant, JA~Labrecque, M~Komorowski, DAMPJ Gommers, J~van Bommel, MJT Reinders, and ME~van Genderen.
\newblock Causal inference using observational intensive care unit data: a scoping review and recommendations for future practice.
\newblock \emph{npj Digital Medicine}, 6\penalty0 (1):\penalty0 221, 2023.

\bibitem[Balagopalan et~al.(2024)Balagopalan, Baldini, Celi, Gichoya, McCoy, Naumann, Shalit, van~der Schaar, and Wagstaff]{balagopalan2024machine}
Aparna Balagopalan, Ioana Baldini, Leo~Anthony Celi, Judy Gichoya, Liam~G McCoy, Tristan Naumann, Uri Shalit, Mihaela van~der Schaar, and Kiri~L Wagstaff.
\newblock Machine learning for healthcare that matters: Reorienting from technical novelty to equitable impact.
\newblock \emph{PLOS Digital Health}, 3\penalty0 (4):\penalty0 e0000474, 2024.

\bibitem[McCradden et~al.(2020)McCradden, Stephenson, and Anderson]{McCradden2020ClinicalIntelligence}
Melissa~D. McCradden, Elizabeth~A. Stephenson, and James~A. Anderson.
\newblock {Clinical research underlies ethical integration of healthcare artificial intelligence}.
\newblock \emph{Nature Medicine 2020 26:9}, 26\penalty0 (9):\penalty0 1325--1326, 9 2020.
\newblock ISSN 1546-170X.
\newblock \doi{10.1038/s41591-020-1035-9}.

\bibitem[Wiesenfeld et~al.(2022)Wiesenfeld, Aphinyanaphongs, and Nov]{wiesenfeld2022ai}
Batia~Mishan Wiesenfeld, Yin Aphinyanaphongs, and Oded Nov.
\newblock {AI} model transferability in healthcare: a sociotechnical perspective.
\newblock \emph{Nature Machine Intelligence}, 4\penalty0 (10):\penalty0 807--809, 2022.

\bibitem[Joshi et~al.(2025)Joshi, Urteaga, van Amsterdam, Hripcsak, Elias, Recht, Elhadad, Fackler, Sendak, Wiens, et~al.]{joshi2025ai}
Shalmali Joshi, I{\~n}igo Urteaga, Wouter~AC van Amsterdam, George Hripcsak, Pierre Elias, Benjamin Recht, No{\'e}mie Elhadad, James Fackler, Mark~P Sendak, Jenna Wiens, et~al.
\newblock {AI} as an intervention: improving clinical outcomes relies on a causal approach to {AI} development and validation.
\newblock \emph{Journal of the American Medical Informatics Association}, page ocae301, 2025.

\bibitem[Behar et~al.(2023)Behar, Levy, and Celi]{behar2023generalization}
Joachim~A Behar, Jeremy Levy, and Leo~Anthony Celi.
\newblock Generalization in medical ai: a perspective on developing scalable models.
\newblock \emph{arXiv preprint arXiv:2311.05418}, 2023.

\bibitem[Rubin(2006)]{rubin2006causal1}
Donald~B Rubin.
\newblock {Causal Inference Through Potential Outcomes and Principal Stratification: Application to Studies with "Censoring" Due to Death 1}.
\newblock \emph{Statistical Science}, 21\penalty0 (3):\penalty0 299--309, 2006.
\newblock \doi{10.1214/088342306000000114}.

\bibitem[Pocock et~al.(2012)Pocock, Ariti, Collier, and Wang]{pocock2012win}
Stuart~J Pocock, Cono~A Ariti, Timothy~J Collier, and Duolao Wang.
\newblock The win ratio: a new approach to the analysis of composite endpoints in clinical trials based on clinical priorities.
\newblock \emph{European heart journal}, 33\penalty0 (2):\penalty0 176--182, 2012.

\bibitem[Even and Josse(2025)]{even2025rethinking}
Mathieu Even and Julie Josse.
\newblock Rethinking the win ratio: A causal framework for hierarchical outcome analysis.
\newblock \emph{arXiv preprint arXiv:2501.16933}, 2025.

\bibitem[Murphy(2003)]{murphy2003optimal}
Susan~A Murphy.
\newblock Optimal dynamic treatment regimes.
\newblock \emph{Journal of the Royal Statistical Society Series B: Statistical Methodology}, 65\penalty0 (2):\penalty0 331--355, 2003.

\bibitem[Gottesman et~al.(2019)Gottesman, Johansson, Komorowski, Faisal, Sontag, Doshi-Velez, and Celi]{gottesman2019guidelines}
Omer Gottesman, Fredrik Johansson, Matthieu Komorowski, Aldo Faisal, David Sontag, Finale Doshi-Velez, and Leo~Anthony Celi.
\newblock Guidelines for reinforcement learning in healthcare.
\newblock \emph{Nature medicine}, 25\penalty0 (1):\penalty0 16--18, 2019.

\bibitem[Madras et~al.(2018)Madras, Pitassi, and Zemel]{madras2018predict}
David Madras, Toni Pitassi, and Richard Zemel.
\newblock Predict responsibly: improving fairness and accuracy by learning to defer.
\newblock \emph{Advances in Neural Information Processing Systems}, 31, 2018.

\bibitem[Mozannar and Sontag(2020)]{mozannar2020consistent}
Hussein Mozannar and David Sontag.
\newblock Consistent estimators for learning to defer to an expert.
\newblock In \emph{International Conference on Machine Learning}, pages 7076--7087. PMLR, 2020.

\bibitem[Ghoummaid and Shalit(2024)]{ghoummaidact}
Marah Ghoummaid and Uri Shalit.
\newblock When to act and when to ask: Policy learning with deferral under hidden confounding.
\newblock In \emph{The Thirty-eighth Annual Conference on Neural Information Processing Systems}, 2024.

\bibitem[Gao and Yin(2023)]{gao2023confounding}
Ruijiang Gao and Mingzhang Yin.
\newblock Confounding-robust policy improvement with human-{AI} teams.
\newblock \emph{arXiv preprint arXiv:2310.08824}, 2023.

\bibitem[Bansal et~al.(2021)Bansal, Nushi, Kamar, Horvitz, and Weld]{bansal2021most}
Gagan Bansal, Besmira Nushi, Ece Kamar, Eric Horvitz, and Daniel~S Weld.
\newblock Is the most accurate {AI} the best teammate? optimizing {AI} for teamwork.
\newblock \emph{Proceedings of the AAAI Conference on Artificial Intelligence}, 35\penalty0 (13):\penalty0 11405--11414, 2021.

\bibitem[Jacobs et~al.(2021)Jacobs, Pradier, McCoy~Jr, Perlis, Doshi-Velez, and Gajos]{jacobs2021machine}
Maia Jacobs, Melanie~F Pradier, Thomas~H McCoy~Jr, Roy~H Perlis, Finale Doshi-Velez, and Krzysztof~Z Gajos.
\newblock How machine-learning recommendations influence clinician treatment selections: the example of antidepressant selection.
\newblock \emph{Translational psychiatry}, 11\penalty0 (1):\penalty0 108, 2021.

\bibitem[Meyer et~al.(2022)Meyer, Khademi, T{\^e}tu, Han, Nippak, and Remisch]{meyer2022impact}
Julien Meyer, April Khademi, Bernard T{\^e}tu, Wencui Han, Pria Nippak, and David Remisch.
\newblock Impact of artificial intelligence on pathologists’ decisions: an experiment.
\newblock \emph{Journal of the American Medical Informatics Association}, 29\penalty0 (10):\penalty0 1688--1695, 2022.

\bibitem[Finlayson et~al.(2021)Finlayson, Subbaswamy, Singh, Bowers, Kupke, Zittrain, Kohane, and Saria]{finlayson2021clinician}
Samuel~G Finlayson, Adarsh Subbaswamy, Karandeep Singh, John Bowers, Annabel Kupke, Jonathan Zittrain, Isaac~S Kohane, and Suchi Saria.
\newblock The clinician and dataset shift in artificial intelligence.
\newblock \emph{New England Journal of Medicine}, 385\penalty0 (3):\penalty0 283--286, 2021.

\bibitem[Perdomo et~al.(2020)Perdomo, Zrnic, Mendler-D{\"u}nner, and Hardt]{perdomo2020performative}
Juan Perdomo, Tijana Zrnic, Celestine Mendler-D{\"u}nner, and Moritz Hardt.
\newblock Performative prediction.
\newblock In \emph{International Conference on Machine Learning}, pages 7599--7609. PMLR, 2020.

\bibitem[Tchetgen et~al.(2020)Tchetgen, Ying, Cui, Shi, and Miao]{tchetgen2020introduction}
Eric J~Tchetgen Tchetgen, Andrew Ying, Yifan Cui, Xu~Shi, and Wang Miao.
\newblock An introduction to proximal causal learning.
\newblock \emph{arXiv preprint arXiv:2009.10982}, 2020.

\bibitem[Shi et~al.(2020)Shi, Miao, and Tchetgen]{shi2020selective}
Xu~Shi, Wang Miao, and Eric~Tchetgen Tchetgen.
\newblock A selective review of negative control methods in epidemiology.
\newblock \emph{Current epidemiology reports}, 7\penalty0 (4):\penalty0 190--202, 2020.

\bibitem[Sverdrup and Cui(2023)]{sverdrup2023proximal}
Erik Sverdrup and Yifan Cui.
\newblock Proximal causal learning of conditional average treatment effects.
\newblock In \emph{International Conference on Machine Learning}, pages 33285--33298. PMLR, 2023.

\bibitem[Gheorghiade and Pang(2009)]{Gheorghiade2009AcuteSyndromes}
Mihai Gheorghiade and Peter~S. Pang.
\newblock {Acute Heart Failure Syndromes}.
\newblock \emph{Journal of the American College of Cardiology}, 53\penalty0 (7):\penalty0 557--573, 2 2009.
\newblock ISSN 07351097.
\newblock \doi{10.1016/j.jacc.2008.10.041}.

\bibitem[Laffey and Kavanagh(2018)]{Laffey2018NegativeWrong.}
John~G Laffey and Brian~P Kavanagh.
\newblock {Negative trials in critical care: why most research is probably wrong.}
\newblock \emph{The Lancet. Respiratory medicine}, 6\penalty0 (9):\penalty0 659--660, 9 2018.
\newblock ISSN 2213-2619.
\newblock \doi{10.1016/S2213-2600(18)30279-0}.

\bibitem[McMurray et~al.(2014)McMurray, Packer, Desai, Gong, Lefkowitz, Rizkala, Rouleau, Shi, Solomon, Swedberg, and Zile]{McMurray2014AngiotensinNeprilysinFailure}
John~J.V. McMurray, Milton Packer, Akshay~S. Desai, Jianjian Gong, Martin~P. Lefkowitz, Adel~R. Rizkala, Jean~L. Rouleau, Victor~C. Shi, Scott~D. Solomon, Karl Swedberg, and Michael~R. Zile.
\newblock {Angiotensin–Neprilysin Inhibition versus Enalapril in Heart Failure}.
\newblock \emph{New England Journal of Medicine}, 371\penalty0 (11):\penalty0 993--1004, 9 2014.
\newblock ISSN 0028-4793.
\newblock \doi{10.1056/NEJMoa1409077}.

\bibitem[Massie et~al.(2010)Massie, O'Connor, Metra, Ponikowski, Teerlink, Cotter, Weatherley, Cleland, Givertz, Voors, DeLucca, Mansoor, Salerno, Bloomfield, and Dittrich]{Massie2010RolofyllineFailure}
Barry~M. Massie, Christopher~M. O'Connor, Marco Metra, Piotr Ponikowski, John~R. Teerlink, Gad Cotter, Beth~Davison Weatherley, John~G.F. Cleland, Michael~M. Givertz, Adriaan Voors, Paul DeLucca, George~A. Mansoor, Christina~M. Salerno, Daniel~M. Bloomfield, and Howard~C. Dittrich.
\newblock {Rolofylline, an Adenosine A 1 −Receptor Antagonist, in Acute Heart Failure}.
\newblock \emph{New England Journal of Medicine}, 363\penalty0 (15):\penalty0 1419--1428, 10 2010.
\newblock ISSN 0028-4793.
\newblock \doi{10.1056/NEJMoa0912613}.

\bibitem[O'Connor et~al.(2011)O'Connor, Starling, Hernandez, Armstrong, Dickstein, Hasselblad, Heizer, Komajda, Massie, McMurray, Nieminen, Reist, Rouleau, Swedberg, Adams, Anker, Atar, Battler, Botero, Bohidar, Butler, Clausell, Corbal{\'{a}}n, Costanzo, Dahlstrom, Deckelbaum, Diaz, Dunlap, Ezekowitz, Feldman, Felker, Fonarow, Gennevois, Gottlieb, Hill, Hollander, Howlett, Hudson, Kociol, Krum, Laucevicius, Levy, M{\'{e}}ndez, Metra, Mittal, Oh, Pereira, Ponikowski, Tang, Tanomsup, Teerlink, Triposkiadis, Troughton, Voors, Whellan, Zannad, and Califf]{OConnor2011EffectFailure}
C.M. O'Connor, R.C. Starling, A.F. Hernandez, P.W. Armstrong, K.~Dickstein, V.~Hasselblad, G.M. Heizer, M.~Komajda, B.M. Massie, J.J.V. McMurray, M.S. Nieminen, C.J. Reist, J.L. Rouleau, K.~Swedberg, K.F. Adams, S.D. Anker, D.~Atar, A.~Battler, R.~Botero, N.R. Bohidar, J.~Butler, N.~Clausell, R.~Corbal{\'{a}}n, M.R. Costanzo, U.~Dahlstrom, L.I. Deckelbaum, R.~Diaz, M.E. Dunlap, J.A. Ezekowitz, D.~Feldman, G.M. Felker, G.C. Fonarow, D.~Gennevois, S.S. Gottlieb, J.A. Hill, J.E. Hollander, J.G. Howlett, M.P. Hudson, R.D. Kociol, H.~Krum, A.~Laucevicius, W.C. Levy, G.F. M{\'{e}}ndez, M.~Metra, S.~Mittal, B.-H. Oh, N.L. Pereira, P.~Ponikowski, W.H.W. Tang, S.~Tanomsup, J.R. Teerlink, F.~Triposkiadis, R.W. Troughton, A.A. Voors, D.J. Whellan, F.~Zannad, and R.M. Califf.
\newblock {Effect of Nesiritide in Patients with Acute Decompensated Heart Failure}.
\newblock \emph{New England Journal of Medicine}, 365\penalty0 (1):\penalty0 32--43, 7 2011.
\newblock ISSN 0028-4793.
\newblock \doi{10.1056/NEJMoa1100171}.

\bibitem[Goldberg et~al.(2009)Goldberg, Kogan, Hammerman, Markiewicz, and Aronson]{Goldberg2009TheInfarction}
Alexander Goldberg, Elena Kogan, Haim Hammerman, Walter Markiewicz, and Doron Aronson.
\newblock {The impact of transient and persistent acute kidney injury on long-term outcomes after acute myocardial infarction}.
\newblock \emph{Kidney International}, 76\penalty0 (8):\penalty0 900--906, 10 2009.
\newblock ISSN 0085-2538.
\newblock \doi{10.1038/KI.2009.295}.

\bibitem[Butler et~al.(2004)Butler, Forman, Abraham, Gottlieb, Loh, Massie, O'Connor, Rich, Stevenson, Wang, Young, and Krumholz]{Butler2004RelationshipPatients}
Javed Butler, Daniel~E Forman, William~T Abraham, Stephen~S Gottlieb, Evan Loh, Barry~M Massie, Christopher~M O'Connor, Michael~W Rich, Lynne~Warner Stevenson, Yongfei Wang, James~B Young, and Harlan~M Krumholz.
\newblock {Relationship between heart failure treatment and development of worsening renal function among hospitalized patients}.
\newblock \emph{American Heart Journal}, 147\penalty0 (2):\penalty0 331--338, 2 2004.
\newblock ISSN 00028703.
\newblock \doi{10.1016/j.ahj.2003.08.012}.

\bibitem[Kane et~al.(2017)Kane, Kim, Haidry, Salciccioli, and Lazar]{Kane2017Discontinuation/doseFraction}
Jesse~A. Kane, Joseph~K. Kim, Syed~Abbas Haidry, Louis Salciccioli, and Jason Lazar.
\newblock {Discontinuation/dose reduction of angiotensin-converting enzyme inhibitors/angiotensin receptor blockers during acute decompensated heart failure in African-American patients with reduced left-ventricular ejection fraction}.
\newblock \emph{Cardiology (Switzerland)}, 137\penalty0 (2):\penalty0 121--125, 2017.
\newblock ISSN 14219751.
\newblock \doi{10.1159/000457946}.

\bibitem[B{\"{o}}hm et~al.(2011)B{\"{o}}hm, Link, Cai, Nieminen, Filippatos, Salem, Solal, Huang, Padley, Kivikko, and Mebazaa]{Bohm2011BeneficialTrial}
Michael B{\"{o}}hm, Andreas Link, Danlin Cai, Markku~S. Nieminen, Gerasimos~S. Filippatos, Reda Salem, Alain~Cohen Solal, Bidan Huang, Robert~J. Padley, Matti Kivikko, and Alexandre Mebazaa.
\newblock {Beneficial association of {$\beta$}-blocker therapy on recovery from severe acute heart failure treatment: Data from the Survival of Patients with Acute Heart Failure in Need of Intravenous Inotropic Support trial}.
\newblock \emph{Critical Care Medicine}, 39\penalty0 (5):\penalty0 940--944, 5 2011.
\newblock ISSN 15300293.
\newblock \doi{10.1097/CCM.0b013e31820a91ed}.

\end{thebibliography}
